\documentclass[twoside,11pt]{article}

\usepackage{jmlr2e}

\usepackage{color}

\jmlrheading{x}{2012}{xxx}{xx/xx}{xx/xx}{H.C. Burger, C.J. Schuler and S. Harmeling}

\ShortHeadings{Image denoising with multi-layer perceptrons, part 2}{H.C. Burger, C.J. Schuler and S. Harmeling}
\firstpageno{1}

\graphicspath{
  {figures2/}
}

\begin{document}

\title{Image denoising with multi-layer perceptrons, part 2:\\
training trade-offs and analysis of their mechanisms}

\author{\name Harold Christopher Burger \email burger@tuebingen.mpg.de \\
       \name Christian J. Schuler \email cschuler@tuebingen.mpg.de \\
       \name Stefan Harmeling  \email harmeling@tuebingen.mpg.de \\
       \addr Max Planck Institute for Intelligent Systems\\
       Spemannstr. 38\\
       72076 T\"{u}bingen, Germany
       }

\editor{}

\maketitle

\begin{abstract}
Image denoising can be described as the problem of mapping from a noisy image
to a noise-free image. In \citet{burgerjmlr1}, we show that multi-layer
perceptrons can achieve outstanding image denoising performance for various
types of noise (additive white Gaussian noise, mixed Poisson-Gaussian noise,
JPEG artifacts, salt-and-pepper noise and noise resembling stripes).  In this
work we discuss in detail which trade-offs have to be considered during the
training procedure. We will show how to achieve good results and which pitfalls
to avoid.  By analysing the activation patterns of the hidden units we are able
to make observations regarding the functioning principle of multi-layer
perceptrons trained for image denoising.
\end{abstract}

\tableofcontents

\section{Introduction}
In \citet{burgerjmlr1}, we show that multi-layer perceptrons (MLPs) mapping a
noisy image patch to a denoised image patch are able to achieve outstanding
image denoising results, even surpassing the previous
state-of-the-art~\citep{dabov2007image}. In addition, the MLPs outperform one
type of theoretical bound in image denoising~\citep{chatterjee2010denoising}
and come a long way toward closing the gap to a second type of theoretical
bound~\citep{levin2012patch}. Related work in image denoising is also discussed
in \citet{burgerjmlr1}. This paper explains the technical trade-offs to achieve
those results.

Achieving good results with MLPs was possible through the use of larger patch
sizes: It is known that larger patch sizes help make the denoising problem less
ambiguous~\citep{levin2010natural}. However, large patches also make the
denoising problem more difficult (the function is higher dimensional). This 
required us to train high-capacity MLPs on a large number of training samples.
Training such MLPs is therefore time-consuming, though modern GPUs
alleviate the problem somewhat.

Training neural networks, especially large ones, is usually performed using
stochastic gradient descent and is sometimes considered more of an art than a
science.  While there exist ``tricks'' to make training
efficient~\citep{lecun1998efficient,GlorotAISTATS2010}, it is still quite
possible that some experimental setups will lead to poor results. In these
cases, it is often poorly understood why the results are bad. One might 
sometimes attribute these bad results to ``bad luck'' such as an unlucky weight
initialization. This becomes a problem especially for time-consuming
large-scale experiments, where multiple restarts are simply not possible. It is
therefore crucial to understand which setups are likely to lead to good results
and which to bad results before launching an experiment.

A common criticism regarding neural networks is that they are ``black boxes'':
Given a neural network, one can merely observe its output for a given input.
The inner workings or logic are usually not open for inspection. Under certain
circumstances, this is not the case: Convolutional neural
networks~\citep{lecun1998gradient} are usually easier to interpret for humans
because the hidden representations can be represented as
images~\citep{lee2009convolutional}. More recently,
\citet{erhan2010understanding} have proposed an \emph{activation maximization}
procedure to find an input maximizing the activation of a hidden unit, and have
shown that this procedure allows for better qualitative evaluation of a
network.

\paragraph{Contributions:} This paper aims to address the above two issues for
MLPs trained to denoise image patches. In the first part of this paper, we
provide a detailed description of a large and varied set of large-scale
experiments. We will discuss various trade-offs encountered during the training
procedure. Certain settings of training parameters can lead to initially good
results, but later lead to a \emph{catastrophic} degradation in performance.
This phenomenon is highly undesirable and we will provide guidelines on how to
avoid it, as well as an explanation of such phenomena.

In the second part of this paper, we show that surprisingly, it is possible to
gain insight into the operating principle or inner workings of an MLP trained
on image denoising. This is the least difficult for MLPs with a single hidden
layer, but we will show that MLPs with more hidden layers are also
interpretable through analysis of the activation patterns of the hidden units.
We also gain insight about denoising auto-encoders \citep{vincent2010stacked}
due to their similarity to our MLPs.

\paragraph{Notation and definitions:} For an MLP with four hidden layers, each
containing $2047$ hidden units, input patches of size $39\times39$ pixels and
output patches of size $17\times17$ pixels, we use the following notation
$(39\times39,2047,2047,2047,2047,17\times17) \equiv (39,4\times2047,17)$. If
the input and output patches are of the same size, we use the following
notation $(17,4\times2047)$ to denote an MLP with four hidden layers of size
$2047$ and input and output patches of size $17\times17$ pixels.

We will periodically halt the training procedure of an MLP and report the
\emph{test performance}, by which we mean the average PSNR achieved on the $11$
standard test images defined in \citep{burgerjmlr1}. When we report the
\emph{training performance}, we mean the average PSNR achieved on the last
$2\times10^{6}$ training samples. The test performance therefore refers to
image denoising performance, whereas the training performance refers to patch
denoising performance.

\section{Training trade-offs to achieve good results with MLPs}
\label{sec:progressduringtraining}
In \citet{burgerjmlr1} we showed that it is possible to achieve
state-of-the-art image denoising results with MLPs. This section will show what
steps are necessary to achieve these results. We do so by tracking the
evolution of the results for different experimental setups during the training
process. In particular, we will vary the size of the training dataset as well
as the architecture of the MLPs.  We will mostly use AWG noise with
$\sigma=25$. Each experiment is the result of many days and sometimes even
weeks of computation time on a modern GPU (we used nVidia's C2050).

\subsection{Long training times do not result in overfitting}
\label{sec:nooverfitting}
In this section, we will use a much smaller training set as the one defined in
\citet{burgerjmlr1}. We will use the $200$ training images from the BSDS300
dataset, which is a subset of the BSDS500 dataset.

\begin{figure}[t]
  \centering
    \includegraphics[width=\columnwidth]{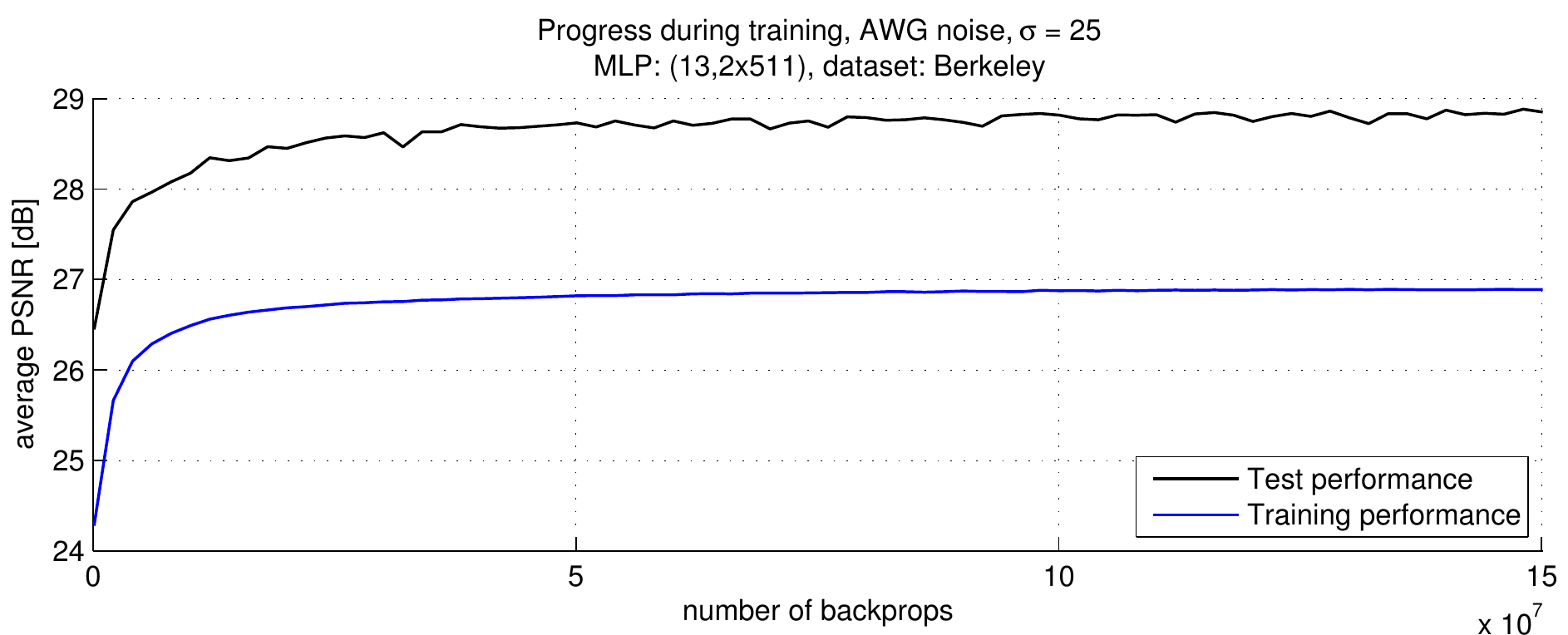}
    \caption{No overfitting even after many updates due to an abundance of training data.}
  \label{fig:progress001}
\end{figure}
We train an MLP with architecture $(13,2\times511)$.  We
report both the training performance and the test performance. The reason why
the test performance is superior to the training performance is that the test
performance refers to the image denoising performance (as opposed to the patch
denoising performance).  The image denoising performance is better than the
patch denoising performance because of the averageing procedure in areas where
patches overlap.  We observe that the training and test performance improve
steadily during the first few million updates.  Results still improve after
$10^8$ updates, albeit more slowly.  On the test set, results occasionally
briefly become worse. We also see that there is no overfitting even though we
are using a rather small training set. This is due to the abundance of training
data (the probability that a noisy patch is seen twice is zero). These results
suggest that overfitting is not an issue.

\subsection{Larger architectures are usually better}
\label{sec:largerarchitectures}
\begin{figure}[t]
  \centering
    \includegraphics[width=\columnwidth]{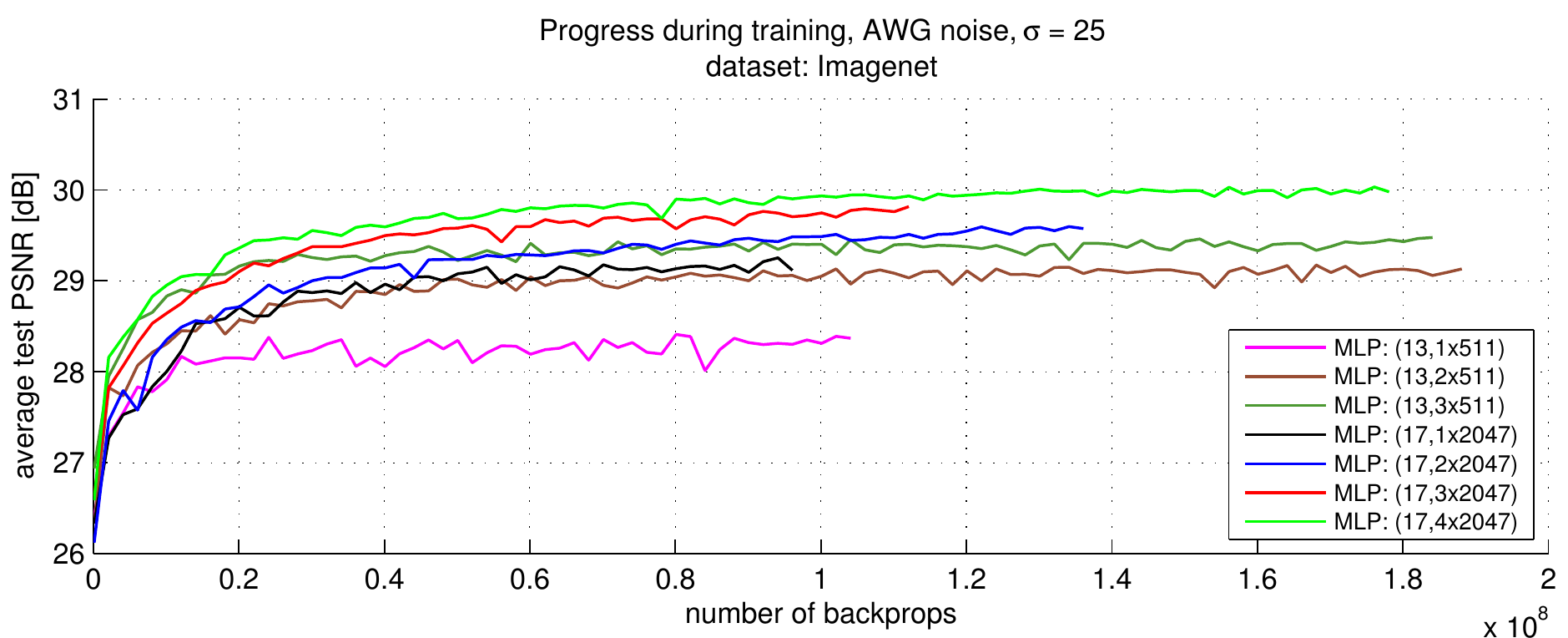}
    \caption{More hidden layers help. Three small hidden layers outperform one large hidden layer.}
  \label{fig:progress004}
\end{figure}
We now use the full training set---as defined in \citet{burgerjmlr1}---and train
various MLPs.  The size of the patches was either $13\times13$ or $17\times17$.
When the patch size was $13\times13$, we used hidden layers with $511$ units.
When the patch size was $17$, we used hidden layers with $2047$ units. We
varied the number of hidden layers, see Figure~\ref{fig:progress004}.

Adding hidden layers seems to always help. Larger patch sizes and wider hidden
layers seem to be beneficial. However, the MLP using patches of size
$13\times13$ and three hidden layers of size $511$ outperforms the MLP using
patches of size $17\times17$ and a single hidden layer of size $2047$.

\begin{figure}[t]
  \centering
    \includegraphics[width=\columnwidth]{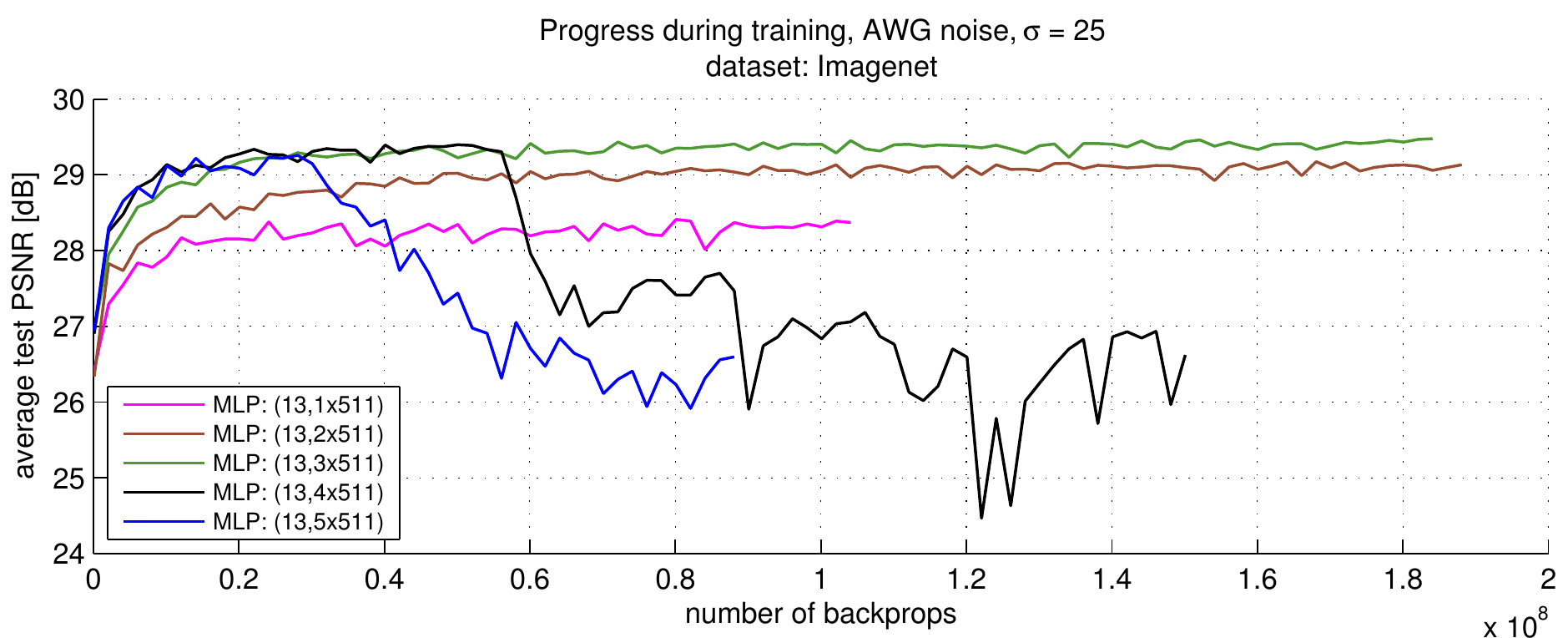}
    \caption{More hidden layers usually help. Too many hidden layers with few
    hidden units cause a catastrophic degradation in performance.}
  \label{fig:progress002}
\end{figure}
\paragraph{Is it always beneficial to add hidden layers?} To answer this
question, we train MLPs with patches of size $13\times13$ and hidden layers of
size $511$ with four and five hidden layers, see Figure~\ref{fig:progress002}.
The MLPs with four and five hidden layers perform well during the beginning of
the training procedure, but experience a significant decrease in performance
later on. The MLP achieving the best performance overall has three hidden
layers. We therefore conclude that it is not always beneficial to add hidden
layers.

A possible explanation for the degradation of performance shown in
Figure~\ref{fig:progress002} is that MLPs with more hidden layers become more
difficult to learn. Indeed, each hidden layer adds non-linearities to the
model. It is therefore possible that the error landscape is complex and
that stochastic gradient descent gets stuck in a poor local optimum from which
it is difficult to escape. In Figure~\ref{fig:progress004}, we see that an
MLP with patches of size $17\times17$ and four hidden layers of size $2047$
does not experience the effect shown in Figure~\ref{fig:progress002}, which is
an indication that deep and narrow networks are more difficult to optimize than
deep and wide networks.

\subsection{A larger training corpus is always better}
We have seen that longer training times lead to better results. Therefore,
seeing more training samples helps the MLPs achieve good results. 

\begin{figure}[t]
  \centering
    \includegraphics[width=\columnwidth]{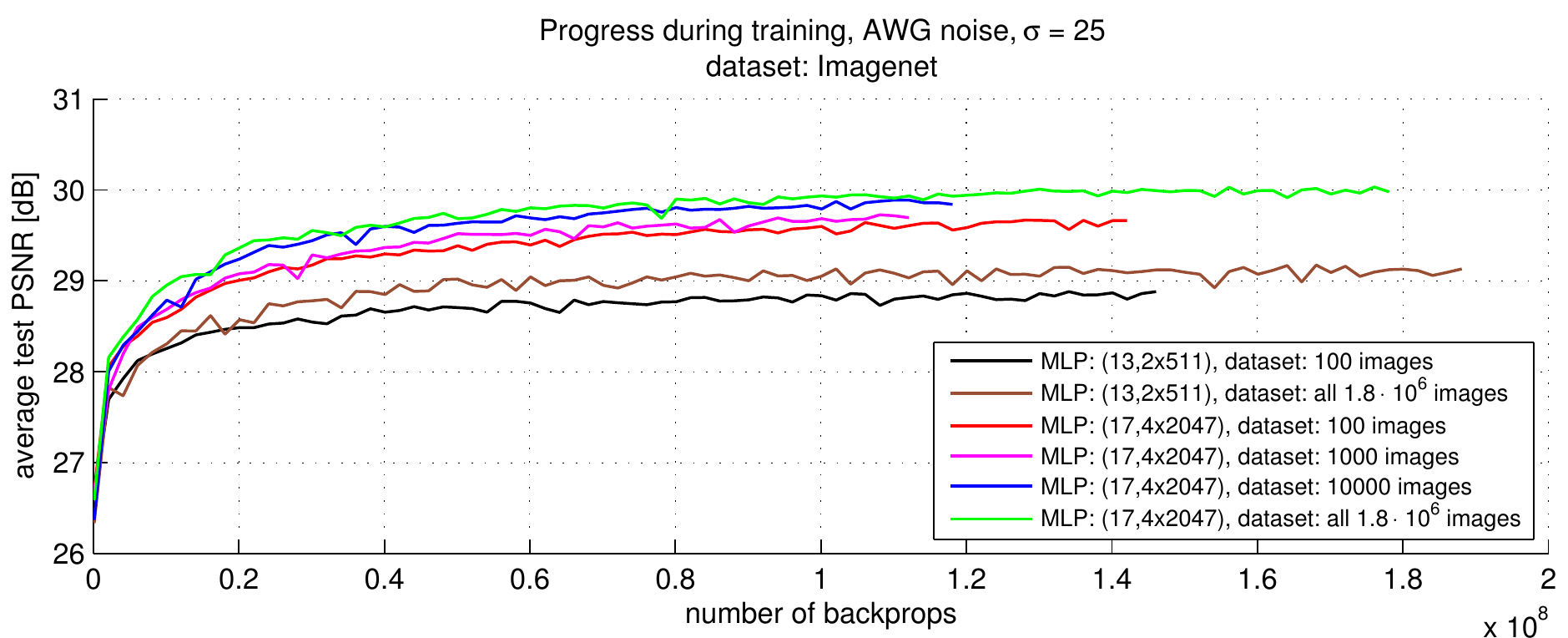}
    \caption{Using a larger corpus of training data helps.}
  \label{fig:progress003}
\end{figure}
We now ask the question: What is the effect of the number of images in the
training corpus? To this end, we have trained MLPs with identical architectures
on training sets of different sizes, see Figure~\ref{fig:progress003}. We used
either the full ImageNet training set or various subsets ($100$, $1000$ and
$10000$ images) of the same training set. We see that significant gains can be
obtained from using more training images. In particular, using even $10000$
training images delivers results that are clearly worse than results
obtained when training on the full ($\sim 1.8\cdot10^6$ image) training set. 
We also never observe a degradation in performance by using more training
images.

\subsection{The trade-off between small and large patches}
\label{sec:idealpatchsize}
We ask the question: Is it better to use small or large patches? We first
restrict ourselves to situations where the input and output patches are of the
same size.

\begin{figure}[t]
  \centering
    \includegraphics[width=\columnwidth]{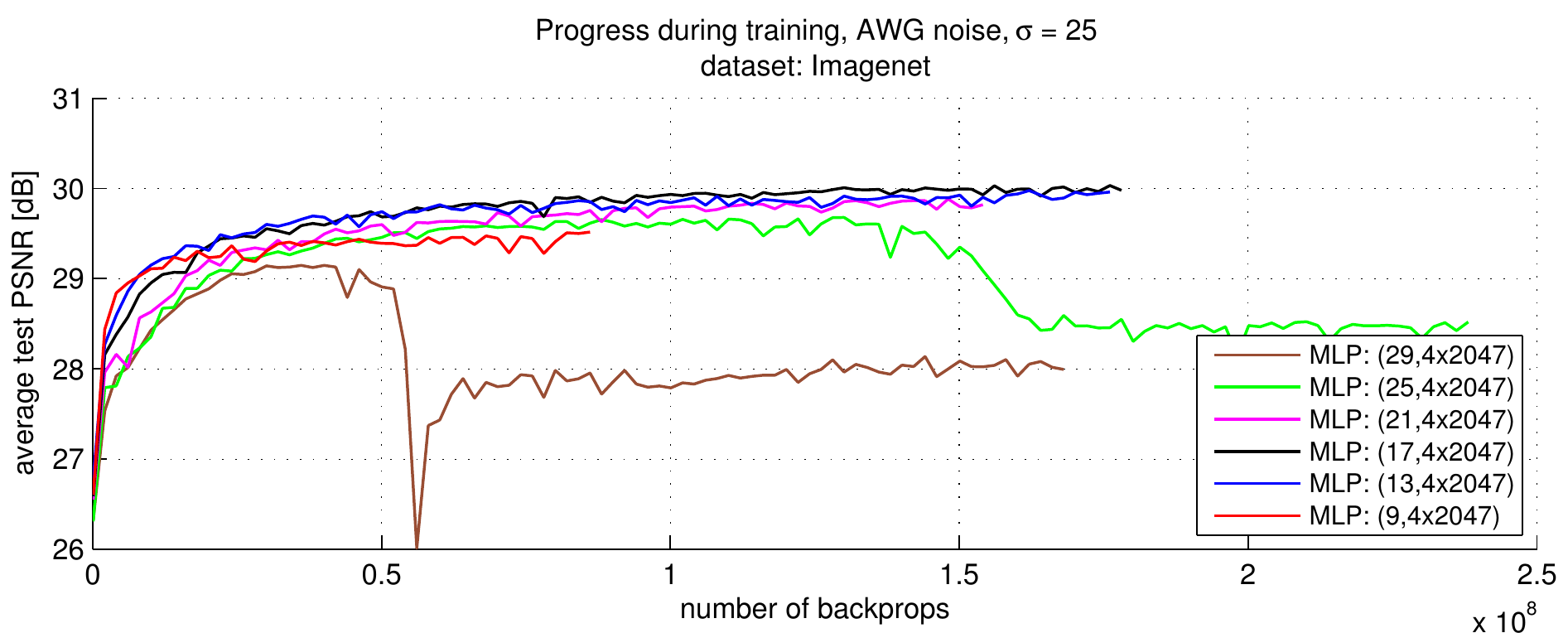}
    \caption{Larger patches lead to better results, up to a point.}
  \label{fig:progress005}
\end{figure} 
Figure~\ref{fig:progress005} shows the results obtained with MLPs with four
hidden layers of size $2047$ and various patch sizes. We see that up to a patch
size of $17\times17$, an increase in patch size leads to better results. This
is in agreement with~\cite{levin2010natural}: Using a larger support size makes
the denoising problem less ambiguous.

However, increasing the patch size further leads to worse results. The results
obtained using patches of size $21\times21$ are worse than those obtained using
patches of size $17\times17$. Using patches of size $25\times25$ leads to
results that are still worse and even leads to a degradation in performance
after approximately $10^8$ updates. For patches of size $29\times29$ we observe
still worse results and a deterioration of results after approximately
$5\cdot10^7$ updates. The performance later recovers slightly, but never
reaches the levels achieved before the degradation in performance.  
For this observation, we provide an explanation similar to the one provided in
section~\ref{sec:largerarchitectures}: Larger patch sizes increase the
dimensionality of the problem and therefore also the difficulty.  The model is
therefore more difficult to optimize when large patches are used, and
stochastic gradient descent may fail.

Therefore, when the input and output patches are of the same size, an ideal
patch size exists (for our architectures, it seems to be approximately
$17\times17$). Patches that are too small result in a denoising \emph{function}
that does not deliver good results, whereas patches that are too large results
in a \emph{model} that is difficult to optimize.

\paragraph{Larger input than output patches:} What happens when we remove
the restriction that the input patches be of the same size as the output
patches? We expect bad results when the output patches are larger than the
input patches: This would require hallucinating part of the patch. A more
interesting question is: What happens when the output patches are smaller than
the input patches?

\begin{figure}[t]
  \centering
    \includegraphics[width=\columnwidth]{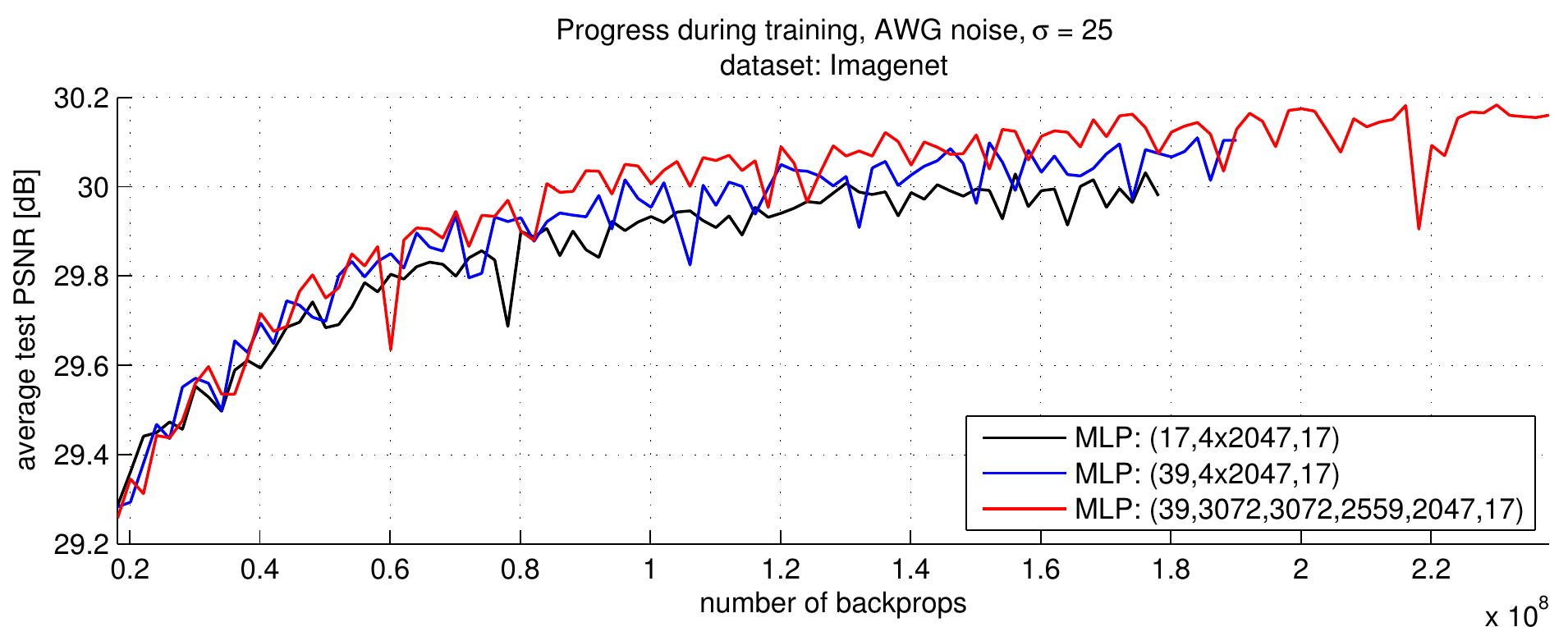}
    \caption{Larger input patches help.}
  \label{fig:progress008}
\end{figure}
Figure~\ref{fig:progress008} shows that using input patches that are larger
than the output patches delivers slightly better results. Using an architecture
with even more hidden units leads to even slightly better results.

\begin{figure}[t]
  \centering
    \includegraphics[width=\columnwidth]{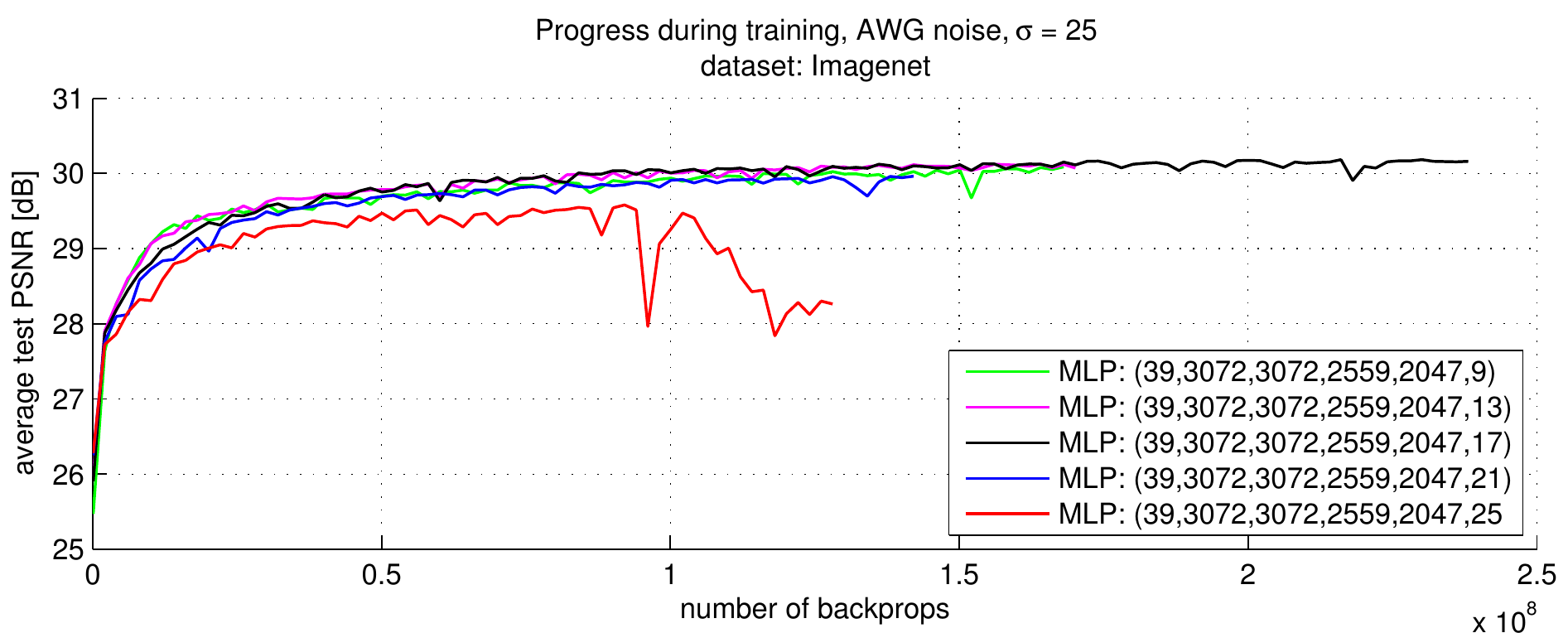}
    \caption{For a given input patch size, there exists an ideal output patch
    size. Output patches that are too large can create problems.}
  \label{fig:progress012}
\end{figure}
We now keep the size of the input patches fixed at $39\times39$ pixels and vary
the size of the output patches, see Figure~\ref{fig:progress012}. We observe
that increasing the size of the output patches helps only up to a point, after
which we observe a degradation in performance. The ideal output patch size
seems to be the same as when the input and output patches are of the same size
($17\times17$). Our explanation is again that output patches that are too large
result in a model that is difficult to optimize.

Finally, we investigate if the patch size has an effect on the best choice of
architecture.
\begin{figure}[t]
  \centering
    \includegraphics[width=\columnwidth]{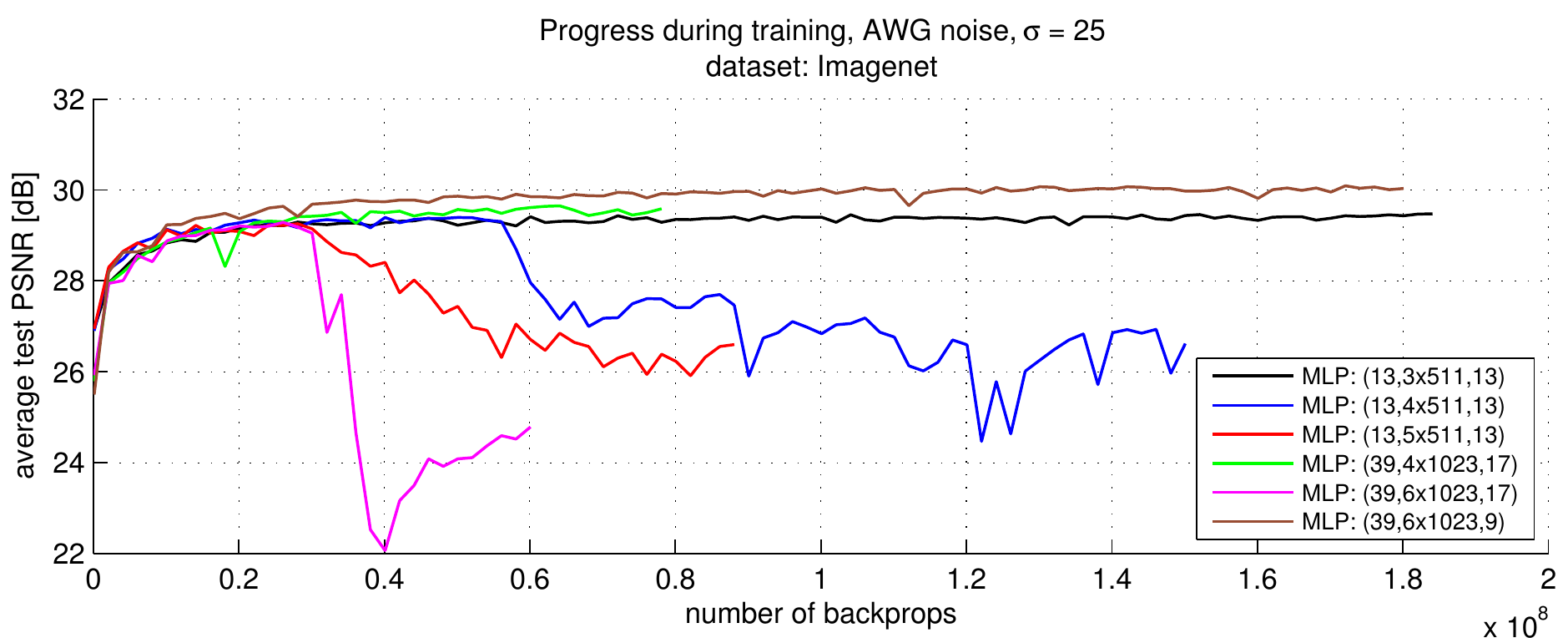}
    \caption{Too many hidden layers combined with large output patches creates problems.}
  \label{fig:progress006}
\end{figure}
Figure~\ref{fig:progress006} shows the results obtained with different patch
sizes and architectures. We see again that with hidden layers of size $511$,
using more than three hidden layers creates a degradation of performance when
combined with patches of size $13\times13$. With hidden layers of size $1023$,
four hidden layers combined with input patches of size $39\times39$ and output
patches of size $17\times17$, no degradation in performance is observed.  Using
the same patch sizes with six hidden layers of size $1023$ quickly results in a
degradation in performance. However, using the same architecture, but using
output patches of size $9\times9$ results in no degradation in performance and
even yields the best results in this comparison. We therefore conclude that it
is the combination of deep and narrow networks combined with large output
patches that are the most difficult to optimize. 

\paragraph{Conclusions concerning MLP architectures:} We have learned that
hidden layers with more units are always beneficial. Similarly, larger input
patches are also always helpful. However, too many hidden layers may lead to
problems in the training procedure. Problems are more likely to occur if the
hidden layers contain few hidden units or if the  size of the output patches is
large.

\subsection{Important gains in performance through ``fine-tuning''}
\label{sec:finetuning}
In all previous experiments, we observed that the test error fluctuates
slightly.  We attempt to avoid or at least reduce this behavior using a
``fine-tuning'' procedure: We initially train with a large learning rate and
later switch to a lower learning rate. The large learning rate is supposed to
encourage faster learning, whereas the low learning rate is supposed to
encourage more stable results on the test data.
\begin{figure}[htbp]
  \centering
    \includegraphics[width=\columnwidth]{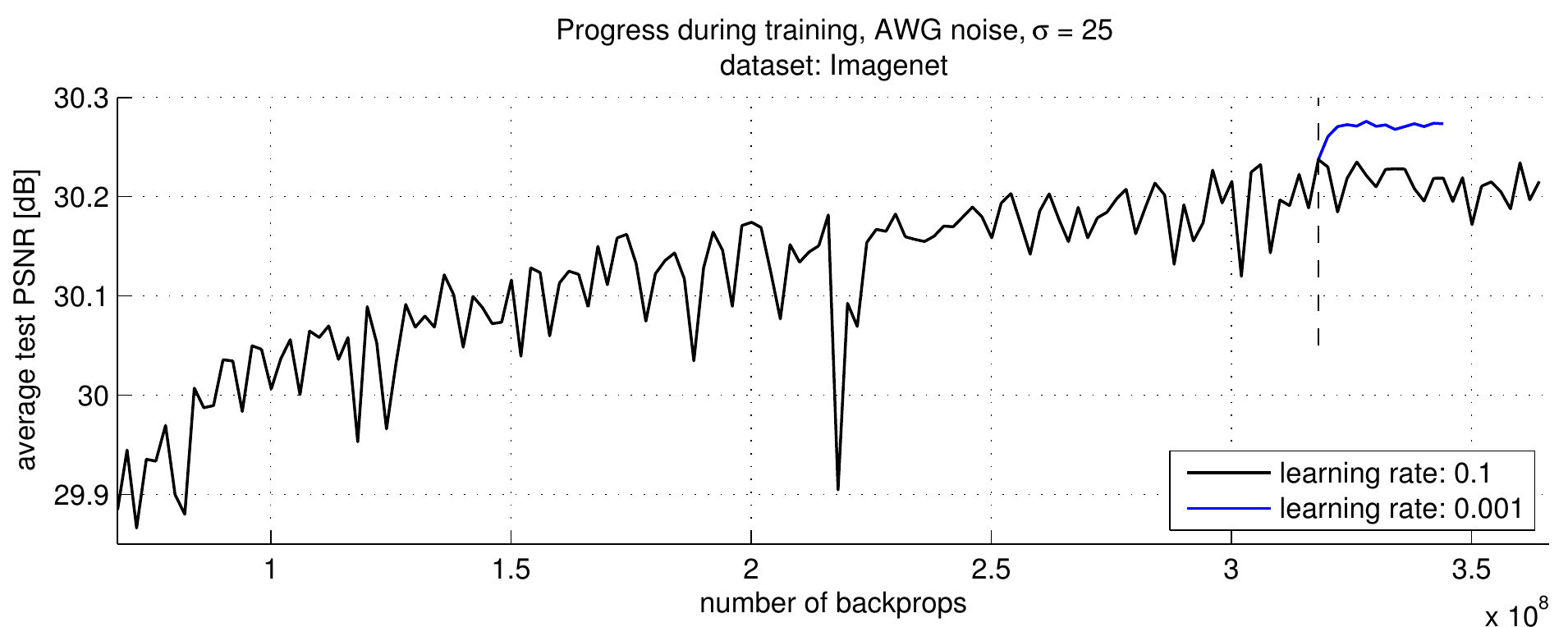}
    \caption{Fine-tuning improves results and reduces fluctuations in the test
    error. The vertical dashed line indicates where the learning rate was
    switched from $0.1$ to $0.001$. The two curves therefore only disagree
    starting at the dashed vertical line.}
  \label{fig:progress010}
\end{figure}
Figure~\ref{fig:progress010} shows that we can indeed reduce fluctuations in
the test error using a fine-tuning procedure. In addition, the switch to a
lower learning rate leads to an improvement of approximately $0.05$dB on the
test set. We conclude that it is important to use a fine-turning procedure to
obtain good results.

\subsection{Other noise variances: smaller patches for lower noise}
\label{sec:othernoisevariances}
\begin{figure}[t]
  \centering
    \includegraphics[width=\columnwidth]{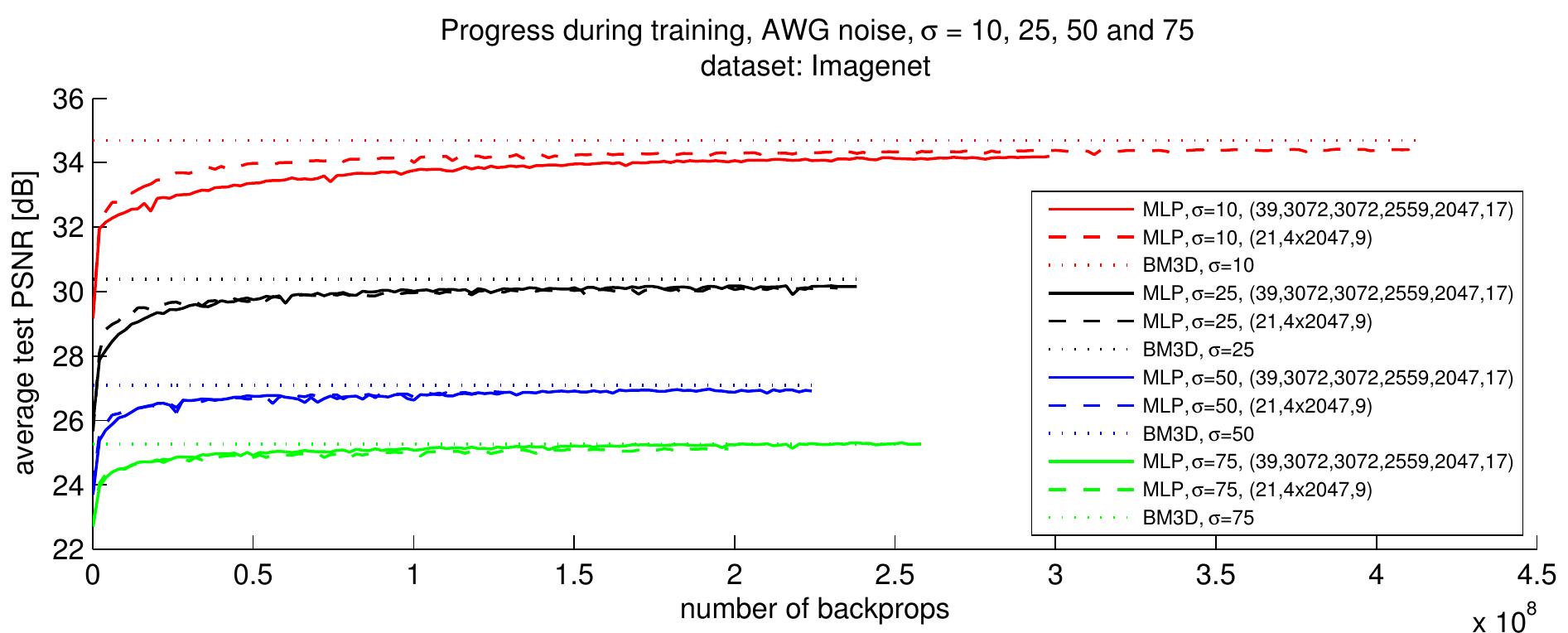}
    \caption{Progress on different noise levels compared to BM3D. The higher the noise level, the faster the progress.}
  \label{fig:progress009_nobm}
\end{figure}

Figure~\ref{fig:progress009_nobm} shows the improvement of the test results
(the average result obtained on the $11$ standard test images) during training
for different values of $\sigma$. The test results achieved by the MLPs is
compared against the test results achieved by BM3D. We used input patches of
size $39\times39$, output patches of size $17\times17$ and hidden layers of
sizes $3072$, $3072$, $2559$ and $2047$. We also experimented with smaller
patches (``smaller patches'' in Figure~\ref{fig:progress009_nobm}): Input
patches of size $21\times21$ and output patches of size $9\times9$. In that
case, we also used a somewhat smaller architecture: Four hidden layers of size
$2047$.

Most MLPs never reach the test results achieved by BM3D because of the
relatively bad performance on image ``Barbara''.  For $\sigma=50$, we approach
the results achieved by BM3D faster than for $\sigma=25$ and for $\sigma=25$,
we approach the results achieved by BM3D faster than for $\sigma=10$. For
$\sigma=75$, we approach the results achieved by BM3D the fastest and even
slightly outperform the results.  We see that the gap between our results and
those of BM3D becomes smaller when the noise is stronger. The slower
convergence for lower noise levels can be explained by the fact that the
overall error is lower (or equivalently: the PSNR values are higher), which
causes the updates during the training procedure to be smaller.

For $\sigma=10$, better results are achieved with smaller patches. For
$\sigma=25$, $\sigma=50$ and $\sigma=75$, better results are achieved with
larger patches.  The reason larger patches achieve better results for
$\sigma=25$, $\sigma=50$ and $\sigma=75$ is that larger patches are necessary
when the noise becomes stronger~\citep{levin2010natural}. This implies that it
is not necessary to use large patches when the noise is weaker. Indeed, using
patches that are too large can cause the optimization to become difficult, see
section~\ref{sec:idealpatchsize}. Therefore, the ideal patch sizes are
influenced by the strength of the noise. We used $21\times21$ and $9\times9$
patches for $\sigma=10$ and $39\times39$ and $17\times17$ patches for the other
noise levels.

\section{Training trade-offs for block-matching MLPs}
We have seen in \citet{burgerjmlr1} that MLPs can be combined with a
block-matching procedure and that doing so can lead to improved results on some
images. In this section, we discuss the training procedure of block-matching
MLPs in more detail. We write $(39,14x13, 4x2047,13)$ to denote a
block-matching MLP with a search window of size $39\times39$ pixels, taking as
input $14$ patches of size $13\times13$ pixels, four hidden layers with $2047$
hidden units each, and an output patch size of $13\times13$ pixels.

\begin{figure}[htbp]
  \centering
    \includegraphics[width=\columnwidth]{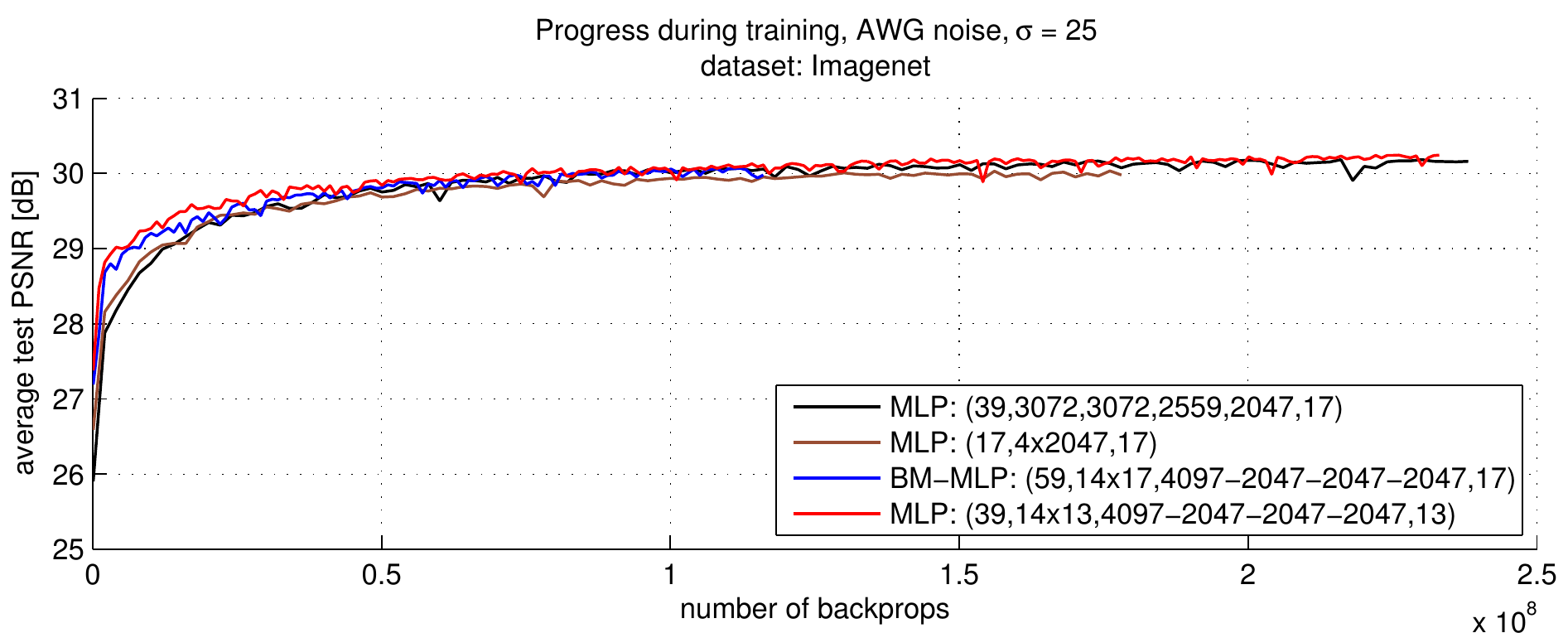}
    \caption{Block matching helps at the beginning of the training procedure.}
  \label{fig:progress007}
\end{figure}

\subsection{Block-matching MLPs can learn faster} 
We see in Figure~\ref{fig:progress007} that progress during training with the
block-matching MLPs is similar to progress with the best MLPs that do not use
block-matching. We see an improvement over the plain MLPs particularly at the
beginning of the training procedure. Later on, the advantage of the
block-matching procedure over plain MLPs is less evident. The block-matching
procedure using patches of size $13\times13$ and a search window of size
$39\times39$ performs slightly better than the block matching procedure using
patches of size $17\times17$ and a search window of size $59\times59$. The
search window size of $39\times39$ is the same as the size of the patches the
best-performing plain MLP takes as input. This means that the block-matching
MLP achieving the better results always uses less information as input than the
plain MLP achieving the best results, yet still achieves similar results.

\begin{figure}[htbp]
  \centering
    \includegraphics[width=\columnwidth]{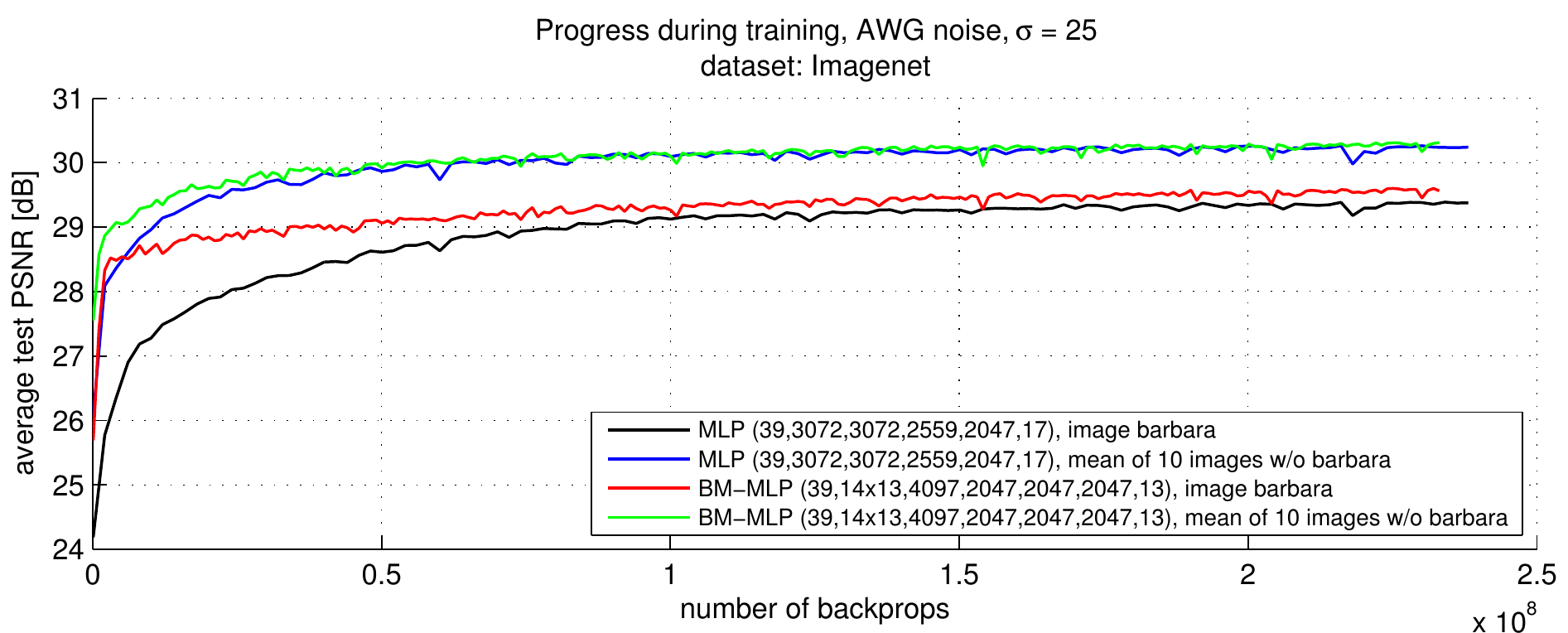}
    \caption{Block matching helps particularly for image Barbara.}
  \label{fig:progress011}
\end{figure}

Figure~\ref{fig:progress011} compares the progress of the winning plain MLP to
the block-matching MLP using patches of size $13\times13$ on image ``Barbara''
against the remaining $10$ of the $11$ standard test images. We see that on
image ``Barbara'', the block-matching MLP has a clear advantage, particularly
at the beginning of the training procedure. On the remaining images, the
advantage is less clear.  Still, the results at the beginning of the training
procedure are better for the block-matching MLP.

This answers our question: The block-matching procedure helps on images with
regular structure. However, the improvement is rather small at the end of the
training procedure.

\subsection{Are block-matching MLPs useful on all noise levels?} 
We train MLPs in combination with the block matching procedure on noise levels
$\sigma=10$, $\sigma=50$ and $\sigma=75$. We again use $k=14$ and patches of
size $13\times13$.

\begin{figure}[htbp]
  \centering
    \includegraphics[width=\columnwidth]{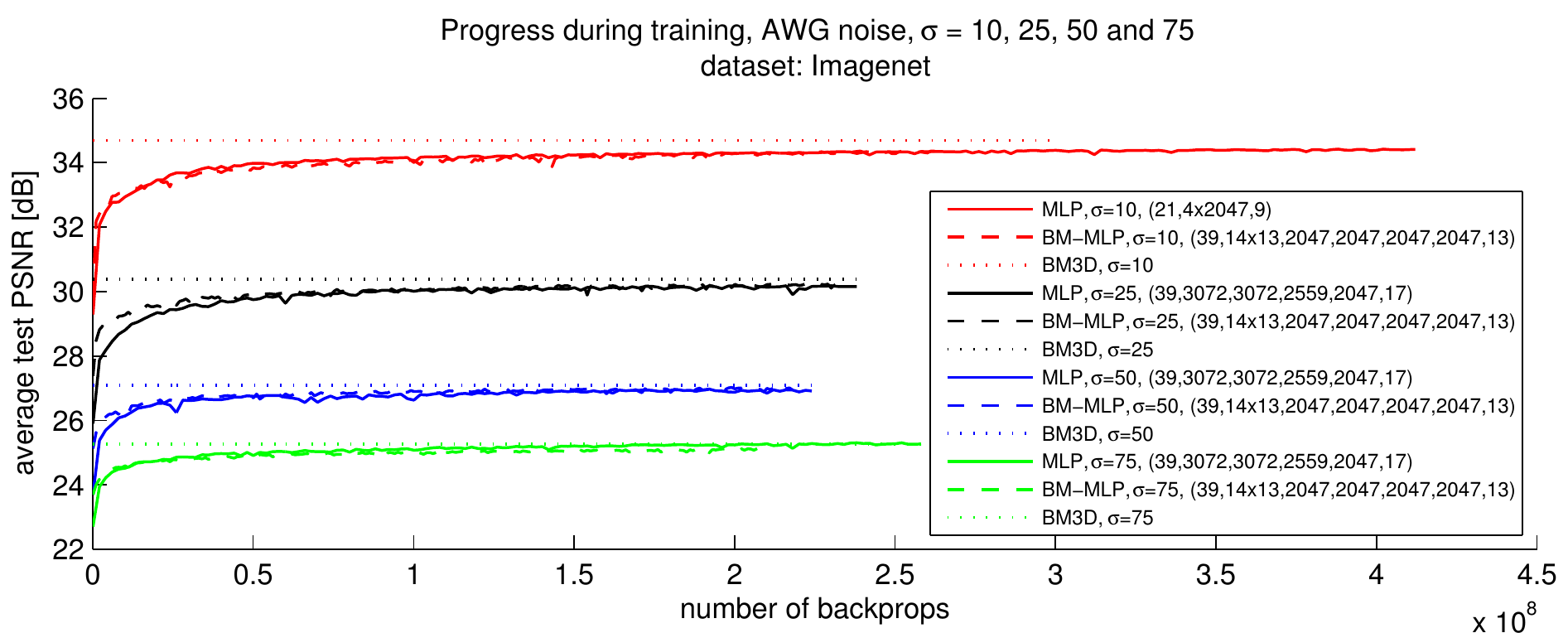}
    \caption{Progress on different noise levels compared to BM3D.
    Block-matching is the most useful at $\sigma=25$ and $\sigma=50$.}
  \label{fig:progress009}
\end{figure}

Figure~\ref{fig:progress009} shows the progress during training for the
different noise levels. For $\sigma=10$, the block-matching procedure seems to
present no advantage over the best MLP without block-matching procedure. For
$\sigma=25$ and $\sigma=50$, the block-matching procedure provides better
results at the beginning of the training procedure. In the later stages of the
training procedure, it is not clear if the block-matching procedure achieves
superior results. For $\sigma=75$, the block-matching procedure presents no
clear advantage at the beginning of the training procedure and also achieves
worse results than the plain MLP in the later stages of the training procedure.
A possible explanation for the deterioration of the results achieved with
block-matching compared to plain MLPs at increasing noise levels is that it
becomes more difficult to find patches similar to the reference patch. A
possible solution would be to employ a coarse pre-filtering step such as the
one employed by BM3D.

\section{Analysis of hidden activation patterns}
We have seen in \citep{burgerjmlr1} that our method can achieve good results on
medium to high noise levels. We have also shown which steps are important and
which are to be avoided in order to achieve good results.  We now ask the
question: Can we gain insight into how the MLP works?  An MLP is a highly
non-linear function with millions of parameters. It is therefore unlikely that
we will be able to perfectly describe its behavior.  This section describes a
set of experiments that will nonetheless provide some insight about how the MLP
works. 

\paragraph{Definitions:} 
Weights connecting the input to one unit in the first hidden layer can be
represented as a patch. We refer to these weights as \emph{feature detectors}
because they can be interpreted as filters.  The weights connecting one unit in
the last hidden layer of an MLP to the output can also be represented as a
patch and we will refer to these as \emph{feature generators}. 

When feeding data into an MLP, we are interested not only in the weights, but
also in the \emph{activations}, by which we mean the values taken by the hidden
units, due to the input.  We will attempt to find inputs maximizing the
activation of a specific hidden unit and refer to such an input as an
\emph{input pattern}. Conversely, we refer to the output caused by the
activation of a single hidden unit as an \emph{output pattern}. 

The input pattern maximizing the
activation of a hidden unit in the first hidden layer is the same as the feature detector
corresponding to the hidden unit. Also, the output pattern corresponding to a
hidden unit in the last hidden layer is the same as the feature generator associated to the same hidden
unit.

\subsection{MLPs with a single hidden layer}
We start by analyzing an MLP with a single hidden layer. We use an MLP with the
architecture ($17\times17,2047,17\times17$) for that purpose. Such an MLP is identical
to a denoising auto-encoder with AWG noise~\citep{vincent2010stacked}.

\paragraph{Weights as patches:} 
The feature detectors of this MLP can be represented as patches of size
$17\times17$ pixels. The feature generators have the same size of the feature detectors.

\begin{figure}[htbp]
  \centering
  \rotatebox{90}{\tiny input} 
  \includegraphics[width=0.95\columnwidth]{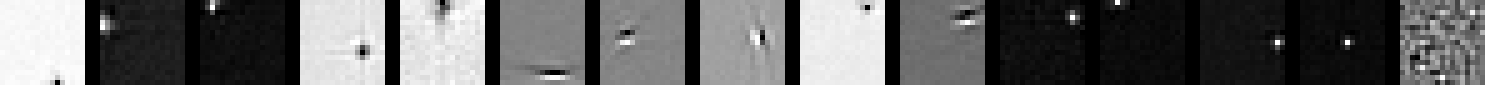} \\
  \rotatebox{90}{\tiny output} 
  \includegraphics[width=0.95\columnwidth]{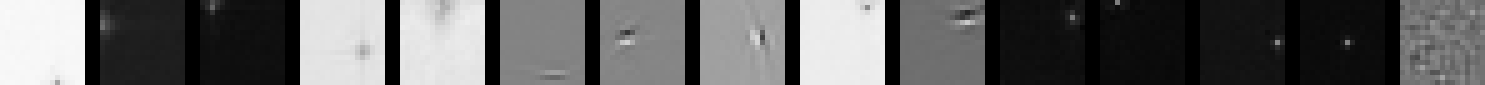} 
  \caption{Random feature detectors (top) and the corresponding feature
  generators (bottom) in a trained MLP with one hidden layer.}
\label{fig:patchespaired}
\end{figure}
Figure~\ref{fig:patchespaired} shows some feature detectors (top row) and the
feature generators corresponding to each feature detector (bottom row). Scaling
of the pixel values was performed separately for each pair of feature detector
and feature generator. The feature detectors are similar in appearance to the
corresponding feature generators, up to a scaling factor. The feature detectors
can be classified into three main categories: 1) feature detectors resembling
Gabor filters 2) feature detectors that focus on just a small number of pixels
(resembling a dot), and 3) feature detectors that look noisy. Most feature
detectors belong to the first and second category. The Gabor filters occur at
different scales, shifts and orientations. Similar dictionaries have also been
learned by other denoising approaches. It should be noted that MLPs are not
shift-invariant, which explains why some patches are shifted versions of each
other.  Similar features have been observed in denoising
auto-encoders~\citep{vincent2010stacked}. 

\begin{figure}[htbp]
  \centering
  \rotatebox{90}{\small \hspace{30pt} feature detectors}
  \includegraphics[width=0.95\columnwidth]{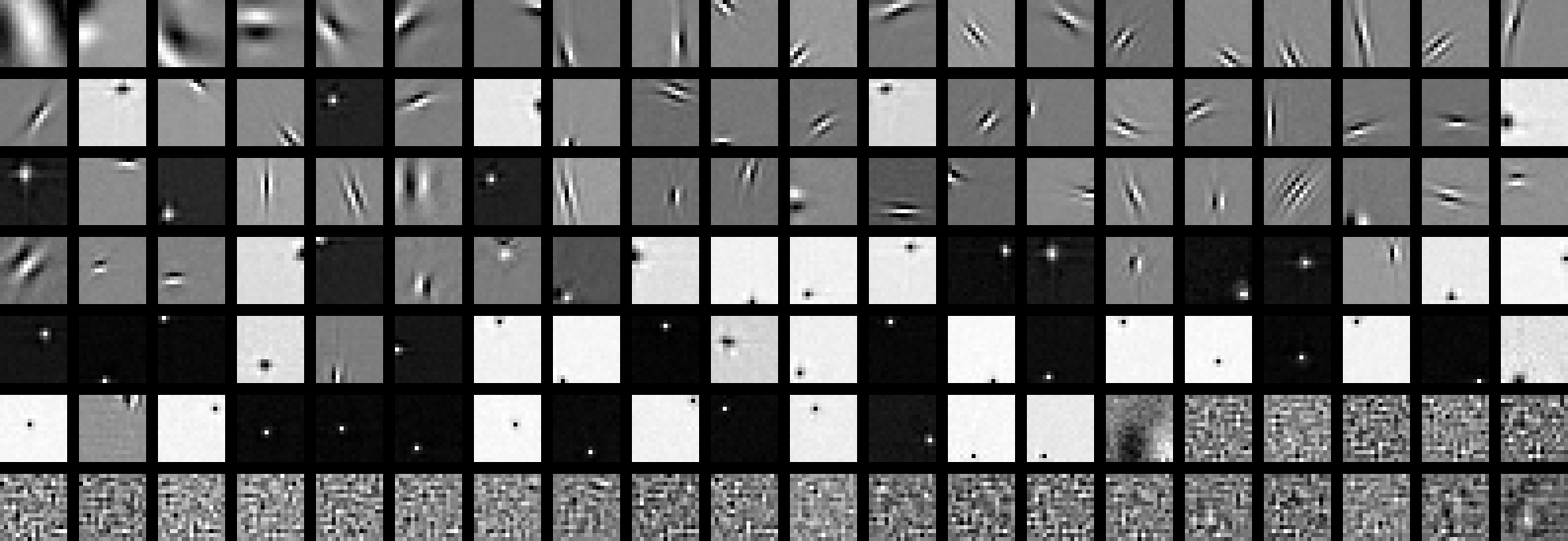}
  \caption{Selection of feature detectors in an MLP with a single hidden layer
  sorted according to their standard deviation. We chose every $15$th feature
  detector in the sorted list.  The sorting is from left to right and from top
  to bottom: The top-left patch has the highest standard deviation, the
  bottom-left patch the lowest.}
\label{fig:patchessortedstd}
\end{figure}
In Figure~\ref{fig:patchessortedstd}, the feature detectors have been sorted
according to their standard deviation. We see that the feature detectors that
look noisy have the lowest standard deviation. The noisy feature detectors
therefore merely look noisy because of the normalization according to which
they are displayed. Because the noisy-looking feature detectors have different
mean values, we can interpret them as various DC-component detectors.

Denoising auto-encoders are sometimes trained with ``tied'' weights: The
feature detectors are forced to be identical to the output bases. We observe
that the learned feature detectors and feature generators look identical up to
a scaling factor without the tying of the weights. This suggests that the
intuition behind weight tying is reasonable. However, our observation also
suggests that better results might be achieved if the feature detectors and
feature generators are tied, but allowed to have different scales.

\paragraph{Activations:} The MLP learned a dictionary in the output layer
resembling the dictionaries learned by sparse coding methods, such as KSVD.
This suggests that the activations in the last hidden layer might be sparse.
We therefore ask the question: What is the behavior of the activations in the
hidden layer?

\begin{figure}[htbp]
  \centering
  \begin{tabular}{cc}
    \includegraphics[width=0.45\textwidth]{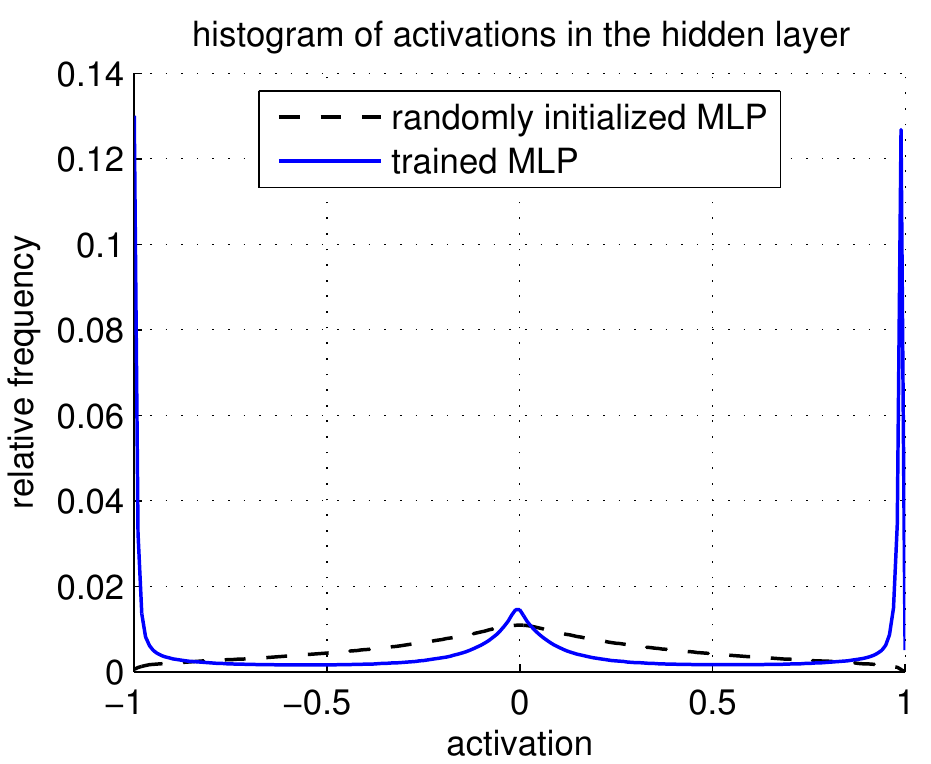} &
    \includegraphics[width=0.45\textwidth]{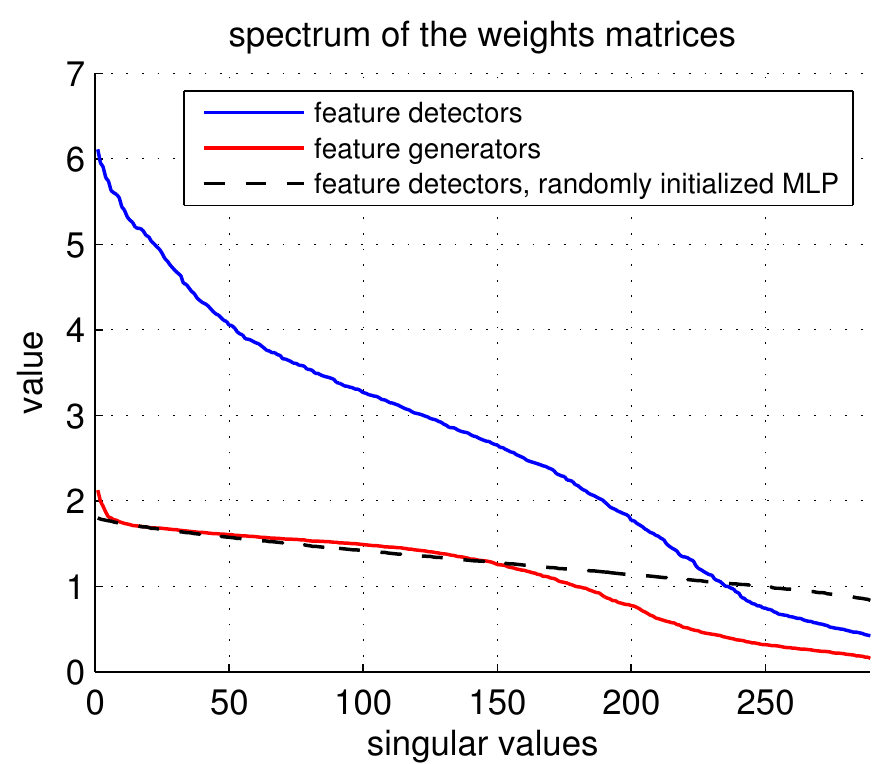}\\
    (a) & (b)
  \end{tabular}
  \caption{(a) Histograms of the activations in the hidden layer of a one
  hidden layer MLP. (b) Spectrum of the feature detectors and feature generators.}
  \label{fig:histogram003}
\end{figure}
\begin{figure}[t]
  \centering
  \begin{tabular}{cc}
    \includegraphics[width=0.45\textwidth]{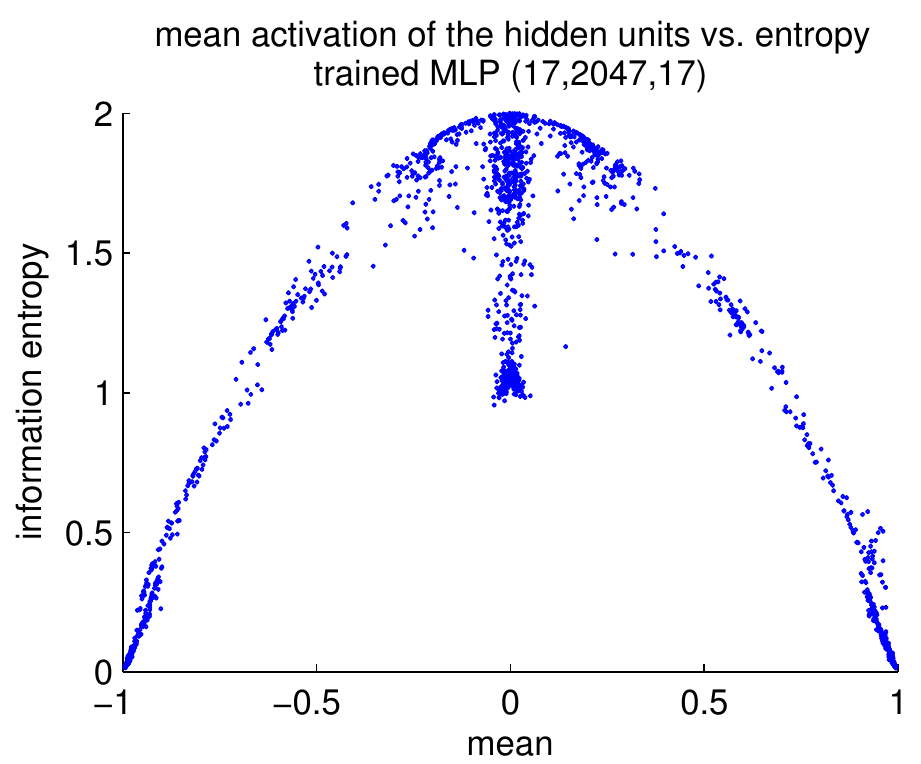} &
    \includegraphics[width=0.45\textwidth]{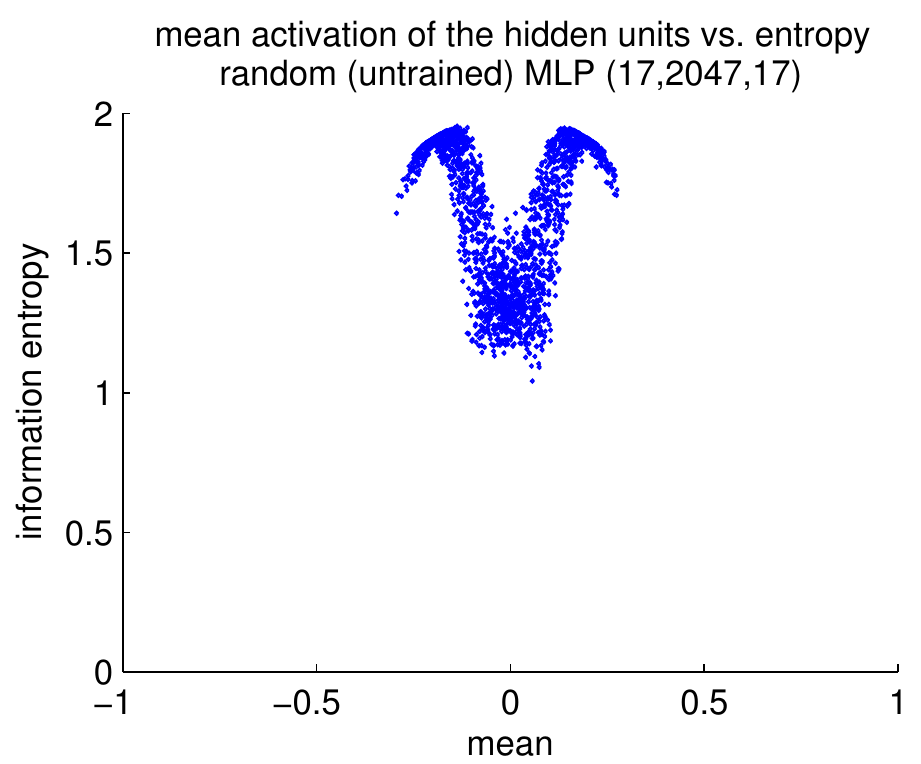} \\
    (a) & (b)
  \end{tabular}
    \caption{(a) In a trained MLP, units with small mean tend to have high
    entropy. This means that these units are highly active; only their mean
    activation is close to $0$. (b) In an untrained MLP, no units have high mean.}
  \label{fig:meanactivations001}
\end{figure}

Figure~\ref{fig:histogram003}a shows a histogram of the activations of all
hidden units in both a trained MLP and a random MLP, evaluated on the $500$
images in the Berkeley dataset. The activations are centered around zero in the
case of the random MLP. The activations in the trained MLP however are almost
completely binary: The activations are either close to $-1$ or close to $1$,
but seldom in between. This is an indication that the training process is
completed: The activities lie on the saturated parts of the $\tanh$ transfer
function, where the derivative is close to zero. The gradient that is
back-propagated to the first layer is therefore mostly zero.  This also answers
our question: The activations are not sparse. We will provide a further
interpretation for this observation later in this section.

\paragraph{Entropy:} 
To measure the usefulness of neurons, we estimate the information entropy of
their activation distributions. We plot the mean activations of hidden units
against their entropy $H(X) = -\sum_{i=1}^{N} p(x_{i})\log_{2} p(x_{i})$ with
four bins of equal size in Figure~\ref{fig:meanactivations001}a.  We repeat the
experiment for an untrained MLP in Figure~\ref{fig:meanactivations001}b.  We
see that units with high entropy tend to have a low absolute mean, and that
units with low absolute mean have high entropy. The reverse is also true: units
with low entropy have a high absolute mean and units with a high absolute mean
have low entropy. The entropy of the units in a random MLP is higher than in a
trained MLP. This is explained by the fact that the random MLP has no units
with high absolute mean.  These observations allow us to conclude that the
units that have a mean close to $0$ also have a binary behavior: They are
either $1$ or $-1$ and seldom have a value in between. In fact, we can say that
these units take value $1$ in approximately $50\%$ of the cases and value $-1$
in the remaining cases. They therefore have a high entropy. 

\begin{figure}[htbp]
  \centering
    \rotatebox{90}{\tiny feature detectors}
    \includegraphics[width=0.95\columnwidth]{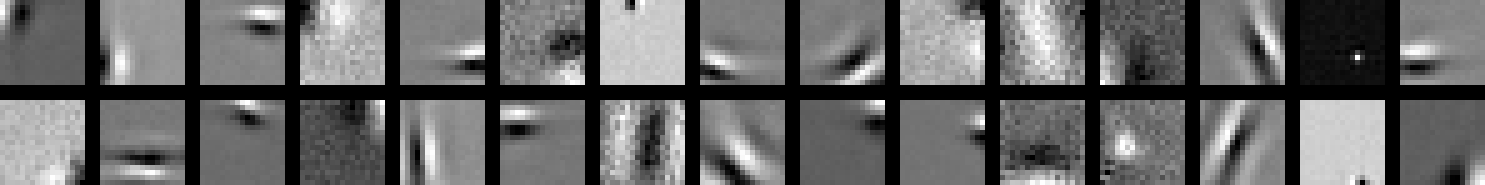}
      
    \vspace{5pt} 
     
    \rotatebox{90}{\tiny feature detectors}
    \includegraphics[width=0.95\columnwidth]{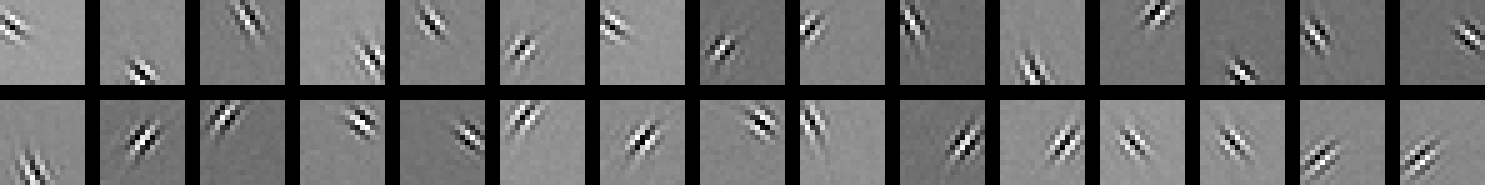}
    \caption{Feature detectors of the units with the highest (top) and lowest (bottom) entropy.}
  \label{fig:entropy_patches}
\end{figure}

Figure~\ref{fig:entropy_patches} shows the feature detectors of the units with
the highest and lowest entropy. The feature detectors with the lowest entropy
all resemble high-frequency Gabor filters of different positions and
orientations. A possible explanation for their low entropy is that these
filters are highly selective. Only few patches cause these filters to
activate.

\paragraph{Weight spectrum:} We perform an SVD-decomposition of the weight
matrices of both the trained and the random MLP and plot the spectrum of the
singular values, see Figure~\ref{fig:histogram003}b. For the random MLP, we
omit the spectrum of the feature generators because its shape is identical to the
spectrum of the feature detectors. This is due to the initialization procedure
and symmetrical architecture.

The similar shape of the spectra in the trained MLP was expected: the feature
detectors and feature generators are similar in appearance, see
Figure~\ref{fig:patchespaired}. The larger singular values for the feature
detectors is a reflection of the fact that the norms of the feature detectors
is larger than the norm of the feature generators (also seen in
Figure~\ref{fig:patchespaired}). 

The spectrum for both the feature detectors and the feature generators is
relatively flat, which is an indication that the feature detectors are diverse:
Strong correlations between feature detectors would cause small singular
values. The fact that there are no singular values with value zero means that
the output bases matrix has full rank. The spectrum of the random MLP is even
flatter: it also has full rank. This means that the output bases of both the
trained and the random MLP are able to approximate any patch.

\begin{figure}[htbp]
  \centering
  \begin{tabular}{cc}
    \includegraphics[width=0.45\textwidth]{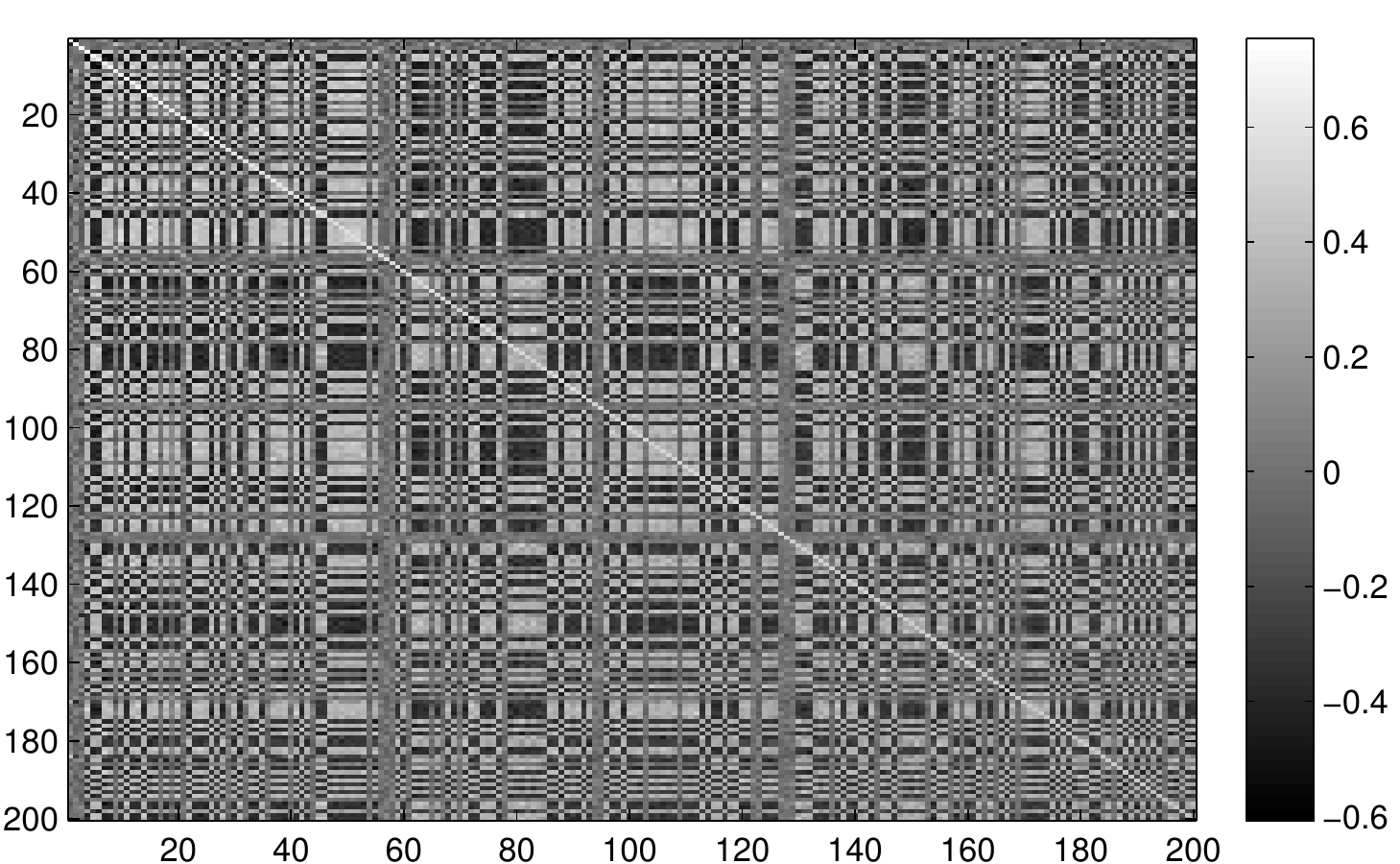} &
    \includegraphics[width=0.45\textwidth]{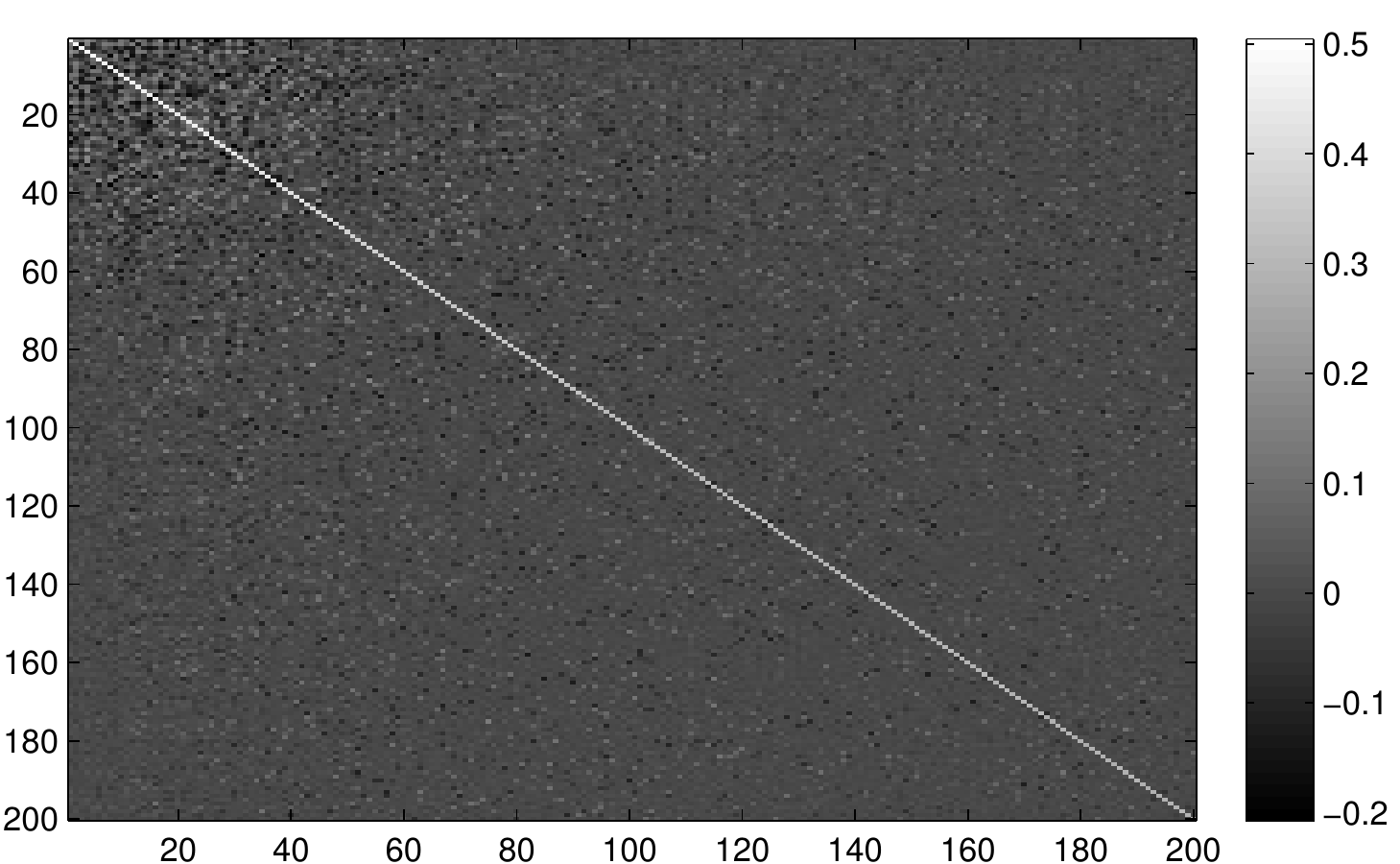} \\
    (a) & (b)
  \end{tabular}
  \caption{(a) Many hidden units are strongly correlated when image data is
  given as input. (b) The hidden units are not strongly correlated when noise
  is given as input.}
  \label{fig:activationcorrelations}
\end{figure}

\paragraph{Activation correlations:}
Figure~\ref{fig:activationcorrelations}a shows the covariance matrix between
the $200$ hidden units of the trained MLP with the highest variance, when image
data is provided as input. We see that activations between units are highly
correlated. This is a reflection of the fact that many of the features detected
by the filters tend to occur simultaneously in image patches.
Figure~\ref{fig:activationcorrelations}b shows that this observation does
not hold when noise is provided as input.

\begin{figure}[htbp]
  \centering
  \begin{tabular}{cc}
    \includegraphics[width=0.45\textwidth]{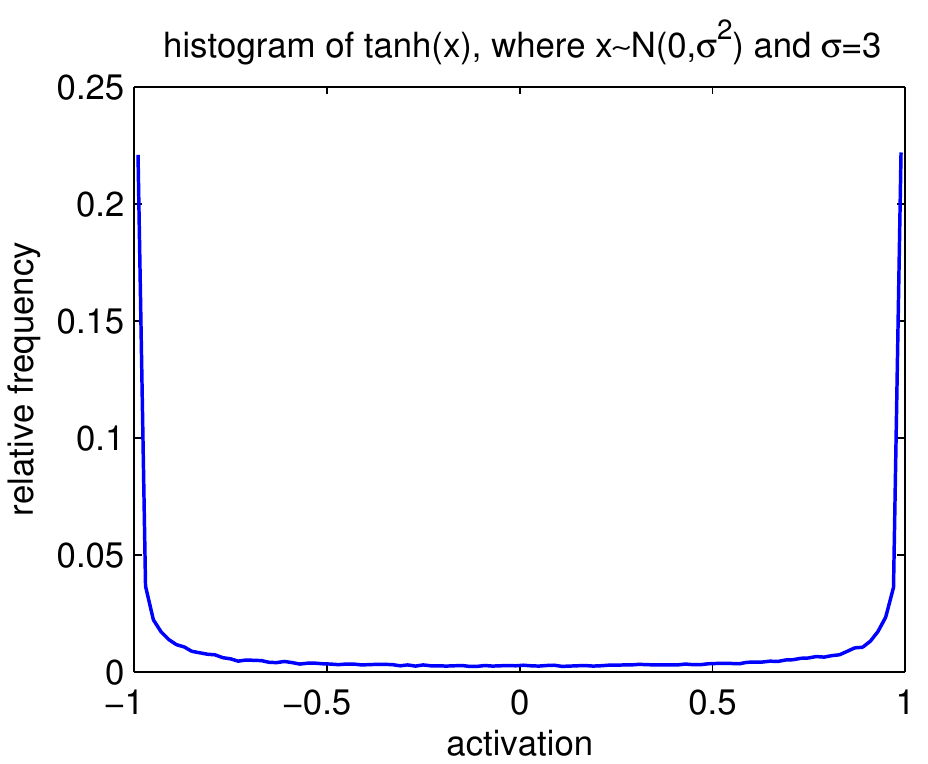} &
    \includegraphics[width=0.45\columnwidth]{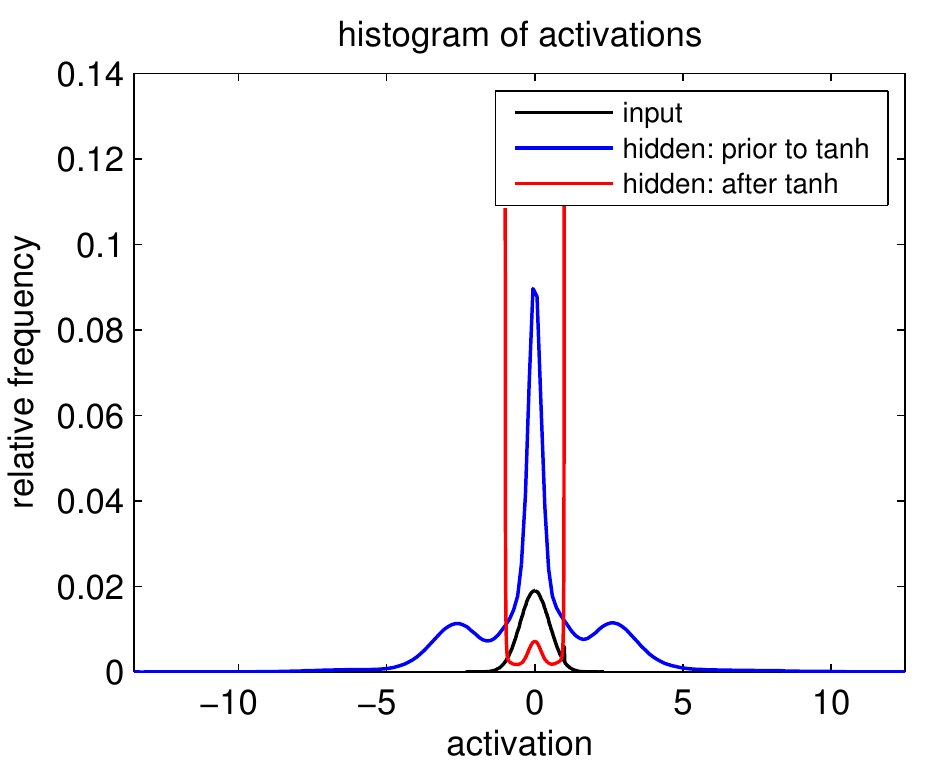} \\
    (a) & (b)
  \end{tabular}
    \caption{(a) Applying a squashing function on a normally distributed vector
    with high variance creates a binary distribution. (b) AWG noise with
    $\sigma=25$ will cause mostly binary activations in the hidden layer.}
  \label{fig:binarization001}
\end{figure}

\paragraph{How do the binary codes arise?} We observed in
Figure~\ref{fig:histogram003}a that the codes in the hidden layer are almost
completely binary. This observation is surprising: The binary distribution was
not explicitly enforced and the distribution of activations is usually 
different~\citep{GlorotAISTATS2010}. A possible explanation would be if the activities prior to
the application of the $\tanh$-function have high variance.  Applying the
$\tanh$-function on a normally-distributed vector with high variance indeed
creates a binary distribution, see Figure~\ref{fig:binarization001}a.

Is this explanation plausible? A supporting argument is that the feature
detectors shown in Figure~\ref{fig:patchespaired} have high norm compared to
their corresponding output bases. The high norm of the filters could cause high
activations in the hidden layers.

We now feed AWG noise with $\sigma=25$ into the MLP. The histograms of the
activations prior and after application of the $\tanh$-layer are shown in
Figure~\ref{fig:binarization001}b. We observe that the activations before the
$\tanh$-layer indeed have high variance and that the activations after the
$\tanh$-layer are indeed mostly binary. We conclude that the binary activities
in the hidden layer are due to activities with high variance prior to the
$\tanh$-layer, which are in turn due to feature detectors with high norm.

\begin{figure}[htbp]
  \centering
    \includegraphics[width=\columnwidth]{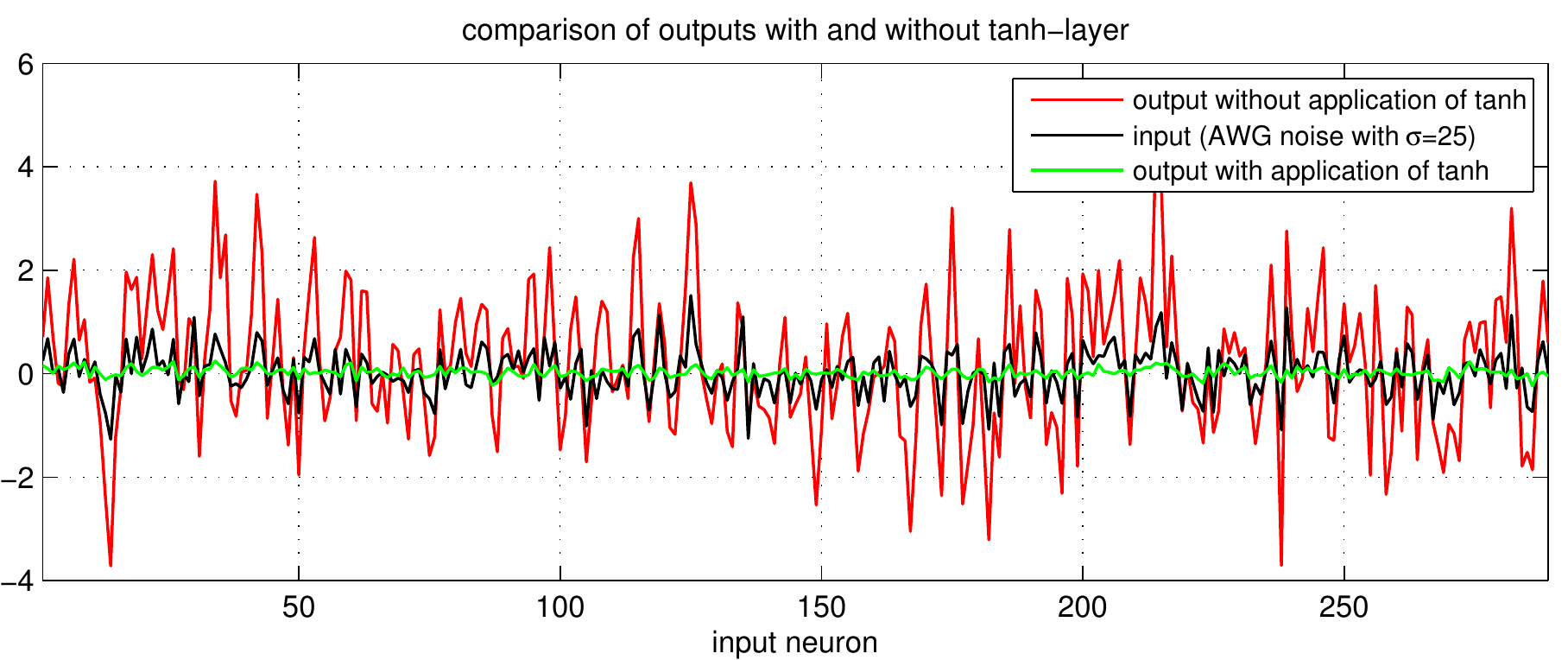}
    \caption{The $\tanh$-layer has a denoising effect due to saturation.}
  \label{fig:binarization003}
\end{figure}

\begin{figure}[htbp]
  \centering
    \includegraphics[width=\columnwidth]{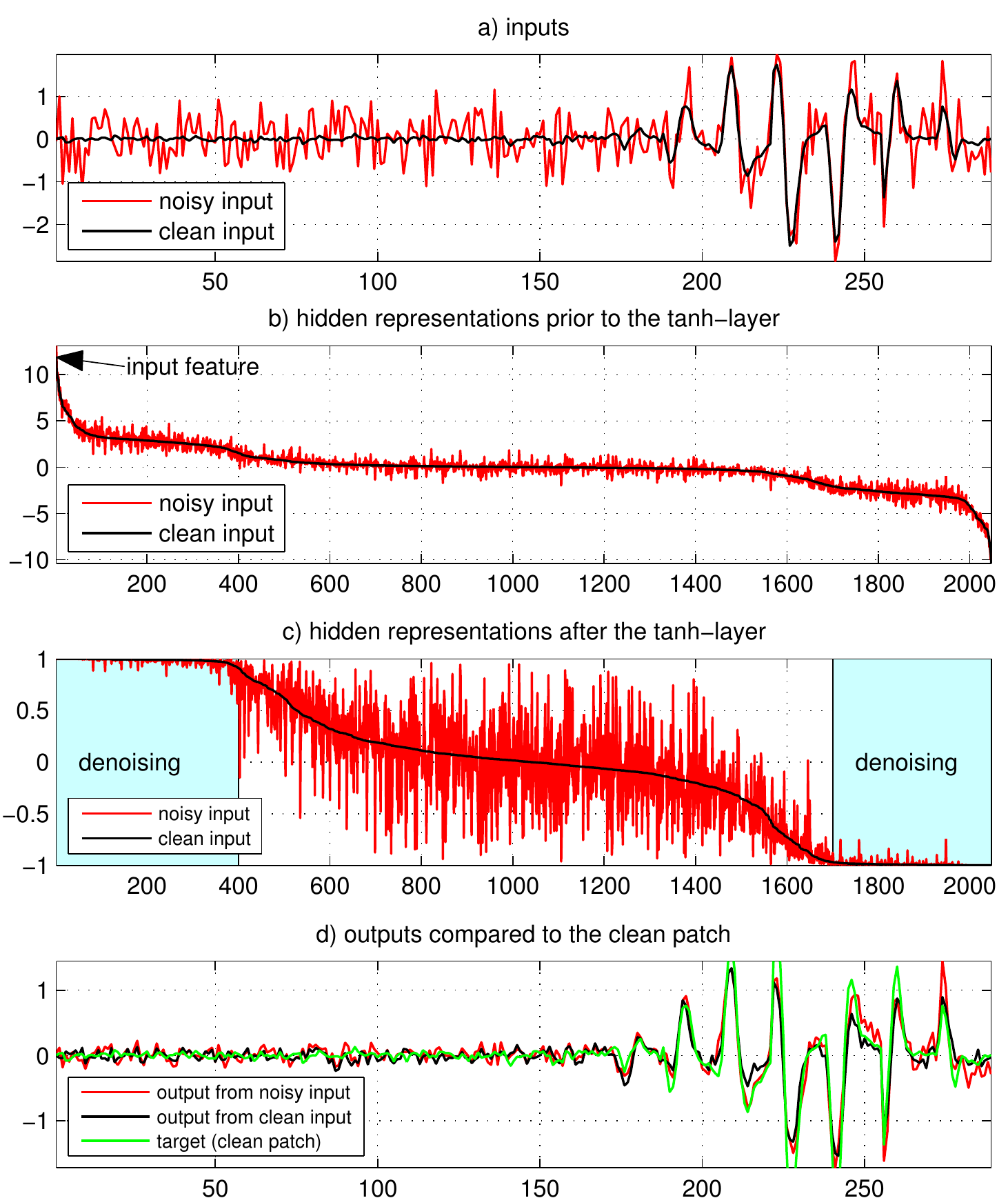}
    \caption{Denoising in action: The input maximizes the activity of one
    feature detector (hidden neuron). Other feature detectors are also strongly
    activated. After the $\tanh$-layer, the noise has almost no effect on the
    feature detectors that are highly active. The activations in b) are sorted and
    the activations in c) use the same sorting. Denoising happens mostly in the
    blue areas in c).}
  \label{fig:binarization004}
\end{figure}

\begin{figure}[htbp]
  \centering
    \includegraphics[width=\columnwidth]{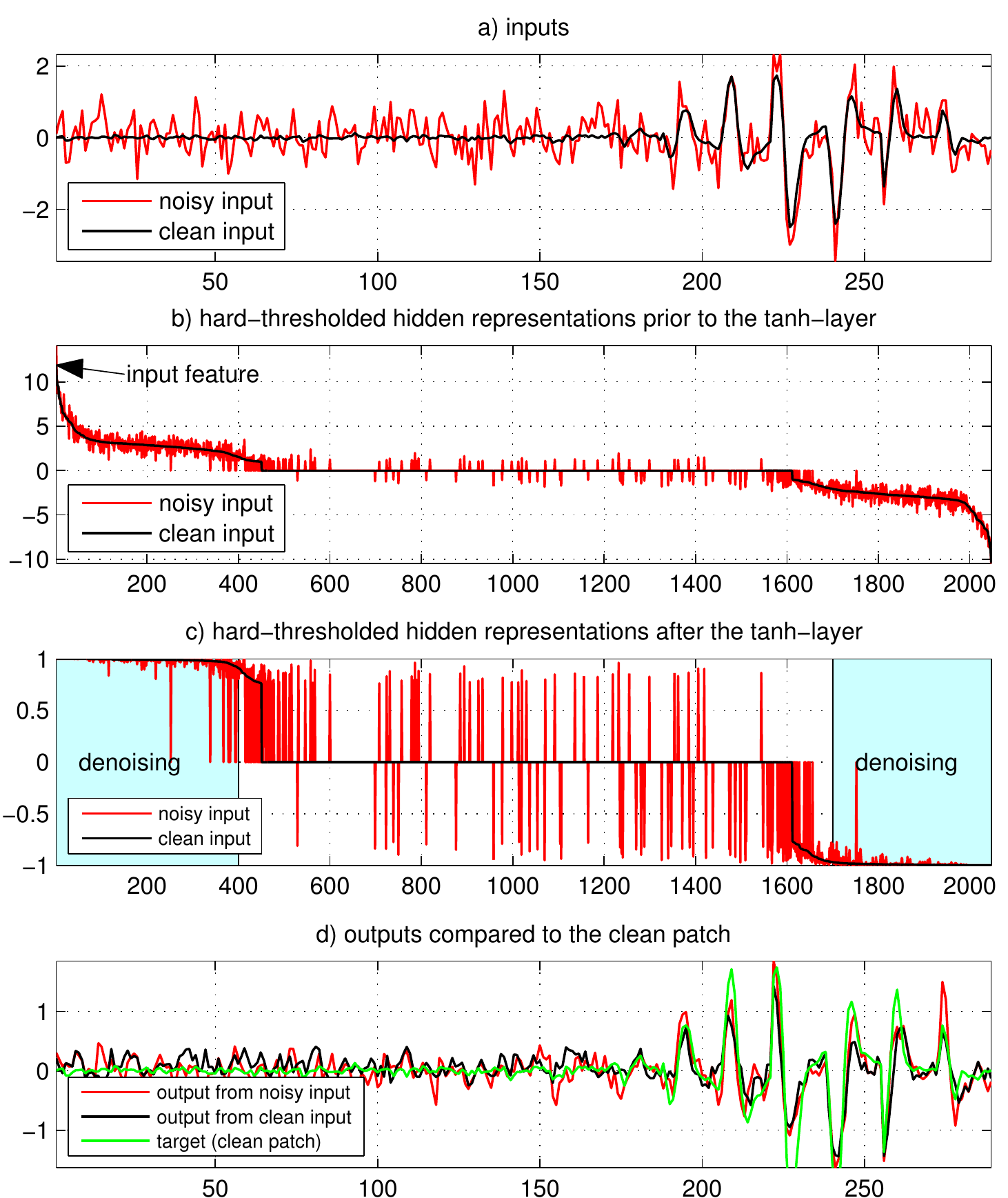}
    \caption{Denoising through hard-thresholding: Setting the less import
    feature detectors to $0$ also produces a denoising effect. The activations in b)
    are sorted and the activations in c) use the same sorting.}
  \label{fig:binarization005}
\end{figure}

\paragraph{How is denoising achieved?} We have made a number of
observations regarding the behavior of the MLP but have not yet explained why
the MLP is able to denoise. Is the binarization effect observed in
Figure~\ref{fig:binarization001} an important factor? To answer this question,
we feed an image patch containing only AWG noise with $\sigma=25$ through the
MLP. We compare the output when the $\tanh$-layer is applied to when the
$\tanh$-layer is not applied, see Figure~\ref{fig:binarization003}. Without
$\tanh$-layer, the output is more noisy than the input. With $\tanh$-layer
however, the output is less noisy than the input. We can therefore conclude
that the same thresholding operation responsible for the binary codes is also
at least partially responsible for the denoising effect of the MLP.

Thresholding for denoising has been thoroughly studied and dates back at least
to ``coring'' for reducing television noise \citep{rossi1978digital}. Typically,
a thresholding operation is performed in some transform domain, such as a
wavelet domain \citep{portilla2003image}. However, the thresholding operations
typically affect small values most strongly: In the case of \emph{hard
thresholding}, values close to zero are set to zero and all other values are left
unchanged. In the case of \emph{soft thresholding}, all values are reduced by a
fixed amount. Then, values close to zero are set to zero. In the MLP, the
situation is reversed: Values close to zero are left unchanged. Only large
absolute values are modified by the $\tanh$-layer. We call this effect
\emph{saturation}.

We have seen that the saturation of the $\tanh$-layer can explain why noise is
reduced. However, denoising can always be trivially achieved by removing both
noise and image information.  We therefore ask the question: Why are image
features preserved? We proceed by example. As input, we will use the feature
detected by one of the feature detectors. As a comparison, we will use as input
a noisy version of this feature, see Figure~\ref{fig:binarization004}a. The
clean input has the effect of maximizing the activity of its corresponding
feature detector prior to the $\tanh$-layer, see
Figure~\ref{fig:binarization004}b. Other feature detectors also have a high
value, which should be expected, given the high covariance of the hidden units,
see Figure~\ref{fig:activationcorrelations}.  We see that the noisy input
creates a hidden representation that looks quite different from the one created
from the clean input: The noise is still clearly present. After application of
the $\tanh$-layer, the noise is almost completely eliminated on the feature
detectors with high activity, see Figure~\ref{fig:binarization004}c. This is
due to the saturation of the $\tanh$-layer. The outputs look similar to the
clean input, see Figure~\ref{fig:binarization004}d. In particular, the noise
from the noisy input has been attenuated.

We repeat the experiment performed in Figure~\ref{fig:binarization004}, but
this time hard-threshold the hidden activities: Activities in the hidden layer
prior to the $\tanh$-layer with an absolute value smaller than $1$ are set to
$0$. Doing so still produces a denoising effect, see
Figure~\ref{fig:binarization005}. This observation brings us to the conclusion
that the feature detectors with a high activity are the more important ones.
This is convenient, because the noise on the feature detectors with high
activities disappears due to saturation.

We summarize the denoising process in a one-hidden-layer MLP as follows. Noise
is attenuated through the saturation of the $\tanh$-layer. Image features are
preserved due to the high activation values of the corresponding feature
detectors.

\paragraph{Relation to stacked denoising autoencoders (SDAEs):} MLPs with a single
hidden-layer which are trained on the denoising task are exactly equivalent to
denoising autoencoders. Denoising autoencoders can be stacked into
SDAEs~\citep{vincent2010stacked}. The difference between SDAEs and MLPs with
multiple hidden layers is that SDAEs are trained sequentially: One layer is
trained at a time and each layer is trained to denoise the output provided by
the previous layer (or the input data in the case of the first layer). While
our MLPs are trained to optimize denoising performance, SDAEs are trained to
provide a useful initialization for a different supervised task.

It has been suggested by \citet{bengio2007greedy} that deep learning is useful due
to an \emph{optimization} effect: Greedy layer-wise training helps to optimize
the training criterion. However, later work contradicts this interpretation:
\citet{erhan2010does} suggest that SDAEs and other deep pre-trained
architectures such as deep belief nets (DBNs) are useful due to a \emph{regularization} effect:
Supervised training of an architecture (especially a deep one) using stochastic
gradient descent is difficult because of an abundance of local minima, many of
them poor (in the sense that they do not generalize well). The unsupervised
pre-training phase imposes a restriction on the regions of parameter space that
stochastic gradient descent can explore during the supervised phase and reduces
the number of local minima that stochastic gradient descent can fall into.
Pre-training thus initializes the architecture in such a way that stochastic
gradient descent finds a better basin of attraction (again in the sense of
generalization). 

The fact that activations in the hidden layers of a SDAE are almost completely
binary (see Figure~\ref{fig:histogram003}) and relatively high entropy (see
Figure~\ref{fig:meanactivations001}a) was not mentioned
by \citet{erhan2010does}, but is in agreement with the regularization
interpretation: The fact that the denoising task forces the hidden
representations to be binary is a restriction and therefore also a form of
regularization. In addition, information about the input should be preserved in
order for the hidden representations to be useful. Information about the input
is preserved by virtue of the denoising task: The hidden representations have
to contain sufficient information to reconstruct the uncorrupted input. The
fact that the hidden units have relatively high information entropy is an
additional indication that information is preserved. 

We have not answered the question if the binary restriction is better than a
more classical form of regularization, such as $\ell_{1}$ or $\ell_{2}$
regularization. However, \citet{erhan2010does} suggest that
pre-training achieves a form of regularization that is different from and
indeed more useful than $\ell_{1}$ or $\ell_{2}$ regularization on the
parameters ($\ell_{2}$ regularization on the weights is approximately
equivalent to $\ell_{2}$ regularization on the activations). Another argument is
that binary vectors are easier to manipulate (e.g. classify) than vectors with
small norm.

\paragraph{Relation to restricted Boltzmann machines and deep belief nets:} The
binary activations in the hidden layer of our MLP are reminiscent of restricted
boltzmann machines (RBMs) and deep belief nets (DBNs), which usually employ
stochastic binary activations during the unsupervised training phase
\citep{hinton2006fast}.  An additional similarity is that it has been shown
that DBNs and stacked denoising autoencoders extract similar features when
trained on either hand-written digits or natural image patches
\citep{erhan2010understanding}.

\begin{figure}[htbp]
  \centering
    \rotatebox{90}{\tiny feature detectors in an RBM}
    \includegraphics[width=0.95\columnwidth]{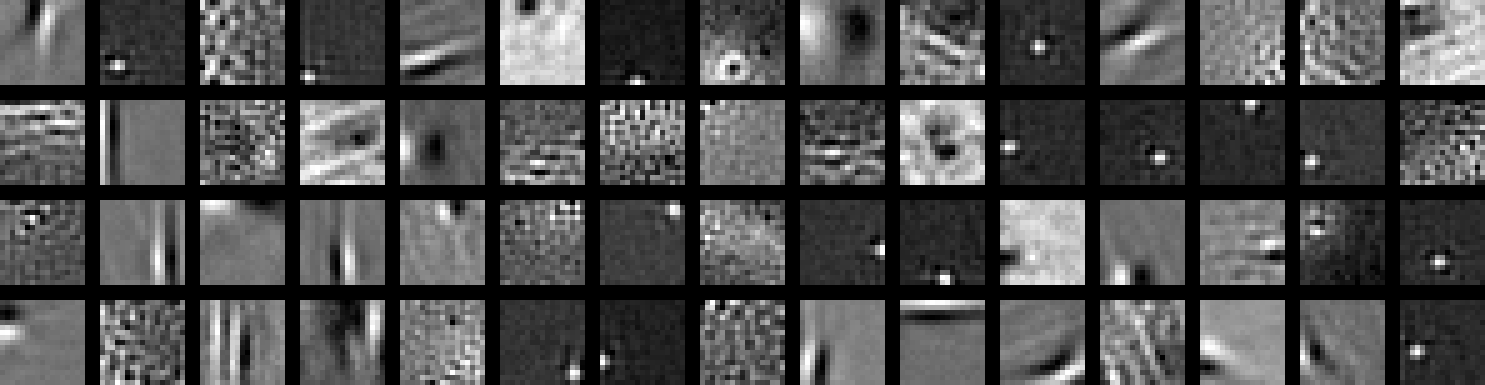}
    \caption{Some filters learned by an RBM trained on natural image data (in an
    unsupervised fashion) with Gaussian visible units, patches of size
    $17\times17$ and $512$ stochastic binary hidden units. The filters resemble
    those learned by an MLP on the denoising task.}
  \label{fig:filters027}
\end{figure}

We trained an RBM with Gaussian visible units on image patches of size
$17\times17$ using contrastive
divergence~\citep{hinton2002training,hinton2010practical}.
Figure~\ref{fig:filters027} shows that the filters learned by the RBM are 
similar in appearance to the filters learned by our one-hidden layer MLP, which
is in agreement with the findings of \citet{erhan2010understanding}. 

\begin{figure}[htbp]
  \centering
  \begin{tabular}{cc}
    \includegraphics[width=0.45\columnwidth]{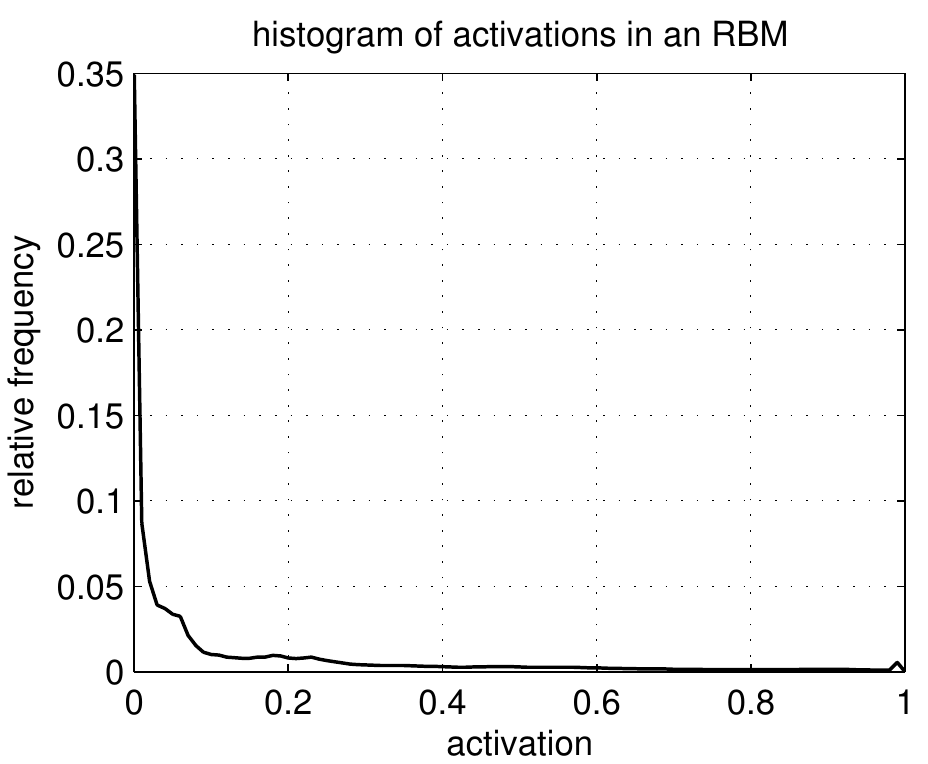} &
    \includegraphics[width=0.45\columnwidth]{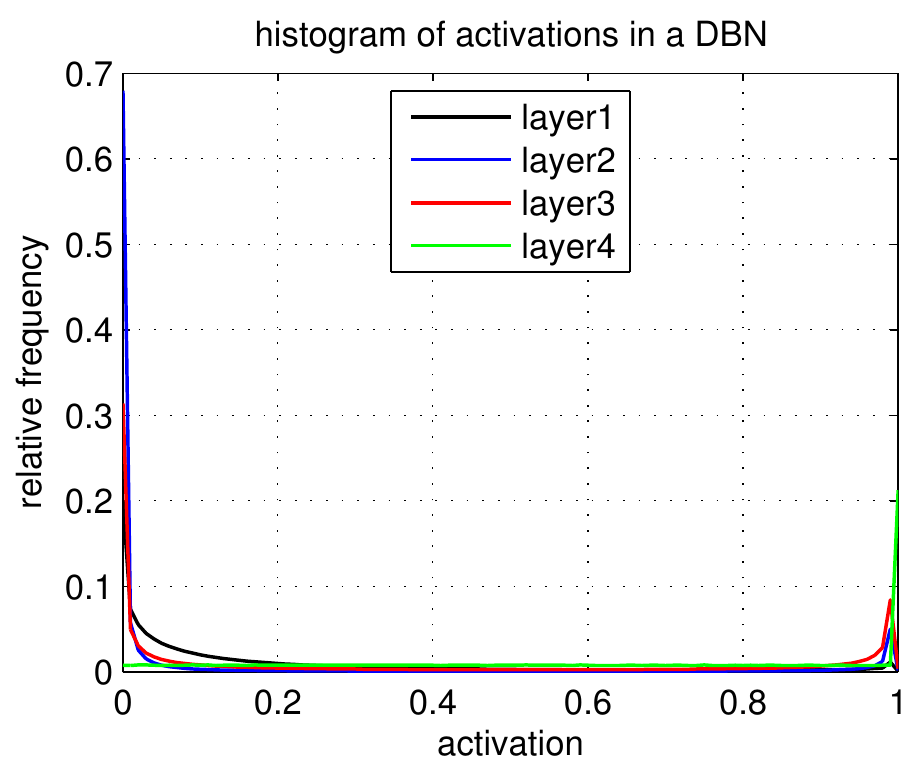} \\
    (a) & (b)
  \end{tabular}
  \caption{(a) Histogram of activations in the hidden layer of an RBM trained
  in an unsupervised fashion on natural image patches.  (b) Histogram of
  activations in the hidden layers of a DBN trained in an unsupervised fashion
  on handwritten digits~\citep{hinton2006reducing}.}
  \label{fig:historbm}
\end{figure}

The activations of the RBM are binary and stochastic during the unsupervised
pre-training phase. It is possible to use the weights learned during
pre-training for a supervised task, in which case the hidden units are allowed
to take real values. After unsupervised learning of our RBM, we observe the
distribution of the real-valued activations in the hidden layer, see
Figure~\ref{fig:historbm}a. The activations lie between $0$ and $1$ instead
of between $-1$ and $1$ for our MLP because of the use of the logistic function
instead of $\tanh$. We see that the activations are sparse and do not show the
binary behavior exhibited by our MLP.

We also used the code provided by \citet{hinton2006reducing} to train a deep
belief net (DBN) on hand-written digits. After pre-training, the activation in
all layers is also sparse, see Figure~\ref{fig:historbm}b. We see that
sparsity occurs in the hidden layers even when not explicitly enforced, as
proposed by \citet{hinton2010practical}.

\paragraph{Summary:} MLPs with one hidden layer denoise by detecting
features in the noisy input patch. Each feature detector responds maximally to
a single feature, but usually many features are detected simultaneously (see
Figure~\ref{fig:activationcorrelations}). The denoised output corresponds to
a weighted sum of each feature detector, see Figure~\ref{fig:patchespaired},
where the weight depends on the response of the feature detector. The features
are mostly Gabor filters of different scales, locations and orientations.
Similar features are observed when training other models on natural image data,
such as RBMs, see Figure~\ref{fig:filters027}. The features are informative in
the sense that many hidden units have high information entropy, see
Figure~\ref{fig:meanactivations001}b. Noise is removed through
saturation of the $\tanh$-layer. Saturation is achieved through feature
detectors with high norm, which in turn leads to activations with high variance
in the hidden layer before the $\tanh$-layer and mostly binary activations after
the $\tanh$-layer, see Figure~\ref{fig:binarization001}. The binary
distribution of activations is surprising given the fact that it has not been
explicitly enforced, but is useful for denoising and also fits well into
the regularization interpretation of denoising auto-encoders proposed
by \citet{erhan2010does}.

\subsection{MLPs with several hidden layers}
The behavior of MLPs with a single hidden layer is easily interpretable.
However, we have seen in Section~\ref{sec:largerarchitectures} that MLPs with
more hidden layers achieve better results. Unfortunately, interpreting the
behavior of an MLP with more hidden layers is more complicated. The weights in
the input layer and in the output layer can still be represented as image
patches, but the layer or layers between the input and output are not so easy
to interpret.  MLPs with a single hidden layer are identical to denoising
autoencoders. This is not true anymore for MLPs with more hidden layers.

\begin{figure}[htbp]
  \centering
  \rotatebox{90}{\small feature detectors}
  \includegraphics[width=0.95\columnwidth]{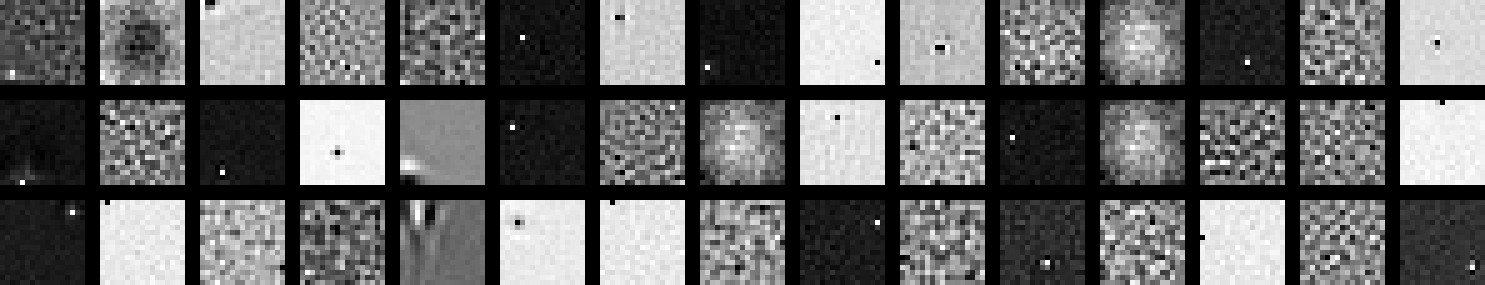}

  \vspace{5pt}
  
  \rotatebox{90}{\small feature generators}
  \includegraphics[width=0.95\columnwidth]{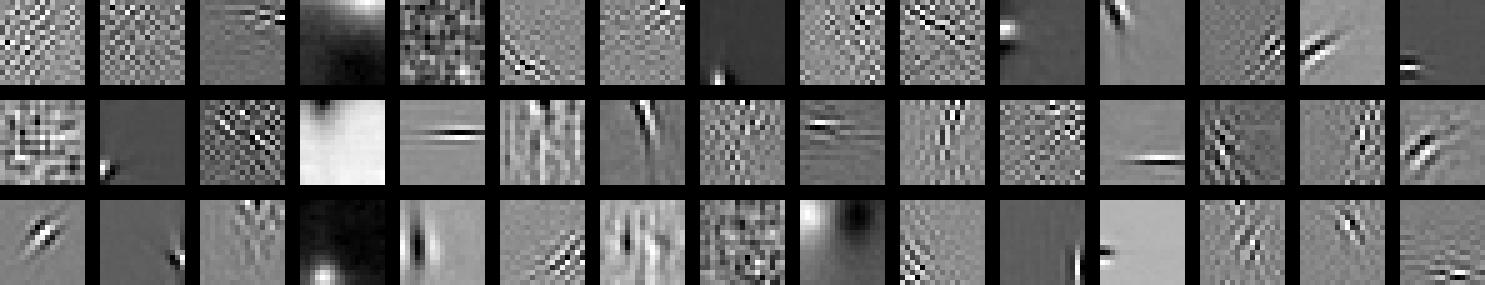}
  \caption{Random selection of weights in the input layer (top) and output layer (bottom) for an MLP with two hidden layers.}
\label{fig:patchesinout_2hl}
\end{figure}

\paragraph{Two hidden layers:} We will start by studying an MLP with
architecture ($17\times172$, $2047$, $2047$,$17\times172$). We repeat the
experiment we performed on an MLP with a single hidden layer and look at the
feature detectors and feature generators of the MLP, see
Figure~\ref{fig:patchesinout_2hl}. We notice that the feature generators look
relatively similar to the output bases of the MLP with a single hidden layer.
However, the feature detectors now look different: Many look somewhat noisy
(perhaps resembling grating filters) or seem to extract a feature that is
difficult to interpret. Intuition would suggest that these filters are in some
sense worse than those learned by the single hidden layer MLP. However, we have
seen in Figure~\ref{fig:progress004} that better results are achieved with the
MLP with two hidden layers than with one hidden layer.

\begin{figure}[htbp]
  \centering
  \rotatebox{90}{\small feature detectors}
  \includegraphics[width=0.95\columnwidth]{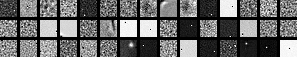}

  \vspace{5pt}
  
  \rotatebox{90}{\small feature generators}
  \includegraphics[width=0.95\columnwidth]{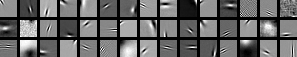}
  \caption{Random selection of weights in the input layer (top) and output layer (bottom) for an MLP with four hidden layers.}
\label{fig:patchesinout_4hl}
\end{figure}

\paragraph{Four hidden layers:} We look at the feature detectors and the output
bases of an MLP with architecture ($17\times17$, $2047$, $2047$, $2047$, $2047$,
$17\times17$), see Figure~\ref{fig:patchesinout_4hl}. The output bases resemble those
of the MLPs with one and two hidden layers. The feature detectors however look
still noisier than those of the MLP with two hidden layers. The results
achieved with the MLP with four hidden layers are again better than those
achieved with a two hidden layer MLP, see Figure~\ref{fig:progress004}. We
conclude that feature detectors that look noisy or are just difficult to
interpret do not necessarily lead to worse denoising results.

\begin{figure}[htbp]
  \centering
  \rotatebox{90}{\small feature detectors}
  \includegraphics[width=0.95\columnwidth]{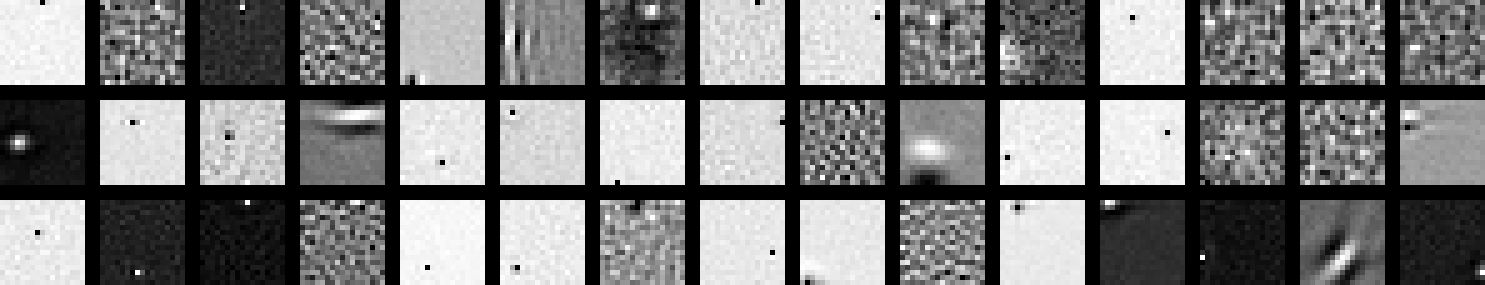}

  \vspace{5pt}
  
  \rotatebox{90}{\small output patterns}
  \includegraphics[width=0.95\columnwidth]{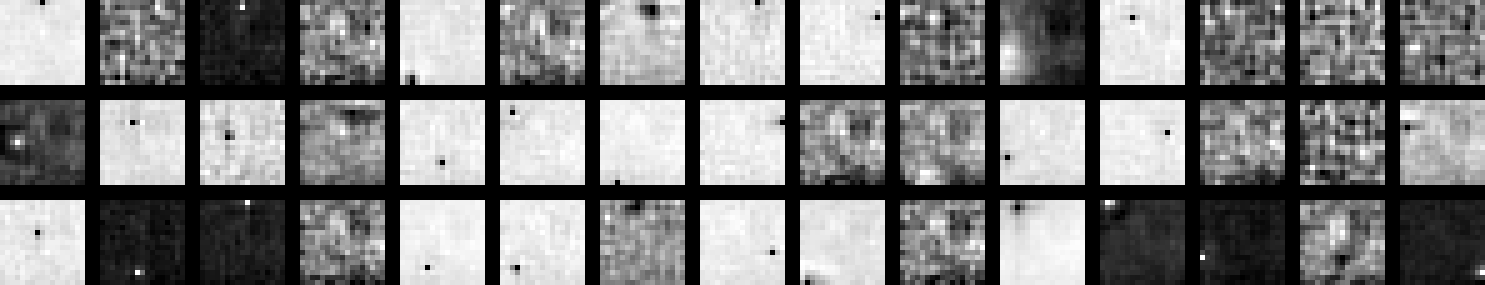}
  \caption{Feature detectors (top) and outputs (bottom) corresponding to each
  feature detector, using an MLP with two hidden layers. The detection of one feature
  causes the generation of a similar feature in the output.}
\label{fig:patchcorrespondences_2hl}
\end{figure}

\begin{figure}[htbp]
  \centering
  \rotatebox{90}{\small feature detectors}
  \includegraphics[width=0.95\columnwidth]{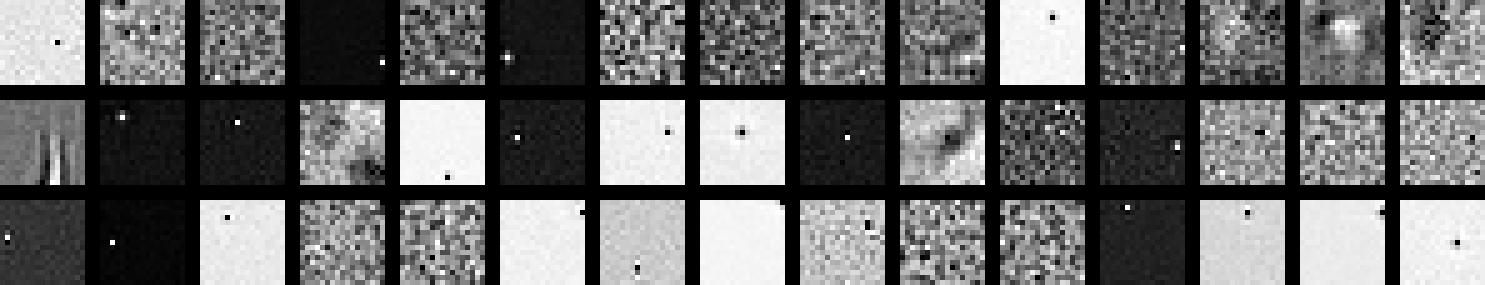}

  \vspace{5pt}
  
  \rotatebox{90}{\small output patterns}
  \includegraphics[width=0.95\columnwidth]{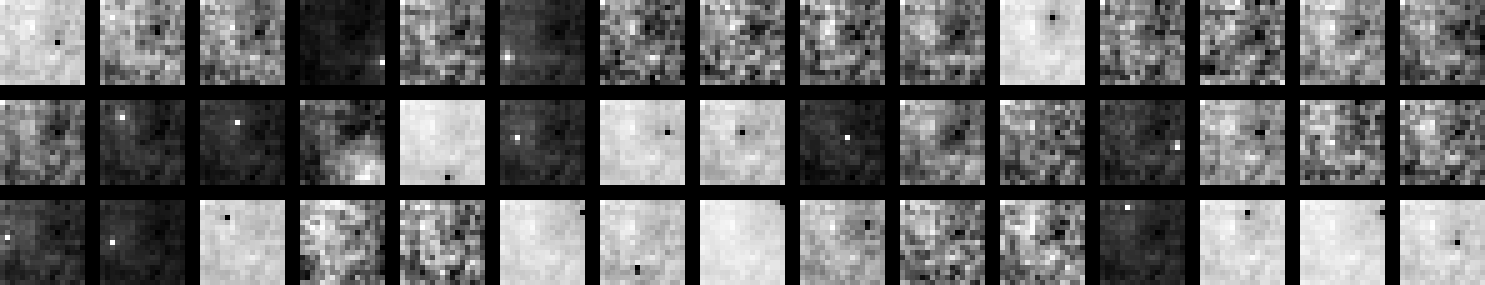}
  \caption{Feature detectors (top) and outputs (bottom) corresponding to each
  feature detector, using an MLP with four hidden layers. The detection of one feature
  causes the generation of a similar feature in the output.}
\label{fig:patchcorrespondences_4hl}
\end{figure}

\paragraph{Outputs corresponding to feature detectors:} In the MLP with a
single hidden layer, there was a clear correspondence between feature detectors
and feature generators: The feature generators looked identical to their corresponding
feature detectors.  This correspondence is lost in MLPs with more hidden
layers, due to the additional hidden layer separating feature detectors from
output bases. Can we still find a connection between feature detectors and
corresponding outputs? To answer this question, we activate a single unit in
the first hidden layer: The unit is assigned value $1$ and all other units are
set to $0$. We then perform a forward pass through the MLP, but completely
ignore the input of the MLP. Doing so provides us with an tentative answer to
the question: What output is caused by the detection of one feature? The answer
is only tentative because several features are usually detected simultaneously.
The activation of more hidden units can cause additional non-linear effects due
to the $\tanh$-functions in the MLP.
Figure~\ref{fig:patchcorrespondences_2hl}~and~\ref{fig:patchcorrespondences_4hl}
show the outputs obtained with an MLP with two and four hidden layers,
respectively. Also shown are the feature detectors corresponding to the hidden
units causing the outputs. We observe a similar correspondence between feature
detectors and outputs as in the case of a single hidden layer MLP. The effect
is more visible with the MLP with two hidden layers than with the MLP with four
hidden layers. The fact that the outputs do not perfectly correspond to their
feature detectors can be explained by the fact that during training, features
are never detected separately, but always in combination with other features.

\begin{figure}[t]
  \centering
  \rotatebox{90}{\small input patterns}
  \includegraphics[width=0.95\columnwidth]{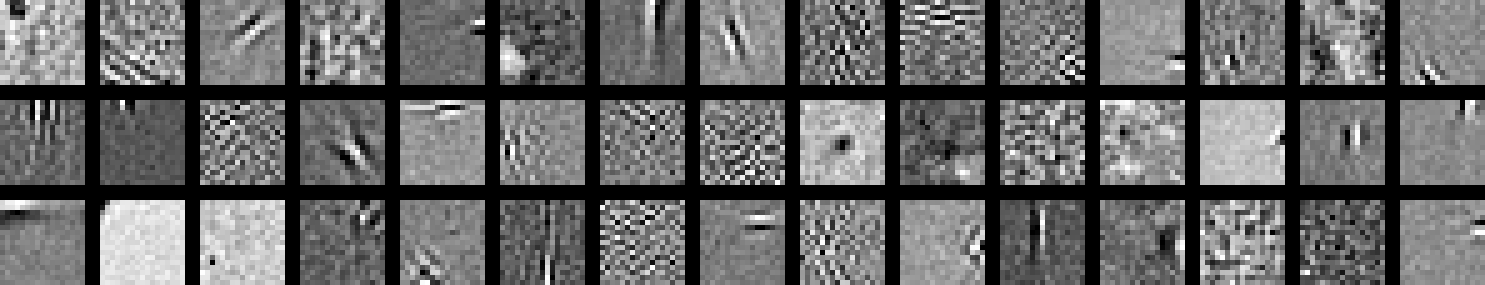}

  \vspace{5pt}
  
  \rotatebox{90}{\small feature generators}
  \includegraphics[width=0.95\columnwidth]{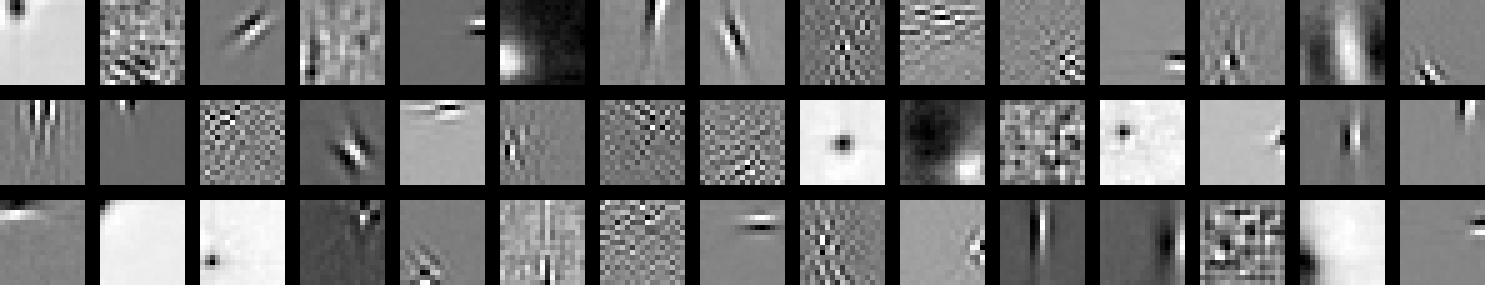}
  \caption{Features discovered through activation maximization (top) and
  corresponding feature generators (bottom), using an MLP with two hidden
  layers. The detection of one feature causes the generation of a similar
  feature in the output.}
\label{fig:AMfeaturecorrespondences_2hl}
\end{figure}

\begin{figure}[t]
  \centering
  \rotatebox{90}{\small input patterns}
  \includegraphics[width=0.95\columnwidth]{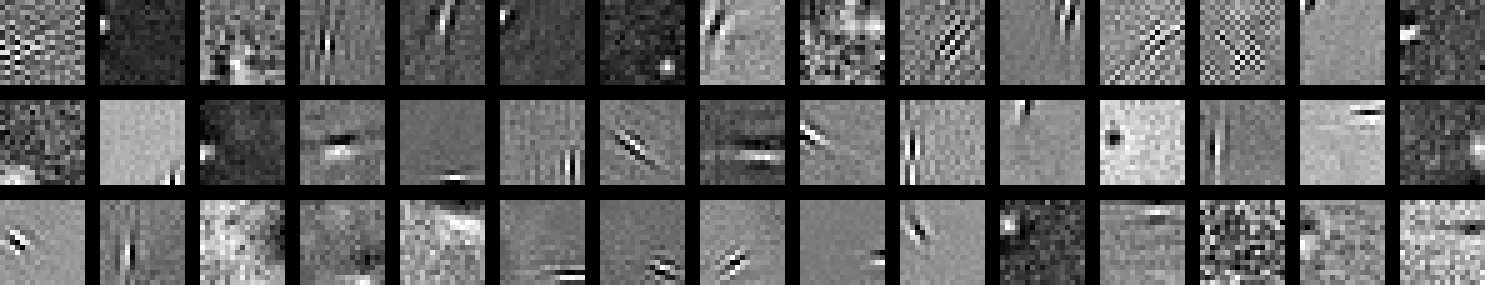}

  \vspace{5pt}
  
  \rotatebox{90}{\small feature generators}
  \includegraphics[width=0.95\columnwidth]{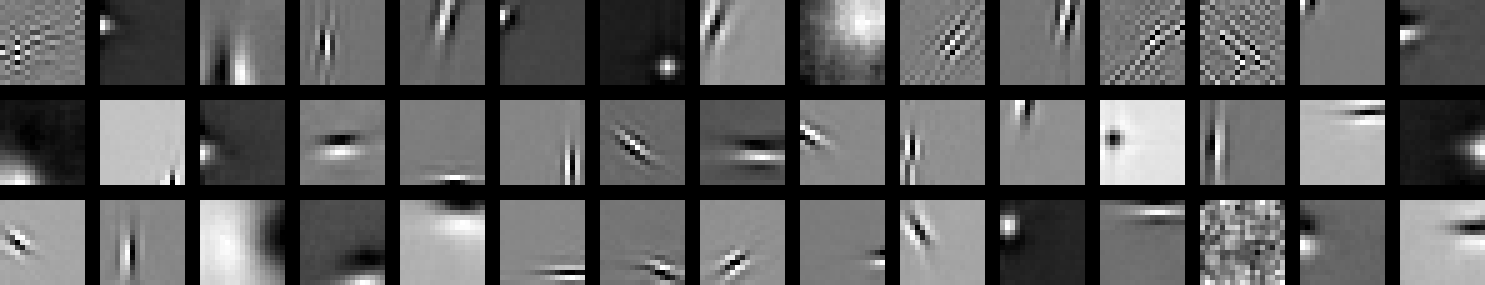}
  \caption{Features discovered through activation maximization (top) and
  corresponding feature generators (bottom), using an MLP with four hidden
  layers. The detection of one feature causes the generation of a similar
  feature in the output.}
\label{fig:AMfeaturecorrespondences_4hl}
\end{figure}

\label{sec:activationmaximization}
\paragraph{Inputs maximally activating single output bases:}
Which inputs cause the highest activation for each hidden neuron? Answering
this question should tell us which features the MLP responds to. We answer this
question using \emph{activation maximization}, proposed by
\citet{erhan2010understanding}.  Activation maximization is a gradient-based
technique for finding an input maximizing the activation of a neuron. We use
activation maximization with a step size of $0.1$. We initialize the patches
with samples drawn from a normal distribution with mean $0$ and unit variance.
We limit the norm of the patch to the norm of the initial patch. 

We apply activation maximization on neurons in the last hidden layer of the
MLPs with two and four hidden layers. The procedure indeed finds interesting
features, see
Figures~\ref{fig:AMfeaturecorrespondences_2hl}~and~\ref{fig:AMfeaturecorrespondences_4hl}.
Even more interesting is the fact that the features found through activation
maximization bear a strong resemblance to the feature generators connected to the
same hidden neuron.

\begin{figure}[htbp]
  \centering
  \rotatebox{90}{\small input patterns}
  \includegraphics[width=0.95\columnwidth]{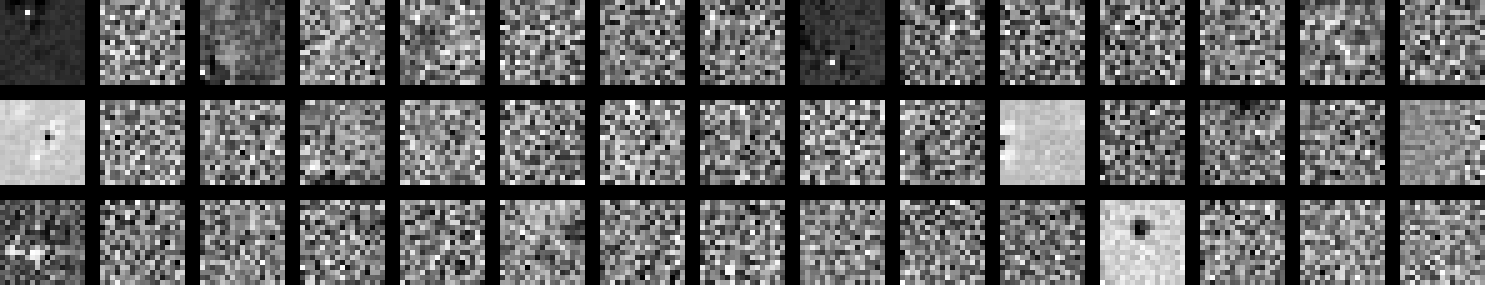}

  \vspace{5pt}
  
  \rotatebox{90}{\small output patterns}
  \includegraphics[width=0.95\columnwidth]{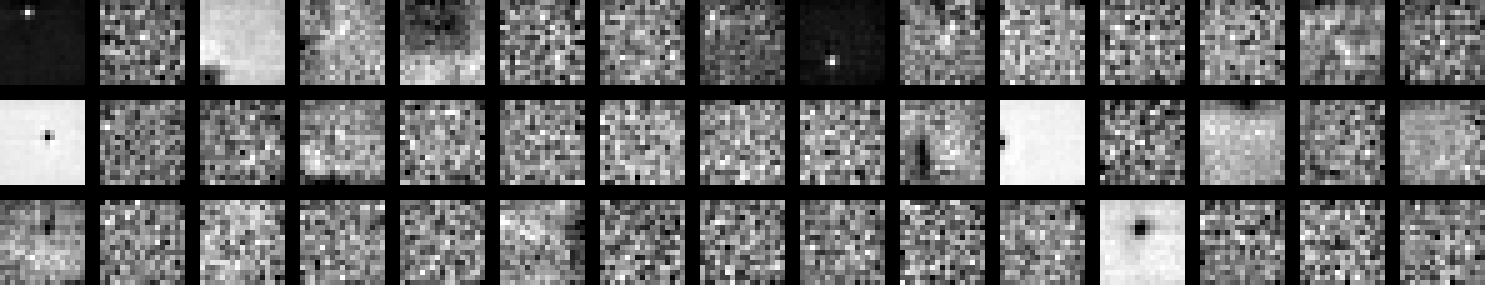}
  \caption{Input patterns discovered through activation maximization (top) and output patterns
  created using one active unit in the hidden layer (bottom), using an MLP with four hidden layers.
  We used the third hidden layer. We see a correspondence between the input and output patterns.}
\label{fig:AMrelatedtooutput_4hl}
\end{figure}

\paragraph{Input patterns vs. output patterns:} We also observe a
correspondence between the input patterns discovered through activation maximization
and output patterns created by activating a single hidden neuron in deeper layers.
Figure~\ref{fig:AMrelatedtooutput_4hl} demonstrates this correspondence in the 
third hidden layer of an MLP with four hidden layers.

\paragraph{Summary:} MLPs with more hidden layers tend to have feature
detectors that are not easily interpretable. In fact, one might be tempted to
conclude that they are inferior in some way to the feature detectors learned by
an MLP with a single hidden layer, because many of the feature detectors look
noisy. However, the denoising results obtained with MLPs with more hidden
layers is superior. The visual appearance of the feature detectors is therefore
not a disadvantage. The better denoising results can be explained by the higher
capacity of MLPs with more hidden layers.  MLPs with more hidden layers also
seem to operate according to the same principle as MLPs with a single hidden
layer: If a feature is detected in the noisy patch, a weighted version of the
feature is added to the denoised patch.

\subsection{MLPs with larger inputs}
We now consider the MLP that provided the best results on AWG noise with
$\sigma=25$, see Figure~\ref{fig:progress004}. The MLP has architecture
($39\times39$, $3072$, $3072$, $2559$, $2047$, $17\times17$). The main difference between
this MLP and the previous ones is that the input patches are larger than the
output patches. An additional difference is the somewhat larger architecture.

\begin{figure}[htbp]
  \centering
  \rotatebox{90}{\small feature detectors}
  \includegraphics[width=0.95\textwidth]{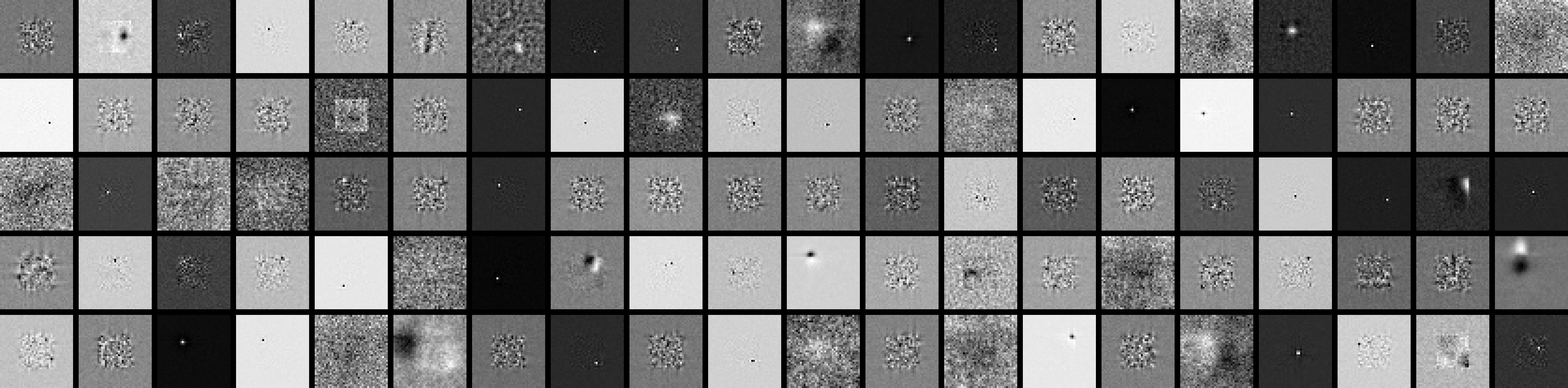}

  \vspace{5pt}
  
  \rotatebox{90}{\small feature generators}
  \includegraphics[width=0.95\textwidth]{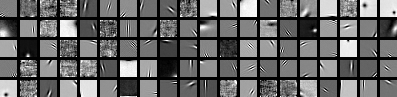}
  \caption{Random selection of weights in the input layer (top) and output
  layer (bottom) of the MLP providing the best results:
  $(39,3072,3072,2559.2047,17)$. This MLP has input patches of size
  $39\times39$ and output patches of size $17\times17$.}
\label{fig:patchesinout}
\end{figure}

\paragraph{Feature detectors and feature generators:}
Figure~\ref{fig:patchesinout} shows a set of feature detectors and feature generators
for the MLP with larger input patches. The feature generators look similar to those
learned by other MLPs. However, the feature detectors again look somewhat
different: many seem to focus on the center area of the input patch. In
addition, many look noisy. The fact that many feature detectors focus on
the center area of the input patch can be explained by the fact that the output
patches are smaller than the input patches. The target patches correspond to
the center region of the input patches. Correlations between pixels fall with
distance, which implies that the pixels at the outer border of the input patch
should be the least important for denoising the center patch.

\begin{figure}[htbp]
  \centering
  \begin{tabular}{cc}
    \includegraphics[width=0.45\textwidth]{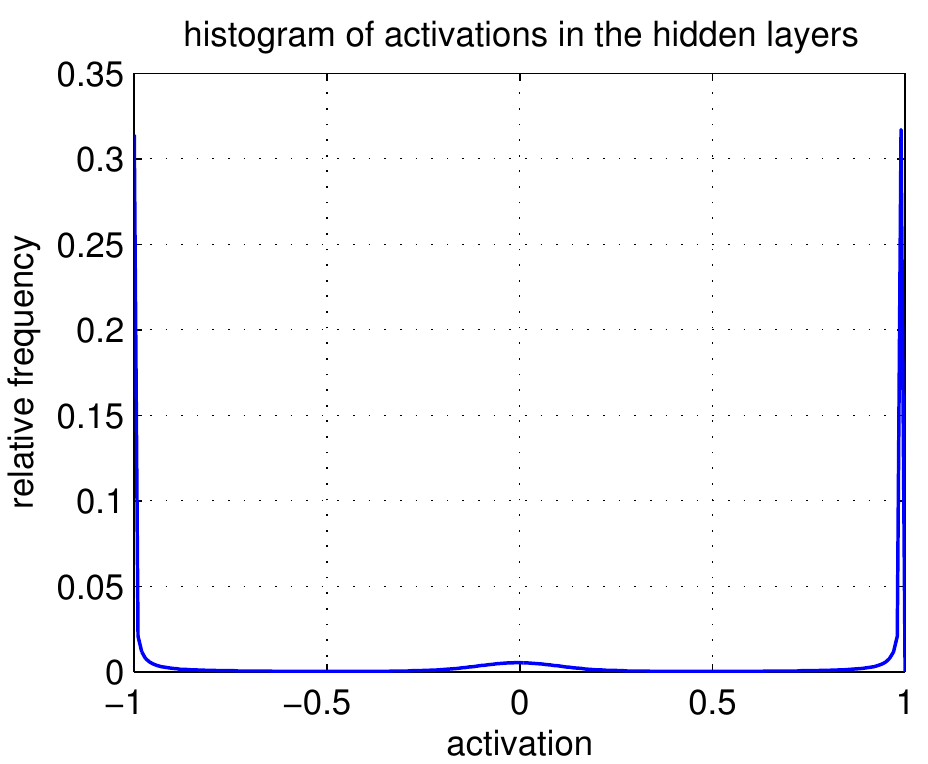} &
    \includegraphics[width=0.45\textwidth]{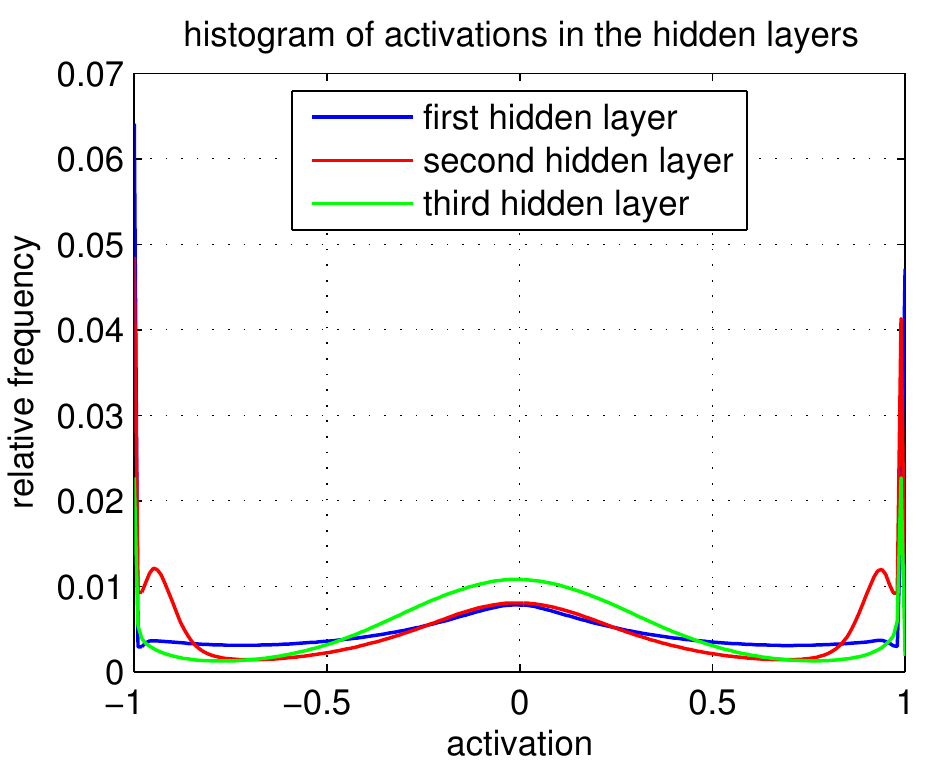} \\
    (a) & (b)
  \end{tabular}
  \caption{(a) Histograms of the activations in the last hidden layer.
  (b) Histograms of the activations in the first three hidden layers.}
  \label{fig:histogram}
\end{figure}

\paragraph{Activations:} The activations in the last hidden layer are
almost completely binary, see Figure~\ref{fig:histogram}a. This effect was also
observed on an MLP with a single hidden layer, but is now even more pronounced.
The activations in the other hidden layers are not binary: They frequently lie
somewhere between $-1$ and $1$, see Figure~\ref{fig:histogram}b and resemble 
a typical distribution~\citep{GlorotAISTATS2010}. The denoised
output patches are therefore essentially constructed from binary codes
weighting elements in a dictionary.

\begin{figure}[htbp]
  \centering
    \includegraphics[width=\columnwidth]{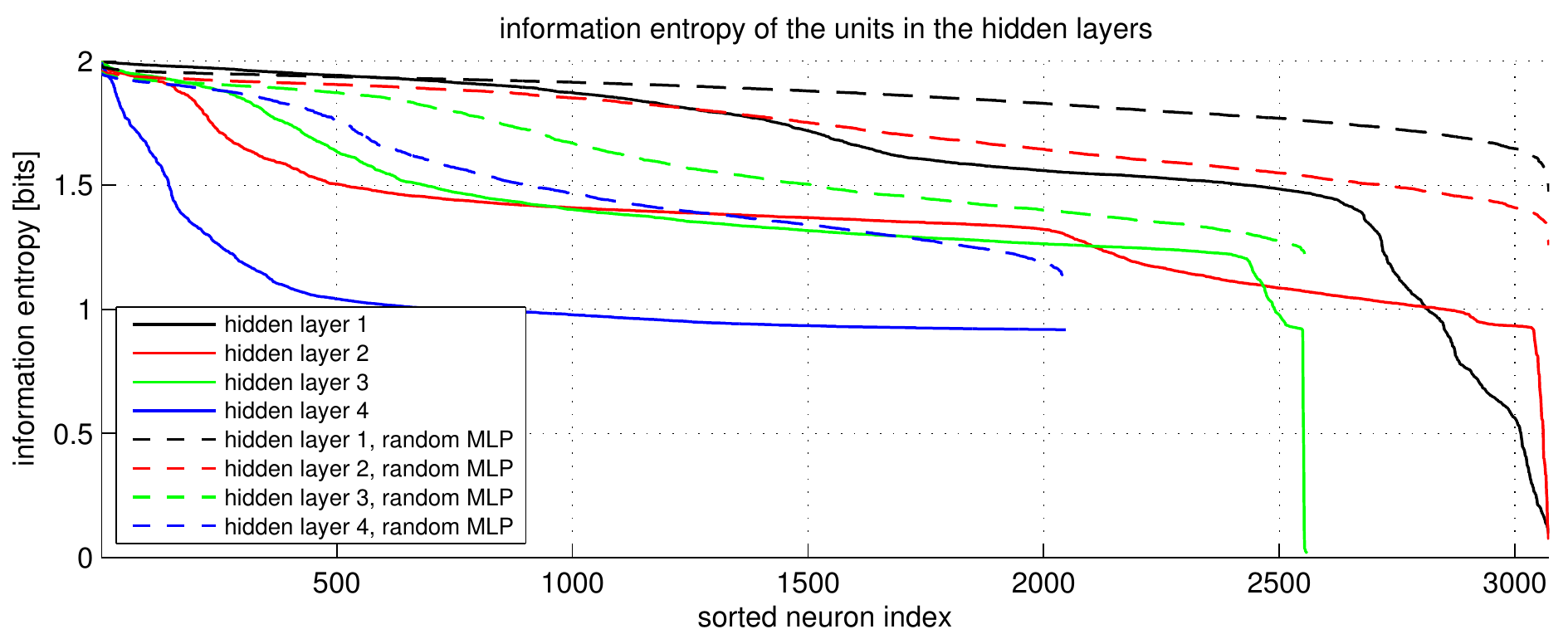}
    \caption{Information entropy of the units in the different hidden layers.
    All units in the last hidden layer have high entropy.}
  \label{fig:entropy002}
\end{figure}

\paragraph{Entropy:} An MLP with a single hidden layer had some hidden units
with entropy close to zero. Is this also the case for MLPs with more hidden
layers?  We evaluate the information entropy of the units in the various hidden
layers, see Figure~\ref{fig:entropy002}. We again used four bins of equal size.
We also compare against a randomly initialized MLP. We observe that the entropy
is lower for the trained MLP than for the randomly initialized MLP, which was
also observed on an MLP with a single hidden layer.  However, this time, all
the units in the last hidden layer have high information entropy. In the
remaining layers, some units have low information entropy.

\begin{figure}[htbp]
  \centering
    \rotatebox{90}{\tiny feature detectors}
    \includegraphics[width=0.95\columnwidth]{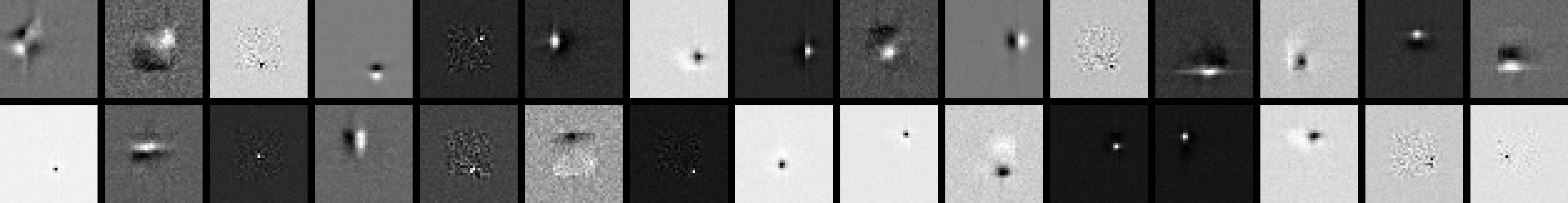}
      
    \vspace{5pt} 
     
    \rotatebox{90}{\tiny feature detectors}
    \includegraphics[width=0.95\columnwidth]{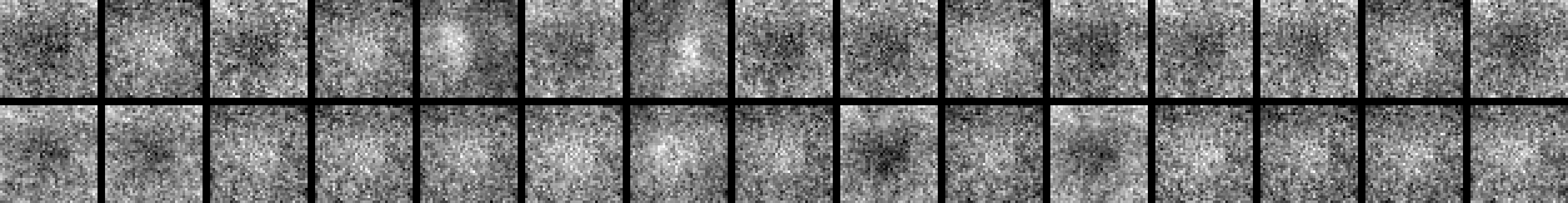}
    \caption{Feature detectors of the units with the highest (top) and lowest (bottom) entropy.}
  \label{fig:entropy_featuredetectors}
\end{figure}
\begin{figure}[htbp]
    \rotatebox{90}{\tiny feature generators}
    \includegraphics[width=0.95\columnwidth]{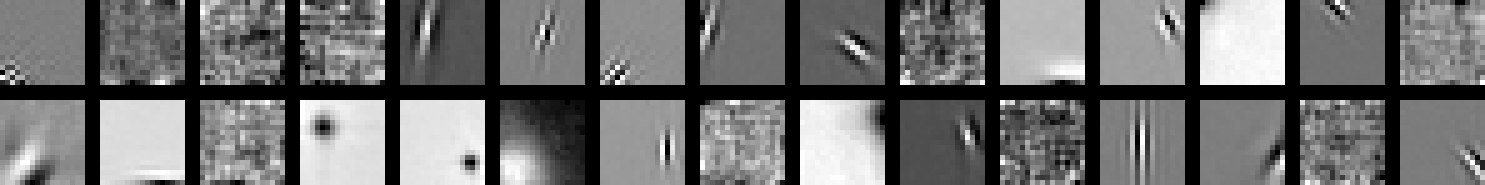}
      
    \vspace{5pt} 
     
    \rotatebox{90}{\tiny feature generators}
    \includegraphics[width=0.95\columnwidth]{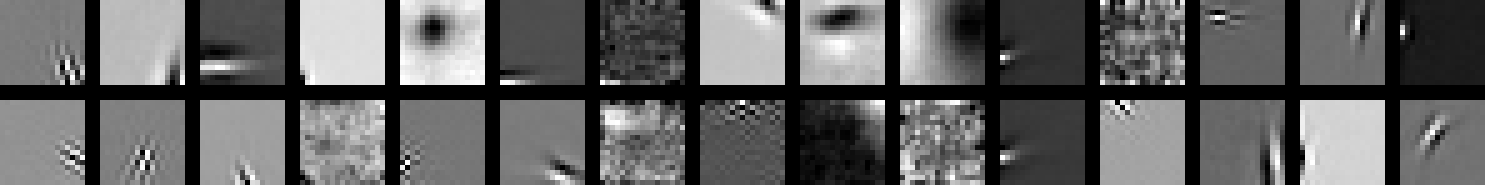}
    \caption{Feature generators of the units with the highest (top) and lowest (bottom) entropy.}
  \label{fig:entropy_outputbases}
\end{figure}

Figure~\ref{fig:entropy_featuredetectors} shows the feature detectors connected
to the units with highest and lowest entropy, respectively.
Figure~\ref{fig:entropy_outputbases} shows the feature generators with the highest
and lowest entropy, respectively. The feature detectors with the highest
entropy look different from the feature detectors with the lowest entropy.
The latter all look similar: All are noisy and seem to loosely focus on a
region in the center of the patch. The feature detectors with the highest
entropy look more clearly defined. For the feature generators, no clear difference is
observed. This is perhaps due to the fact that all output bases have high
information entropy. The feature detectors with the lowest entropy almost
always have the same activation value and are therefore probably also not very
helpful in terms of denoising results. 

\begin{table}
  \centering
  \begin{tabular}{l|c}
    image name & PSNR \\  
    \hline\hline 
    Barbara & 127.45dB \\  
    Boat & 187.76dB \\  
    Cameraman & 166.00dB \\  
    Couple & 192.58dB \\  
    Fingerprint & 198.12dB \\  
    Hill & 192.05dB \\  
    House & 195.68dB \\  
    Lena & 192.19dB \\  
    Man & 190.21dB \\  
    Montage & 174.14dB \\  
    Peppers & 189.49dB \\  
    Noise & 161.28dB
  \end{tabular}
  \caption{Ability of the dictionary to approximate images.}
  \label{tab:resultsApprox}
\end{table}

\begin{figure}[t]
    \includegraphics[width=\columnwidth]{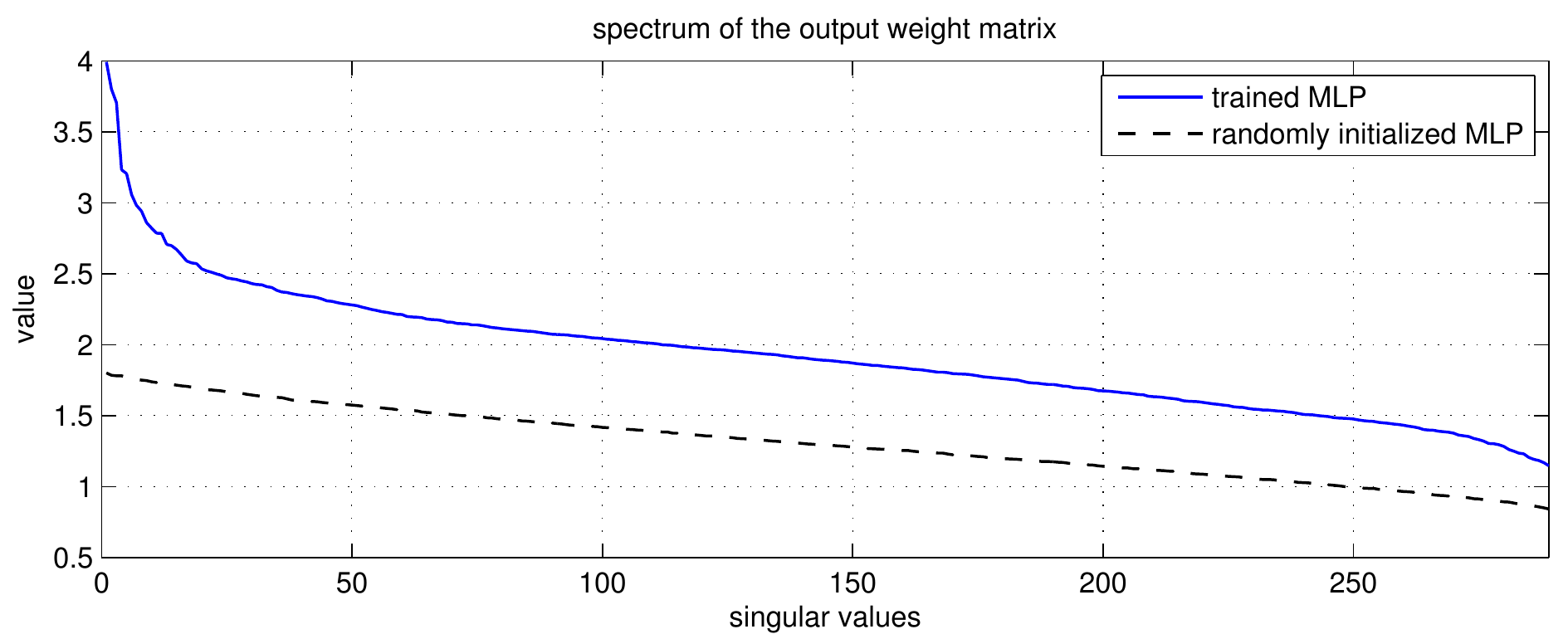}
    \caption{Spectrum of the output bases.}
  \label{fig:weightspectrum002}
\end{figure}

\paragraph{Approximation ability:} We have seen that the MLP does not
perform as well as other methods on the image ``Barbara''. We now ask the
question: Is the dictionary formed by the last layer of the MLP the reason why
some images cannot be denoised well? In other words, is it possible to
approximate any image patch arbitrarily well using that dictionary, or are
there images that are difficult to approximate?  An additional constraint is
that the code vector weighting the dictionary is not allowed to contain values
below $-1$ or above $1$ due to the $\tanh$ layer.

To answer this question, we try to approximate images patch-wise using the
dictionary formed by the last layer. In other words, we try to approximate each
image patch $x$ of a clean image using our dictionary $D$, and proceed in a
sliding-window manner. We average in the regions of overlapping patches. Formally,
we solve the following problem:
\begin{equation}
  \min ||x - D\alpha||_{2} \quad \textrm{ s.t.} -1 \preceq \alpha \preceq 1 .
\end{equation}
Table~\ref{tab:resultsApprox} lists the results obtained on the $11$ standard
test images, as well as one image containing only white Gaussian noise with
$\mu=127.5$ and $\sigma=25$ (row ``Noise'').  We see that all images (including
the noise image) can be almost perfectly approximated, though the result on
image Barbara is slightly worse than on other images. We therefore conclude
that the dictionary in the last layer by itself cannot be the reason why some
images are not denoised well. Any image can be well approximated 
using the dictionary and codes with values in range from -1 to 1.

A related observation is that the weights in the last layer have no zero
singular values, see Figure~\ref{fig:weightspectrum002}. This implies that the
matrix has full rank and can therefore approximate any patch, when the lower-
and upper-bound constraints are disregarded. We also observe that the spectrum
is relatively flat, which was also the case for the MLP with a single hidden
layer. This implies that the output bases are diverse.

\begin{table}
  \centering
  \begin{tabular}{l||ccc}
  image &  KSVD \citep{aharon2006rm} & MLP & ``MLP + OMP'' \\  
  \hline\hline 
  Barbara &  \textit{29.49}dB  &  \textbf{29.52}dB  &  28.23dB  \\  
  Boat &  \textit{29.24}dB  &  \textbf{29.95}dB  &  28.93dB  \\  
  Cameraman &  \textit{28.64}dB  &  \textbf{29.60}dB  &  28.32dB  \\  
  Couple &  \textit{28.87}dB  &  \textbf{29.75}dB  &  28.66dB  \\  
  Fingerprint &  \textit{27.24}dB  &  \textbf{27.67}dB  &  26.88dB  \\  
  Hill &  \textit{29.20}dB  &  \textbf{29.84}dB  &  28.95dB  \\  
  House &  \textit{32.08}dB  &  \textbf{32.52}dB  &  30.12dB  \\  
  Lena &  \textit{31.30}dB  &  \textbf{32.28}dB  &  30.65dB  \\  
  Man &  \textit{29.08}dB  &  \textbf{29.85}dB  &  28.95dB  \\  
  Montage &  \textit{30.91}dB  &  \textbf{31.97}dB  &  30.21dB  \\  
  Peppers &  \textit{29.69}dB  &  \textbf{30.27}dB  &  29.08dB  \\  
  \end{tabular}
  \caption{Using the MLP's dictionary in combination with OMP.}
  \label{tab:resultsOMP}
\end{table}

\paragraph{Combining the dictionary with sparse coding:} Dictionary-based
methods for image denoising such as KSVD typically denoise by approximating a
noisy image patch using a sparse linear combination of the elements in the
dictionary. More formally, one attempts to solve the following problem:
\begin{equation}
\label{eqn:sparsecoding}
  \min ||\alpha||_{0} \quad \textrm{s.t. } ||y - D\alpha||_{2} \leq \epsilon
\end{equation}
where $y$ is a noisy image patch, $\epsilon$ is a pre-defined parameter and
$||\cdot||_{0}$ refers to the $\ell_{0}$ pseudo-norm. Approximate solutions to
this problem can be found using OMP \citep{pati1993orthogonal}. The denoised
patch $\hat{x}$ is given by $\hat{x} = D\alpha$. Denoising is performed in a
sliding-window manner and averaging is performed where patches overlap.

We ask the question: Can the dictionary learned by the MLP be used in
combination with this sparse coding approach? We denoise the $11$ standard test
images with AWG noise, $\sigma=25$ using the dictionary learned by the MLP and
solve equation~(\ref{eqn:sparsecoding}) approximately using OMP. We set
$\epsilon$ similarly to KSVD \citep{aharon2006rm}: $\epsilon =
n((C\sigma)^{2})$, where n is the dimensionality of the patches ($289$) and $C$
is a hyper-parameter. We found the best value of $C$ to be $1.05$. We
normalized all columns of $D$ to have unit norm. The results of this approach
are summarized in Table~\ref{tab:resultsOMP}. The PSNR of the noisy images is 
approximately $20.18$dB.

The denoising results of this approach are not very good. We therefore conclude
that the dictionary's ability to denoise is strongly dependent on the codes
provided to it. The first three hidden layers of the MLP serve as a mechanism
for creating good codes for the last layer.

\begin{figure}[htbp]
  \centering
  \includegraphics[width=\textwidth]{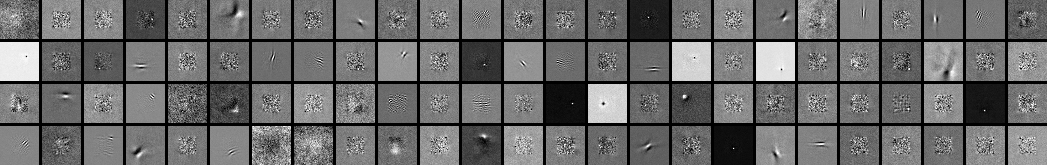}
  (a) Input patterns maximizing the activation of neurons in the second hidden layer.

  \vspace{5pt}
  
  \includegraphics[width=\textwidth]{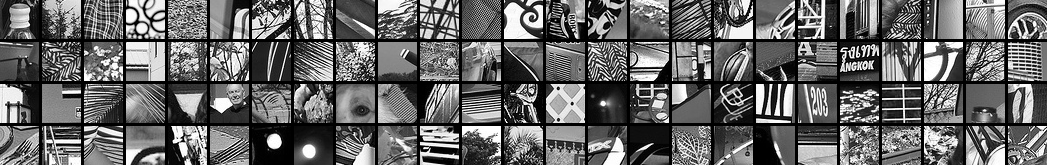}
  (b) Natural image patches maximizing the absolute activation of neurons in the second hidden layer.

  \vspace{5pt}

  \includegraphics[width=\textwidth]{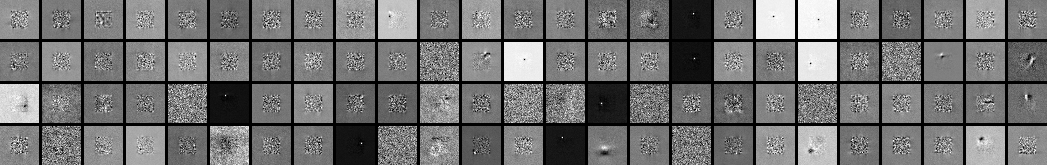}
  (c) Input patterns maximizing the activation of neurons in the third hidden layer.

  \vspace{5pt}
  
  \includegraphics[width=\textwidth]{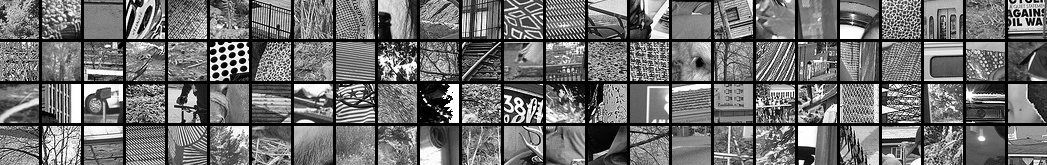}
  (d) Natural image patches maximizing the absolute activation of neurons in the third hidden layer.

  \vspace{5pt}

  \includegraphics[width=\textwidth]{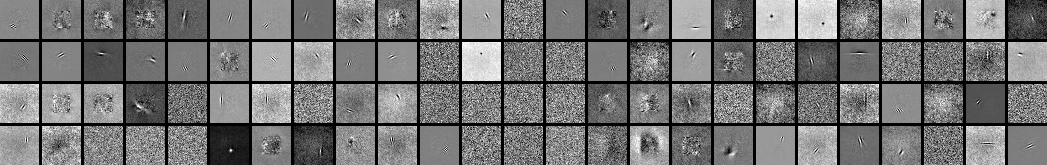}
  (e) Input patterns maximizing the activation of neurons in the fourth hidden layer.
  
  \vspace{5pt}

  \includegraphics[width=\textwidth]{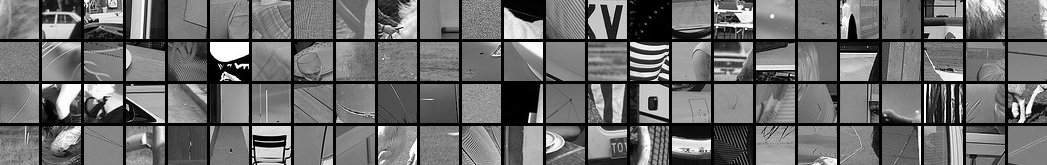}
  (f) Natural image patches maximizing the absolute  activation of neurons in the fourth hidden layer.
\caption{What features does the MLP respond to?}
\label{fig:patches}
\end{figure}

\paragraph{Inputs maximizing the activation of neurons:} 
Which inputs cause the highest activation for each neuron? We answer this
question using two approaches: (i) Activation maximization
\citep{erhan2010understanding} and (ii) evaluating the activation values for a
large number of (non-noisy) image patches.

We perform activation maximization as described in
section~\ref{sec:activationmaximization}. We also run the MLP on a large number
of noise-free natural image patches.  For each neuron, we save the input
maximizing its absolute activation. We used $6768$ natural images, each
containing many thousand patches.  Figure~\ref{fig:patches} shows the input
patterns found through activation maximization as well as the input patches
found by inspecting a larger number of natural image patches. We make a number
of observations.

\begin{itemize}
  \item \textbf{Focus on the center part:} The patterns found through
  activation maximization mostly focus on the center part of the patches. This
  intuitively makes sense: The most important part of the input patch is expected
  to be the area covered by the output patch. In addition, pixel correlations
  fall with distance, so pixels that are further away are expected to be less
  interesting.  There are exceptions however: Some patches seem to focus
  particularly on the patch border.

  \item \textbf{Gabor filters:} Many input patterns resemble Gabor filters. This is
  true for all hidden layers, but particularly for hidden layers two and four. We
  also observed this phenomenon in the output layer weights, see
  Figure~\ref{fig:patchesinout}.

  \item \textbf{Random looking patches:} Many input patterns look as if the pixels
  were set randomly. This is particularly true in hidden layer three.

  \item \textbf{Correlation to natural image patches:} Some input patterns found
  through activation maximization correlate well with patches found through
  exhaustive search through a set of natural image patches. For example the
  patches $6$ and $7$ from the right in the upper row of hidden layer four. In
  many cases however, it is not clear that the two procedures find correlating
  patches.  The fourth hidden layer patches seem to indicate that many neurons
  respond to features with a highly specific location and orientation.
\end{itemize}

\subsection{Comparing the importance of the feature detectors}
Some of the feature detectors look random or noisy, see
Figure~\ref{fig:patchesinout}. Are all the feature detectors useful or are the
noisy looking filters less useful?  We answer this question by observing the
behavior of the MLP when a set of feature detectors is removed (in other words,
when only a subset of feature detectors is used).  We evaluate the average
performance of the network on the $11$ standard test images. We remove a
feature detectors by replacing its weights with the average value of the
feature detector.

\begin{figure}[htbp]
  \centering
    \rotatebox{90}{\tiny feature detectors}
    \includegraphics[width=0.95\columnwidth]{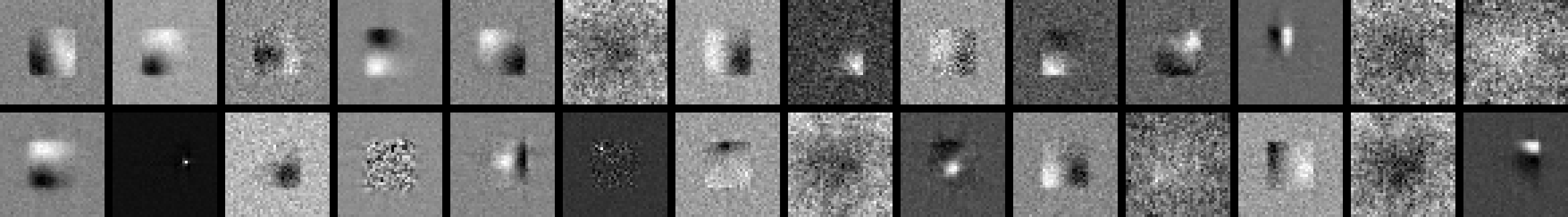}

    \vspace{5pt}

    \rotatebox{90}{\tiny feature detectors}
    \includegraphics[width=0.95\columnwidth]{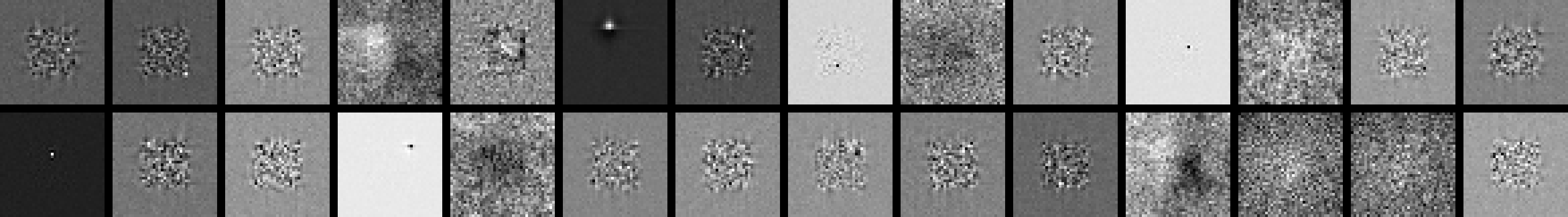}
    \caption{Most (top) and least (bottom) important feature detectors (using 1500 feature detectors).}
  \label{fig:importance001}
\end{figure}
We use an iterative procedure during which $1500$ feature detectors are chosen for each
iteration. The mean PSNR obtained is assigned to the feature detectors used during that
iteration. We average over iterations.  The feature detectors yielding the best results
(on average) are shown in the top row of Figure~\ref{fig:importance001} and the
feature detectors yielding the worst results are shown in the bottom row.

It seems that the feature detectors yielding good results on average are more easily
interpretable than the ones yielding worse results. The feature detectors yielding good
results seem to focus on large-scale features, whereas the filters yielding
worse results look more noisy.

\subsection{Effect of the type and strength of the noise on the feature detectors and feature generators}
All observations we have made on the feature detectors and feature generators
of the MLPs were made on MLPs trained to remove AWG noise with $\sigma=25$. We
will now make a number of observations for different types and strengths of
noise.

\begin{figure}[htbp]
  \centering
  \rotatebox{90}{\small feature detectors}
  \includegraphics[width=0.95\textwidth]{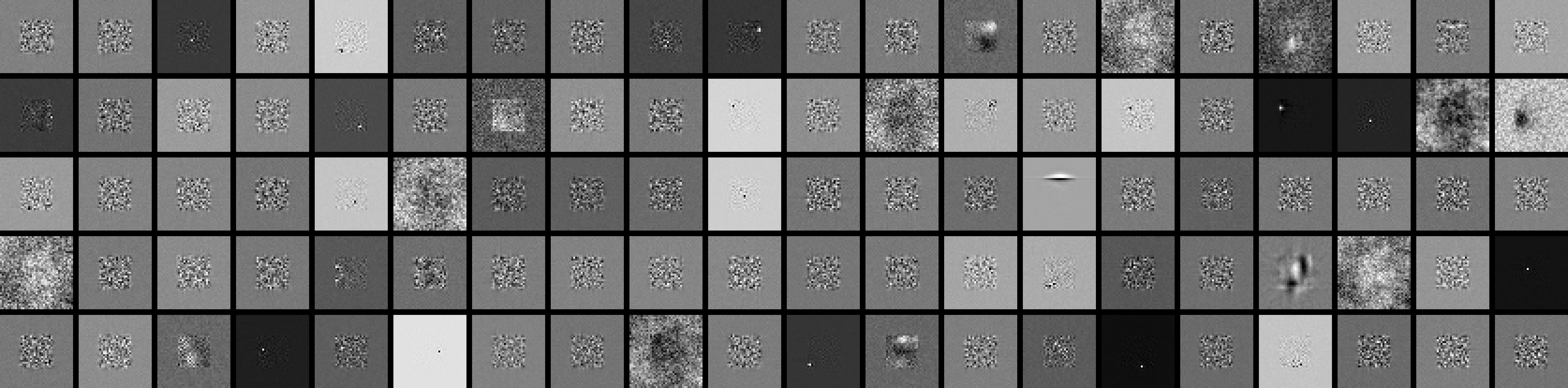}

  \vspace{5pt}
  
  \rotatebox{90}{\small feature generators}
  \includegraphics[width=0.95\textwidth]{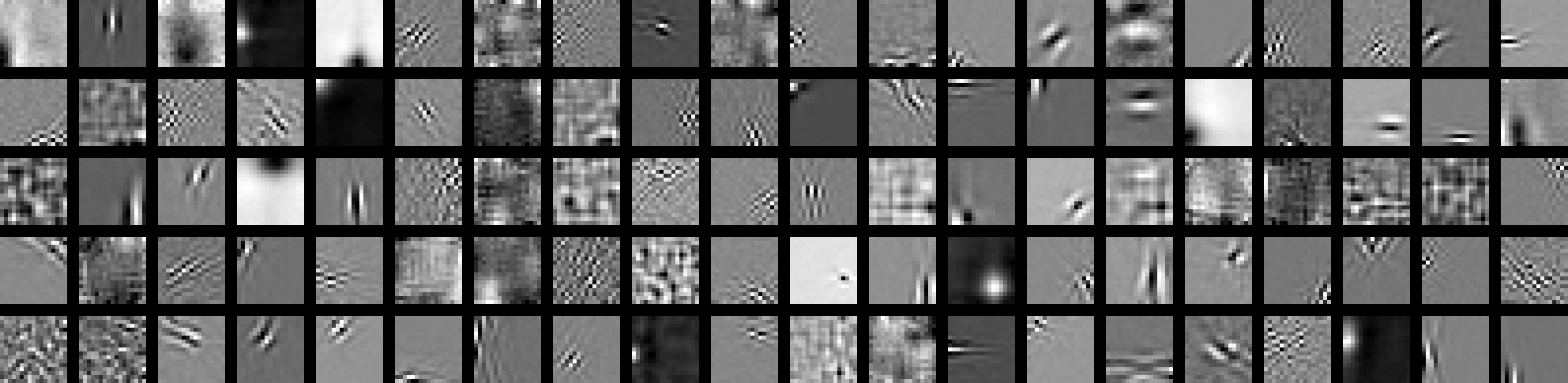}
  \caption{Random selection of weights in the input layer (top) and output layer (bottom) for $\sigma=10$}
\label{fig:patchesinout_sig10}
\end{figure}
\begin{figure}[htbp]
  \centering
  \rotatebox{90}{\small feature detectors}
  \includegraphics[width=0.95\textwidth]{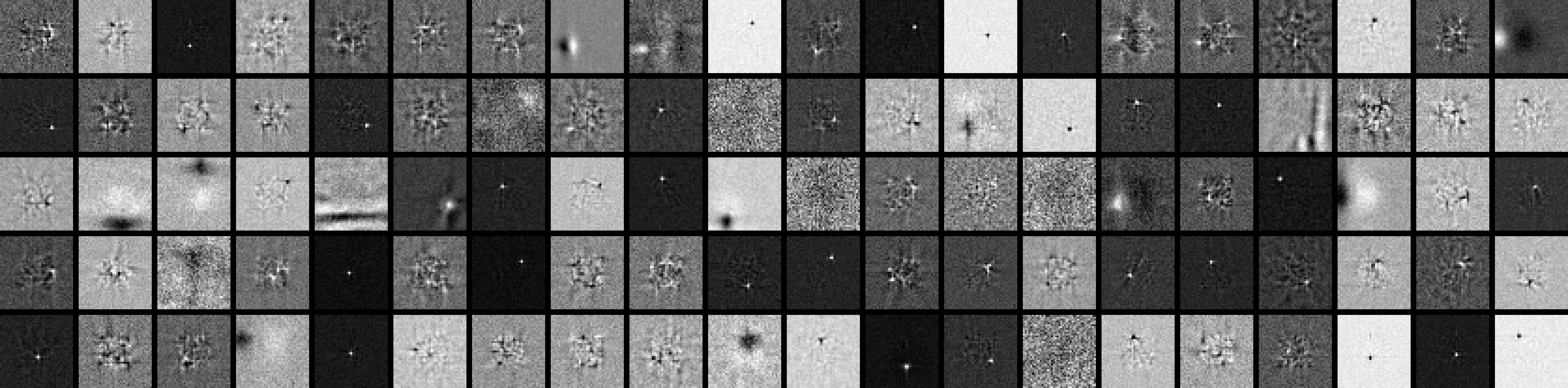}

  \vspace{5pt}
  
  \rotatebox{90}{\small feature generators}
  \includegraphics[width=0.95\textwidth]{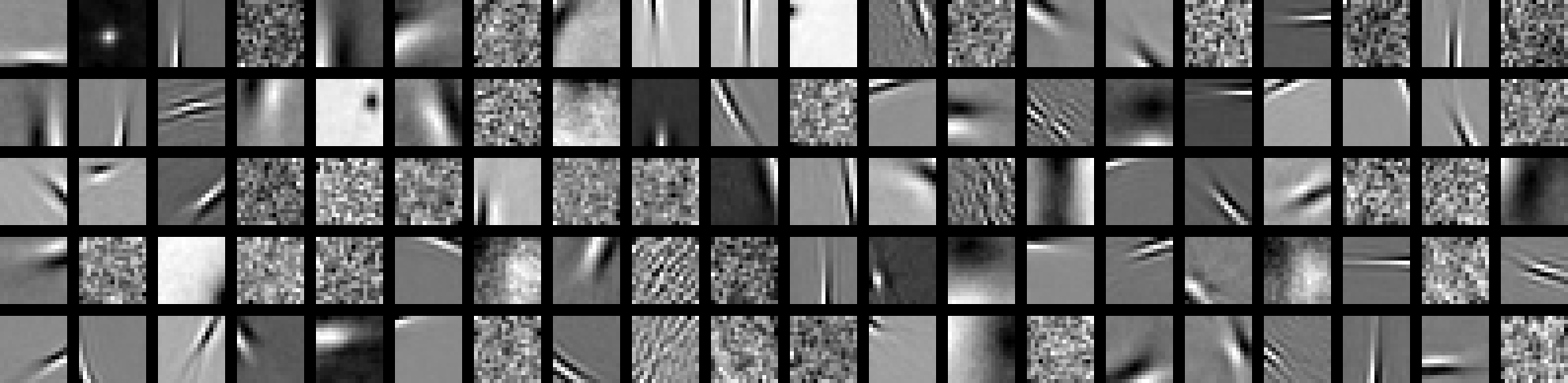}
  \caption{Random selection of weights in the input layer (top) and output layer (bottom) for $\sigma=75$}
\label{fig:patchesinout_sig75}
\end{figure}
\textbf{How does the strength of the noise affect the learned weights?}
Figure~\ref{fig:patchesinout_sig10} and Figure~\ref{fig:patchesinout_sig75}
show the feature detectors and feature generators for $\sigma=10$ and $\sigma=75$,
respectively. The feature generators look similar for the two noise levels.
However, the feature detectors look different: For $\sigma=10$, the feature detectors
almost always focus on the area covered by the output patch, whereas for
$\sigma=75$, the feature detectors also consider pixels that are further away. This is in
agreement with~\citet{levin2010natural}: When the noise is stronger, larger
input patches are necessary to achieve good results. We already provided a
similar explanation in Section~\ref{sec:othernoisevariances}.  This also
implies that it is unnecessary to use large input patches when the noise is
weak and explains why we achieved better results with smaller patches for
$\sigma=10$, see Figure~\ref{fig:progress009_nobm}.

\medskip

\begin{figure}[htbp]
  \centering
  \rotatebox{90}{\small feature detectors}
  \includegraphics[width=0.95\textwidth]{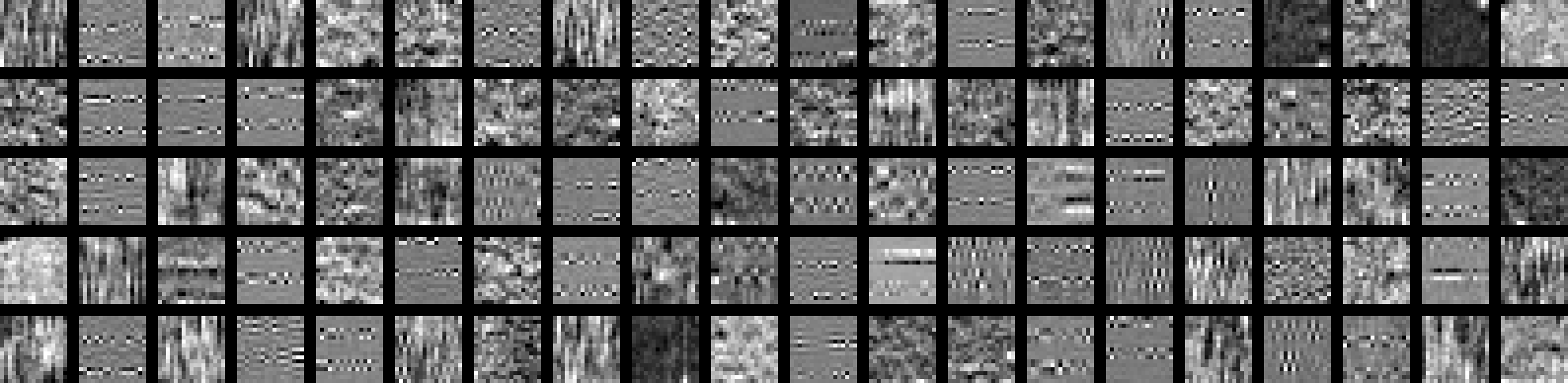}

  \vspace{5pt}
  
  \rotatebox{90}{\small feature generators}
  \includegraphics[width=0.95\textwidth]{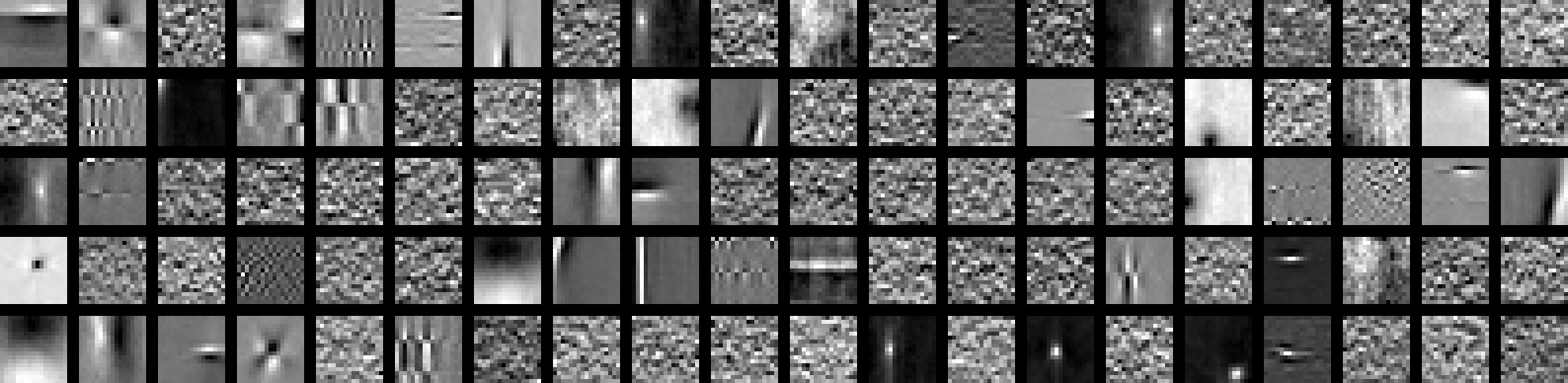}
  \caption{Random selection of weights in the input layer (top) and output layer (bottom) for stripe noise}
\label{fig:patchesinout_stripes}
\end{figure}

\begin{figure}[htbp]
  \centering
  \rotatebox{90}{\small feature detectors}
  \includegraphics[width=0.95\textwidth]{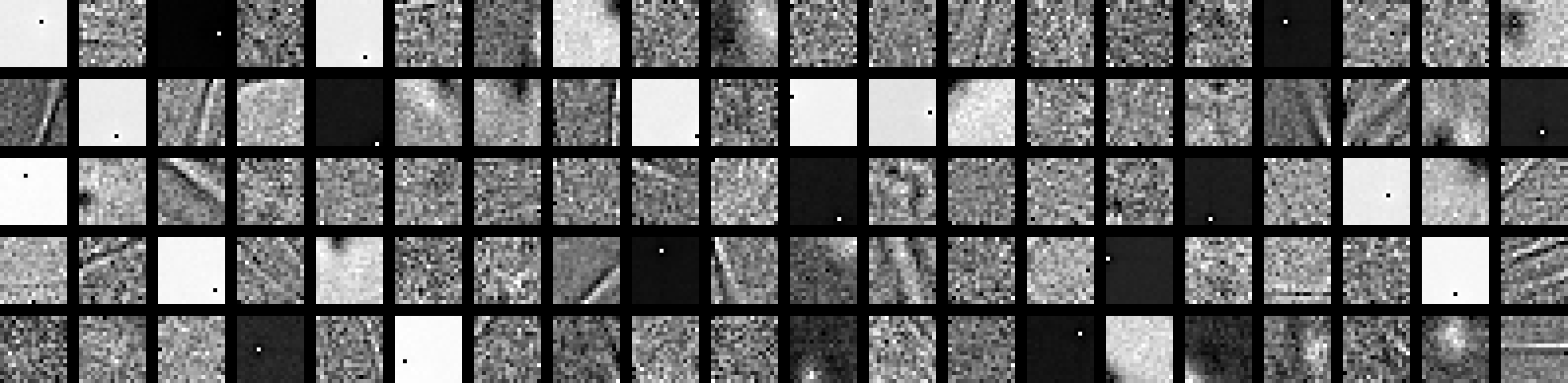}

  \vspace{5pt}
  
  \rotatebox{90}{\small feature generators}
  \includegraphics[width=0.95\textwidth]{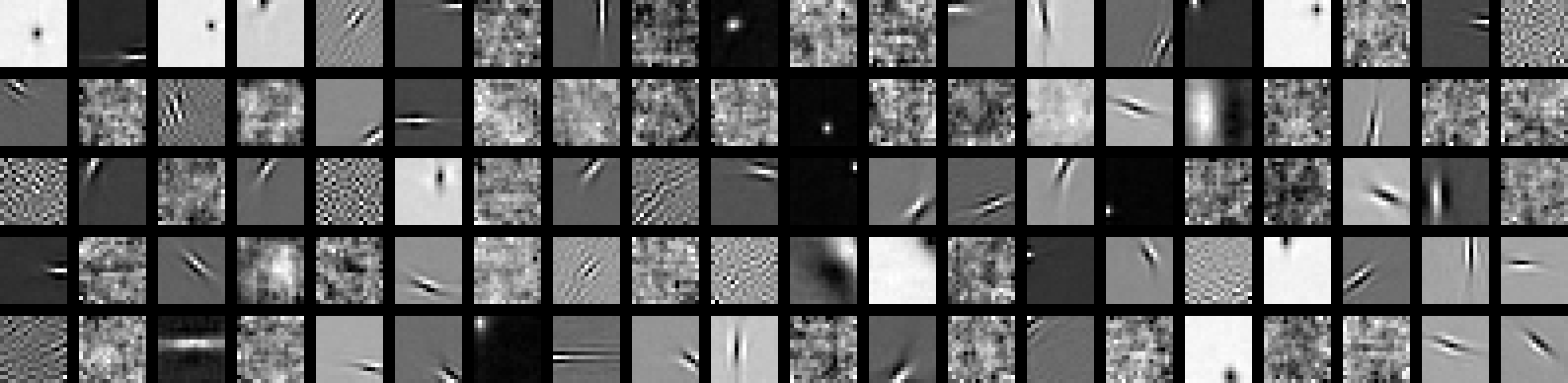}
  \caption{Random selection of weights in the input layer (top) and output layer (bottom) for salt and pepper noise}
\label{fig:patchesinout_saltandpepper}
\end{figure}

\begin{figure}[htbp]
  \centering
  \rotatebox{90}{\small feature detectors}
  \includegraphics[width=0.95\textwidth]{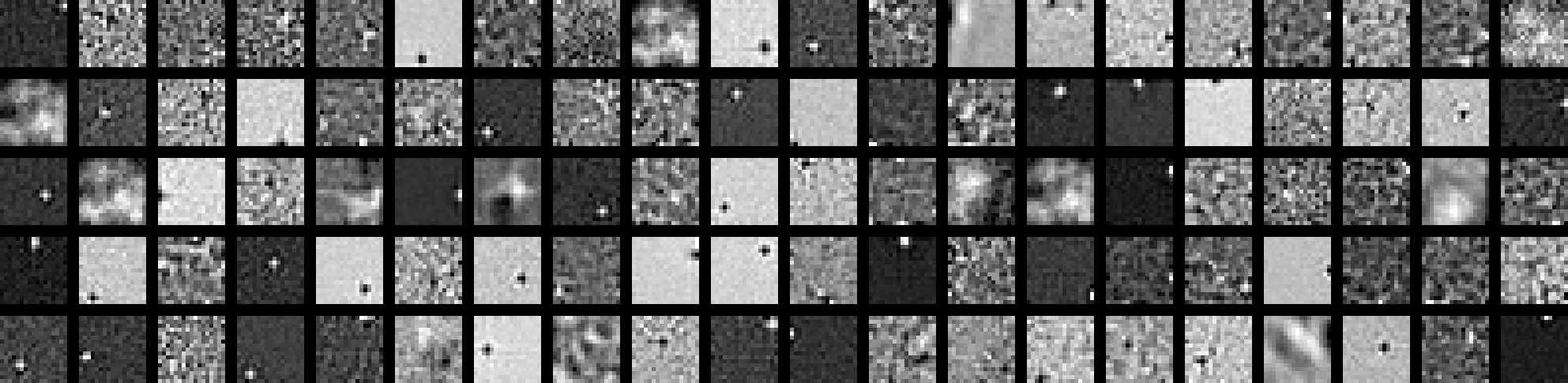}

  \vspace{5pt}
  
  \rotatebox{90}{\small feature generators}
  \includegraphics[width=0.95\textwidth]{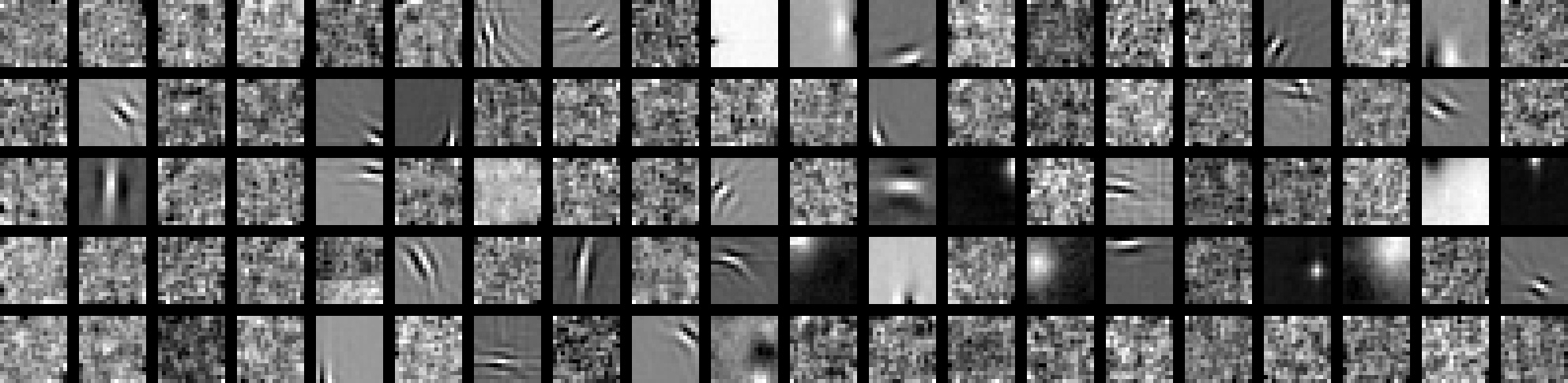}
  \caption{Random selection of weights in the input layer (top) and output layer (bottom) for JPEG noise}
\label{fig:patchesinout_jpeg}
\end{figure}

\paragraph{How does the type of the noise affect the learned weights?}
Figures~\ref{fig:patchesinout_stripes},~\ref{fig:patchesinout_saltandpepper}
and~\ref{fig:patchesinout_jpeg} show the feature detectors and feature
generators learned with stripe noise, salt-and-pepper noise and JPEG artifacts,
respectively. All patches in these figures are of size $17\times17$.  The input
weights are strongly affected by the type of the noise: For horizontal stripe
noise, the feature detectors often have horizontal features that also look like
stripes. For salt-and-pepper noise, the feature detectors are often filters
focussing on long edges. For JPEG artifacts, the feature detectors are close in
appearance to the output weights. The feature generators are also somewhat affected
by the type of the noise. This is especially visible for stripe noise, where
the feature generators seem to sometimes also contain stripes.  It was also
observed by \citet{vincent2010stacked} that the type of the noise has a strong
effect of the learned weights in denoising autoencoders.

\begin{figure}[htbp]
  \centering
  \rotatebox{90}{\small output features}
  \includegraphics[width=0.95\textwidth]{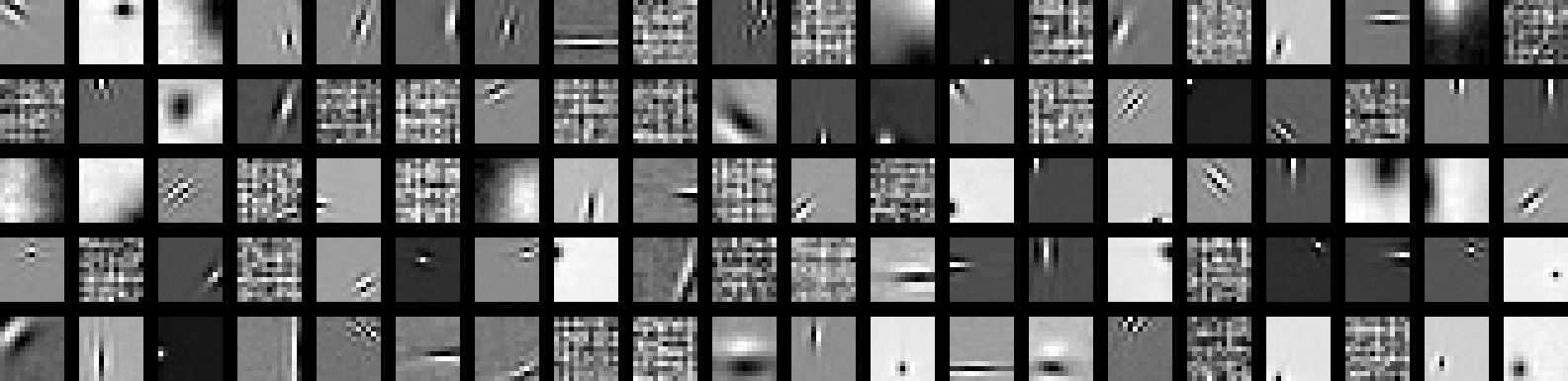}
  \caption{Block matching output weights}
\label{fig:patchesout_bm}
\end{figure}

\subsection{Block-matching filters}
Figure~\ref{fig:patchesout_bm} shows the feature generators learned by the MLP
with block-matching, using $k=14$ and patches of size $13\times13$. The feature
generators look similar to those learned by MLPs without block-matching.

\begin{figure}[htbp]
  \centering
  \rotatebox{90}{\small \hspace{50pt} input features}
  \includegraphics[width=0.95\columnwidth]{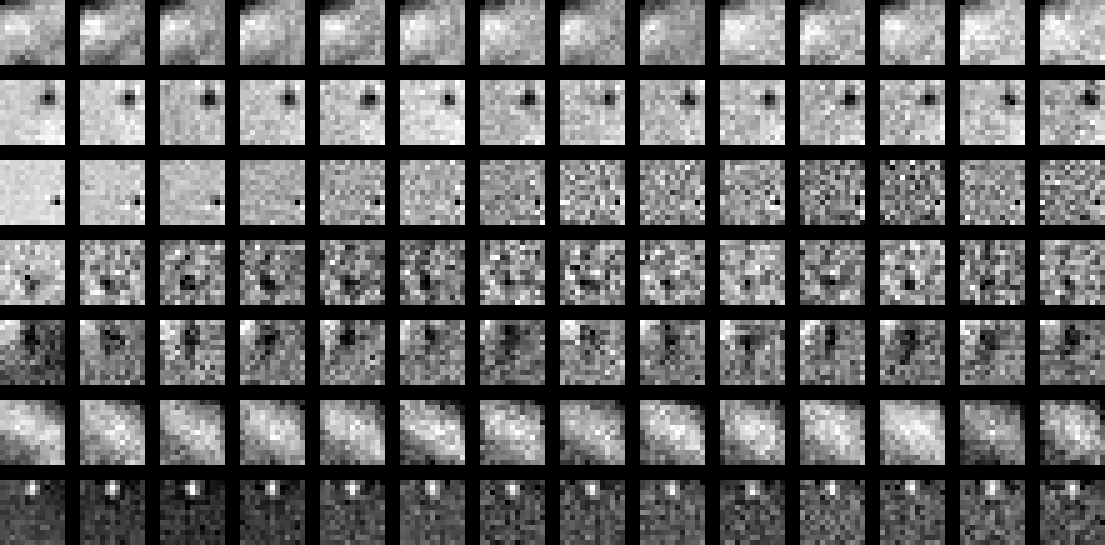}
  \caption{Block matching input weights}
\label{fig:patchesin_bm}
\end{figure}

Figure~\ref{fig:patchesin_bm} shows a selection of feature detectors learned by
the MLP with block-matching. The left-most patch shows the filter applied to
the reference patch, and the horizontally adjacent patches show the filters
applied to the corresponding neighbor patches. The horizontally adjacent
patches all connect to the same hidden neuron. We see that the filters applied
to the neighbor patches are usually similar to the filters applied to the
reference patch. This observation should not be surprising: The updates of the 
weights connecting the input patches to a hidden neuron are defined by (i) the 
gradient at the hidden neuron and (ii) the value of the input pixels. Hence, if
the value of the input pixels are similar (this is ensured by the
block-matching procedure), the weight updates are also similar.

\section{Discussion and Conclusion}
In \citet{burgerjmlr1}, we have shown that it is possible to achieve state-of-the-art
image denoising results using MLPs. In this paper, we have shown how this is
possible. In the first part of this paper, we have discussed which trade-offs
are important during the training procedure. In the second part of this paper,
we have shown that it is possible to gain insight about the inner working of
the trained MLPs by analysing the activation patterns on the hidden units.

\paragraph{How to train MLPs:} We have trained MLPs with varying
architectures on datasets of different sizes.  We have also varied the sizes of
the input as well as of the output patches. The observations made on these
experiments allow us to make a number of conclusions regarding image denoising
with MLPs: (i) More training data is always good, (ii) more hidden units per
hidden layer is always good, (iii) there is an ideal number of hidden layers
for a given problem and a given number of hidden units per hidden layer. Going
above the ideal number of hidden layers can lead to catastrophic degradations
in performance, (iv) increasing the output size requires higher-capacity
architectures, and finally (v) fine-tuning with a lower learning rate can lead 
to important gains in performance.

Other image processing problems such as super-resolution, deconvolution and
demosaicking might also be addressed using MLPs, in which case we expect the
guidelines described in this paper to be useful as well. Other problems
unrelated to images might also benefit from these guidelines. Indeed, we
expect that many difficult problems with high dimensional inputs and outputs 
could benefit from these insights.

\paragraph{Understanding denoising MLPs:} The denoising procedure of MLPs
with a single hidden layer can be briefly summarized as follows. Each hidden
unit detects a feature in the noisy input and copies it to the output patch.
Denoising is achieved through saturation of the $\tanh$-layer. The use of
activation maximization \citep{erhan2010understanding} and observing outputs
obtained by activating a single hidden unit in an MLP allowed us to make
observations concerning the internal workings of MLPs with several hidden
layers. We have seen that MLPs with several hidden layers seem to work
according to the same principle as MLPs with a single hidden layer: The
features required to maximize the activation of a hidden unit are often
remarkably similar to the output caused by the same hidden unit. This 
observation is true for each hidden layer.

Denoising with MLPs requires that the $\tanh$-layer saturates, which naturally
gives rise to binary representations. This is different from RBMs, which force
their hidden representations to be binary. The fact that the representations
are binary lends support to the regularization interpretation of denoising
autoencoders proposed by \citet{erhan2010does}. We also note that binary
representations are unusual for MLPs: Other problems do not give rise to 
binary representations \citep{GlorotAISTATS2010}. 

As an alternative to binary representations, we consider sparse
representations. Sparse representation in higher dimensional spaces have the
well-known benefit of being able to more easily rely on linear operations for a
variety of tasks, see for example \citet{mairal2010online}. Sparsity has been
proposed as a form of regularization to train deep belief networks, see
\citet{marc2007sparse}. Successful architectures for object recognition
\citep{jarretticcv09} also make use of sparse representations, in this case
using a procedure called predictive sparse coding proposed by
\citet{koraypsd08}. In all cases, achieving sparse representations requires
sparsity inducing terms in the optimization criteria, which makes the
optimization procedure more complex. We argue that binary representations have
similar benefits to sparse representations, but that obtaining binary
representations is easier than obtaining sparse representations, using a
denoising criterion.

A further similarity between MLPs trained to denoise images and RBMs and
denoising autoencoders is the similarity of the features (such as Gabor
filters) learned by all three architectures. Unrelated approaches such KSVD
\citep{aharon2006rm} learn similar features.

\vskip 0.2in
\bibliography{mybib}

\end{document}